\pdfoutput=1
\PassOptionsToPackage{dvipsnames,table}{xcolor} 

\documentclass[11pt]{article}

\usepackage[final]{acl}

\usepackage{times}
\usepackage{latexsym}
\usepackage[T1]{fontenc}
\usepackage[utf8]{inputenc}
\usepackage{microtype}          
\usepackage{inconsolata}
\usepackage{graphicx}

\newcommand*{\img}[1]{%
    \raisebox{-.2\baselineskip}{%
        \includegraphics[
        height=\baselineskip,
        width=\baselineskip,
        keepaspectratio,
        ]{#1}%
    }%
}

\usepackage{amsfonts}
\usepackage{amsmath}
\usepackage{amssymb}
\usepackage{booktabs}

\usepackage{algorithm}
\usepackage{algpseudocode}

\usepackage{xcolor}

\usepackage{mathtools}
\usepackage{colortbl}
\usepackage{array}
\usepackage{enumitem}
\usepackage{comment}
\usepackage{multirow}
\usepackage{tcolorbox}
\usepackage{subcaption}

\title{Text Meets Topology \img{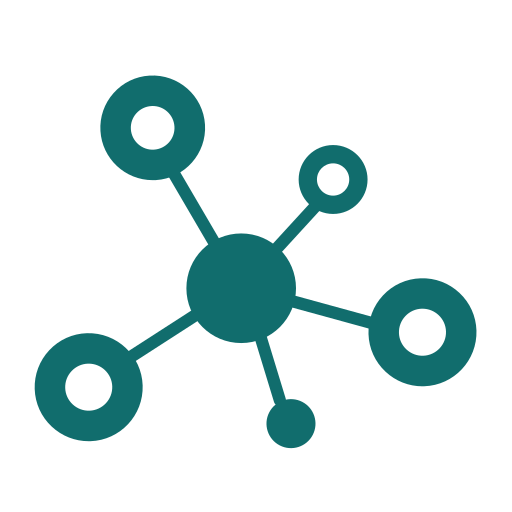}: Rethinking Out-of-distribution Detection in Text-Rich Networks}

\author{Danny Wang, Ruihong Qiu, Guangdong Bai \and Zi Huang\\
        The University of Queensland \\ \texttt{\{danny.wang,r.qiu,g.bai,helen.huang\}@uq.edu.au}}

\begin{document}
\maketitle
\begin{abstract}

Out-of-distribution (OOD) detection remains challenging in \textbf{\textit{text-rich networks}}, where textual features intertwine with topological structures. Existing methods primarily address label shifts or rudimentary domain-based splits, overlooking the intricate textual-structural diversity. For example, in social networks, where users represent nodes with textual features (name, bio) while edges indicate friendship status, OOD may stem from the distinct language patterns between bot and normal users. To address this gap, we introduce the \textbf{TextTopoOOD} framework for evaluating detection across diverse OOD scenarios: \textbf{(1) attribute-level shifts} via text augmentations and embedding perturbations; \textbf{(2) structural shifts} through edge rewiring and semantic connections; \textbf{(3) thematically-guided label shifts}; and \textbf{(4) domain-based divisions}. Furthermore, we propose \textbf{TNT-OOD} to model the complex interplay between \textbf{T}ext a\textbf{N}d \textbf{T}opology  using: \textbf{1) a novel cross-attention module} to fuse local structure into node-level text representations, and \textbf{2) a HyperNetwork} to generate node-specific transformation parameters. This aligns topological and semantic features of ID nodes, enhancing ID/OOD distinction across structural and textual shifts. Experiments on \textbf{11 datasets across four OOD scenarios} demonstrate the nuanced challenge of \textbf{TextTopoOOD} for evaluating OOD detection in text-rich networks.
\footnote{Code is available at \href{https://github.com/DannyW618/TNT}{https://github.com/DannyW618/TNT}.}
\end{abstract}

\section{Introduction}

Text-rich networks (TrN) have emerged as a powerful paradigm for representing the complex interplay between textual content and relational structures, serving as the lingua franca for modeling intricate real-world systems across diverse domains~\cite{patton, TRN_Paper, CaseGNN, CaseLink}. 
Despite the ubiquity of TrNs, machine learning approaches for these hybrid data structures often fail catastrophically when confronted with data that deviates from their training distribution (i.e., out-of-distribution (OOD))~\cite{GOOD, GOODSurvey, GOODAT}. To address this challenge, a surge of effective OOD detection techniques have been proposed~\cite{GKDE, AAGOD, OODGAT, DEGEM,GOOD-D,Dream-OOD}, aiming to identify instances that fall outside the in-distribution (ID) training data.

Nevertheless, the OOD detection problem remains underexplored for TrNs~\cite{NLPOODD, GOODD-survey, OODD_cv_survey}.
\textbf{Existing methods for OOD detection} in TrN learning primarily focus on \textbf{common distribution shifts} like random label-leave-out or temporal splits, largely ignoring the rich textual dimension that characterises real-world networks~\cite{OOD_TAG, Gsyn_tag}. This oversight is critical: in TrNs, semantic shifts in text may precipitate network changes. In product co-purchase networks, for example, shifting product descriptions and customer reviews often precede changes in purchasing patterns and product relationships - shifts that current OOD detection methods fail to capture. 

To address this critical gap, we introduce \textbf{TextTopoOOD}, a comprehensive evaluation framework for TrNs. Unlike previous benchmarks~\cite{GLIP-OOD, OOD_TAG}, TextTopoOOD explores multiple dimensions of distribution shifts through: \textbf{1) Attribute-level modifications} that simulate semantic drift in textual content; 
\textbf{2) Structural alterations }that capture diverse connectivity patterns; 
\textbf{3) Thematically-guided label shifts} that extends beyond random selection; and \textbf{4) Domain-based splits} based on dataset's unique properties.

Additionally, prior OOD detection models for structural data typically add post-hoc scoring functions to the classifier~\cite{GNNSafe, NODESAFE, grasp}. While these methods implicitly integrate text and structure through end-to-end training, the shared projection heads and uniform objectives~\cite{MSP, energy} often fail to capture distribution shifts that manifest differently across modalities and nodes.

In light of this, we propose \textbf{TNT-OOD}, a novel framework that models the interplay between \textbf{T}ext a\textbf{N}d \textbf{T}opology. TNT-OOD consists of three key components: \textbf{(1) a structure encoder} that learns structure-aware representations from local neighborhoods, \textbf{(2) a cross-attention mechanism} that fuses structure-derived and text-derived features to produce contextually grounded representations, and \textbf{(3) a HyperNetwork} that generates effective projection parameters to align the fused representations in a contrastive embedding space. This enables TNT-OOD to model heterogeneity at the node level, capturing textual-topological interactions that static encoders or global projection heads come short. Thereby, improving the separability of ID and OOD data when we encounter misalignment from the textual and topological components. We empirically evaluate TNT-OOD on our TextTopoOOD framework with diverse datasets and OOD scenarios, demonstrating consistent improvements over baselines. Our contributions are:
\begin{itemize}[leftmargin=*, itemsep=0pt, topsep=2pt, parsep=0pt, partopsep=0pt]
\item We introduce \textbf{TextTopoOOD}, the first comprehensive framework for evaluating OOD detection in text-rich networks that captures the interplay between textual features and network topology.
\item We propose \textbf{TNT-OOD}, an effective TrN OOD detector that aligns textual and structural representations using HyperNetwork.
\item We validate the \textbf{efficacy of TNT-OOD for OOD detection} and the \textbf{challenging nature of TextTopoOOD} via 11 TrNs and four OOD scenarios.
\end{itemize}

\section{Preliminaries}
\label{sec:preliminaries}
A \textbf{Text-Rich Network (TrN)} is denoted as $\mathcal{G}=(\mathcal{V}, \mathcal{T}, \mathbf{A})$, where $\mathcal{V}=\{v_1, \ldots, v_n\}$ is the set of $n$ nodes. Each node $v_i$ has a textual content $t_i \in \mathcal{T}$, encoded into a feature vector $\mathbf{x}_i \in \mathbb{R}^d$ through a text encoder (e.g., SBERT) $f_{\text{encoder}}: \mathcal{T} \rightarrow \mathbb{R}^d$. This forms the feature matrix $\mathbf{X} = [\mathbf{x}_1, \ldots, \mathbf{x}_n]^T \in \mathbb{R}^{n \times d}$. The network topology is captured by an adjacency matrix $\mathbf{A} \in \mathbb{R}^{n \times n}$, where $A_{ij} = 1$ indicates a connection between nodes $v_i$ and $v_j$. The node labels $\mathbf{Y} = \{y_1, \ldots, y_n\}$ assign each node to one of $C$ classes, with $y_i \in \{1, \ldots, C\}$. This paper studies TrNs with semantic relationships between nodes (e.g., \textbf{in e-commerce networks}, nodes represent products with textual descriptions, connected by co-purchase relationships). \textit{\textbf{In this paper, we explore two objectives:}}
\begin{figure}[t!]
\centering
    \includegraphics[width=1\columnwidth]{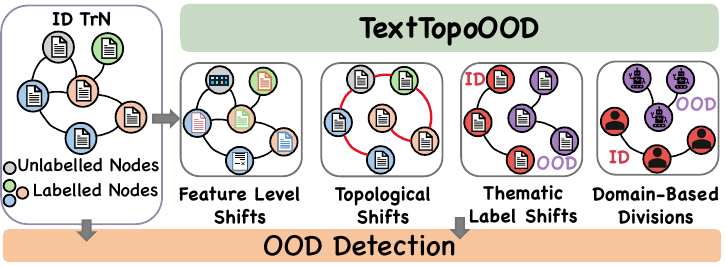}
\caption{Overview of TextTopoOOD framework%
.}
\label{fig:framework}
\end{figure}
\paragraph{Objective 1: In-distribution Class Classification.}
Given the training and test nodes sharing the same distribution, where $P_{train}({\mathcal{G}_{train}}) = P_{test}(\mathcal{G}_{test})$,  
and the conditional distribution $ P_{train}(\mathbf{y}|\mathcal{G}_{train}) = P_{test}(\mathbf{y}|\mathcal{G}_{test})$, 
we aim to design a \textbf{classifier $\boldsymbol{g}$} to accurately predict the label $ \mathbf{y} \in \mathbb{R}^n $ for the test nodes, such that:
$\mathbf{y} = \text{Softmax}(g(\mathcal{G}_{train})).$
\paragraph{Objective 2: Node-Level Out-of-distribution Detection.} 
The goal of OOD detection is to detect nodes with a different distribution to the training data at test time. Typically, we consider OOD shifts as $P_{train}({\mathcal{G}_{train}}) \neq P_{test}(\mathcal{G}_{test})$, and $ P_{train}(\mathbf{y}|\mathcal{G}_{train}) \neq P_{test}(\mathbf{y}|\mathcal{G}_{test})$. The task is to formulate a \textbf{detector $\boldsymbol{F}$} with an \textbf{OOD scoring function $\boldsymbol{S}$} and a given threshold $\tau_{\text{thresh}}$, such that: %
\begin{equation}  
    F\left(\mathcal{G}; g\right)=  
    \begin{cases}  
        \text{OOD}, & S(\mathcal{G};g) \ge \tau_{\text{thresh}}, \\  
        \text{ID}, & S(\mathcal{G};g) < \tau_{\text{thresh}}.  
    \end{cases}  
    \label{eq:ood_criteria}  
\end{equation}

\paragraph{Energy-Based OOD Detection.}
\label{subsec:energy_ood}
For a node $v_i$ in a TnR, the energy score is defined as:
\begin{equation}  
    S(\mathcal{G}_{v_i};g) = e_i = - \log \sum\nolimits_{c=0}^{C-1}{\exp(\mathbf{z}_{i,c})},
    \label{eq:energy}
\end{equation}
where $e_i \in \mathbb{R}$ is the energy score, and $\mathbf{z}_i \in \mathbb{R}^C$ are the logits from the classifier $\mathbf{Z}=g(\mathcal{G})$. 

To leverage topology in OOD detection, \textbf{energy propagation} is proposed with ~\cite{GNNSafe}:
\begin{equation}  
    \mathbf{e}^{(k)} = \alpha \mathbf{e}^{(k-1)} + (1-\alpha) \mathbf{D}^{-1} \mathbf{A} \mathbf{e}^{(k-1)},
    \label{eq:energy_prop}
\end{equation}
where $\mathbf{e}^{(k)}$ represents node energy scores after $k$ propagation steps, $\alpha$ controls energy concentration, and $\mathbf{D}$ is the degree matrix, where $\mathbf{D}_{ii}=\sum^n_{j=1}\mathbf{A}_{ij}$ and $\mathbf{D}_{ij}=0$ for $i\neq j$.

\begin{table*}[t!]
\centering
\setlength{\fboxsep}{8pt}
\setlength{\tabcolsep}{6pt}
\begin{tabular}{p{0.31\textwidth}|p{0.31\textwidth}|p{0.31\textwidth}}
\multicolumn{1}{c|}{\textbf{Original Text}} & 
\multicolumn{1}{c|}{\textbf{Synonym ($\alpha=0.5, p_{char}=0.3$)}} & 
\multicolumn{1}{c}{\textbf{Antonym ($\alpha=0.3, p_{char}=0.3$)}} \\
\midrule
\rowcolor{gray!7}
Recent advances in natural language processing have enabled more efficient text encoding for downstream applications. &
\textcolor{blue}{Curret} advanc\textcolor{blue}{z}es in \textcolor{blue}{normal} language \textcolor{blue}{handlin} have \textcolor{blue}{faacilitated} more \textcolor{blue}{faster} text \textcolor{blue}{representation} for dow\textcolor{blue}{sn}tream \textcolor{blue}{tasks}. &
\textcolor{red}{Ancient} advances in n\textcolor{red}{ta}ural language \textcolor{red}{a}processing have \textcolor{red}{disabled} more \textcolor{red}{inefficient} text encding for \textcolor{red}{upstream} appications. \\
\midrule
I tried both the new and old reddit design and its like that. & 
I tried \textcolor{blue}{either} th \textcolor{blue}{fres} and old r\textcolor{blue}{de}dit \textcolor{blue}{plaan} and its \textcolor{blue}{similar} that.&
I tried bth the \textcolor{red}{old} and \textcolor{red}{young} reddit design and its \textcolor{red}{different} that.\\
\end{tabular}
\vspace{-0.3cm}
\caption{Example of text-level feature shift created by \textsc{TextAugmenT}.}
\label{table:text_augmentation_examples}
\vspace{-0.5cm}
\end{table*}

\section{TextTopoOOD Framework} \label{sec:TextTopoOOD}
This section introduces \textbf{TextTopoOOD}, a comprehensive framework that explores diverse OOD scenarios in text-rich networks. 
TextTopoOOD analyses OOD scenarios across four dimensions: \textbf{(1) attribute-level shifts}, \textbf{(2) structural shifts}, \textbf{(3) thematically-guided label shifts}, and \textbf{(4) domain-based divisions} as shown in Figure~\ref{fig:framework}. %

\subsection{Attribute-Level Shifts}
\label{sec:feature_shifts}
\textbf{Attribute-level shifts targets textual or feature components}, challenging models to detect semantic changes while preserving network structure. For example, in product networks, descriptions might use new marketing terminology (`eco-friendly' $\rightarrow$ `sustainable') while co-purchase relationships remain unchanged - requiring models to recognise semantic evolution beyond structural cues. We provide the following attribute-level shift scenarios.
\paragraph{1) Text Augmentation Shift.} \label{sec:text_aug}
We generate text augmentation shifts by modifying the nodes' raw textual content. Given a node's original text $t_i$, we produce a perturbed version $\tilde{t}_i$ using controlled semantic transformations:
\vspace{-0.2cm}
\begin{equation}
\tilde{t}_i = \textsc{TextAugment}(t_i, \text{type}, \alpha_{\text{text}}, p_{\text{char}})
\label{eq:textaugment}
\vspace{-0.2cm}
\end{equation}
where $\text{type} \in \{\text{synonym}, \text{antonym}\}$ determines the semantic direction of the shift, and $\alpha_{\text{text}} \in [0,1]$ controls the noise level or percentage of words modified, $p_{\text{char}}$ indicates the probability of character-level edits (insertion, deletion, replacements, swaps). We provide the pseudo-code of the \textsc{TextAugment} function in Appendix~\ref{Appendix:text_augment}.

This approach creates OOD text with preserved syntactic structure but varied semantics. Synonyms challenge detection by maintaining textual meaning while altering word distributions from training data. Antonyms create more pronounced shifts by inverting meanings while preserving grammar, producing contextual inconsistencies. Character-level noise simulates typical human typographical errors. Examples are shown in Table~\ref{table:text_augmentation_examples} and Appendix~\ref{Appendix:ood_shifts}.

\paragraph{2) Feature Mixing Shift.} \label{sec:feat_mix}
Following prior works, we implement feature mixing to manipulate the encoded embeddings beyond raw texts~\cite{GNNSafe}. For each encoded node feature vector $\mathbf{x}_i$, we generate a perturbed version:
\begin{equation}
\tilde{\mathbf{x}}_i = (1 - \alpha_{\text{feat}}) \cdot \mathbf{x}_i + \alpha_\text{feat} \cdot \mathbf{m}_i,
\end{equation}
where $\mathbf{m}_i = w \cdot \mathbf{x}_j + (1-w) \cdot \mathbf{x}_k$ is a convex combination of features from randomly selected nodes $j$ and $k$, with $w \sim \text{Uniform}(0,1)$. The parameter $\alpha_\text{feat}$ controls shift intensity, with higher values creating embeddings that increasingly deviate from the original distribution to become OOD. %

\subsection{Structural Shifts}
\label{sec:structural_shifts}
\textbf{Structural shifts alter network connectivity while preserving node attributes}, testing models' ability to detect topological changes. In product networks, co-purchase relationships evolve with changing consumer preferences (sustainable items becoming frequently co-purchased), while product descriptions remain unchanged - creating distribution shifts that are topological than semantic.

\paragraph{1) Structure Rewiring.} \label{sec:struct_rewire}
We implement structure shifts using stochastic block models (SBM)~\cite{SBM} to generate alternative edge distributions that reflect different community structures. Given the original network $\mathcal{G}_{\text{ID}}$ with edge density $\rho$, we generate an SBM network with node classes (determined by labels $\mathbf{Y}$) as blocks. The probability matrix is defined as:
\begin{equation}
P_{ij} = 
\begin{cases}
p_{ii} = \rho \cdot f_{ii} & \text{if } i = j \\
p_{ij} = \rho \cdot f_{ij} & \text{if } i \neq j
\end{cases}
\end{equation}
where $f_{ii}$ and $f_{ij}$ are scaling factors that control intra-block and inter-block connectivity. The resulting SBM adjacency matrix $\mathbf{A}_{\text{SBM}}$ are then interpolated with the original structure $\mathbf{A}$ following:
\begin{equation}
\mathbf{A}_{\text{OOD}} = (1-\beta) \cdot \textsc{Sample}(\mathbf{A}) \cup \beta \cdot \textsc{Sample}(\mathbf{A}_{\text{SBM}})
\end{equation}
where $\textsc{Sample}(\cdot)$ selects edges proportionally to maintain original graph density, and $\beta$ controls the shift intensity.

This alters community structures while preserving textual content. In citation networks, this resembles when research papers maintain their topics but form new citation patterns with previously unrelated work, controlled via $f_{ii}$ and $f_{ij}$ parameters to reinforce or contradict expected content-connectivity relationships.

\paragraph{2) Semantic Connection Shift.} \label{sec:semantic_connect}
The semantic connection shift reconnects  
the network based on node feature similarities, creating a correlation between textual semantics and connectivity patterns:
\begin{equation}
\mathbf{A}_{\text{OOD}} = \textsc{TopK}(S(\mathbf{X}, \mathbf{X}), k, \textsc{mode})
\end{equation}
where $S(\mathbf{X}, \mathbf{X})$ is a pairwise cosine similarity matrix between encoded text representations (e.g., $s_{ij} = \text{CosSim}(\mathbf{x}_i, \mathbf{x}_j)$, and $k 
$ determines the number of edges to select (i.e.,same density as the original network). The selection \textsc{mode} $\in \{\text{top}, \text{bottom}, \text{threshold percentile}\}$ determines which similarity pairs connect.

\paragraph{3) Text Swap Shift.} \label{sec:text_swap}
The text swap shift introduces semantic inconsistencies by exchanging text between nodes, implicitly affecting the structural patterns. This creates a mismatch between node content and network position, challenging models to detect contextual incongruities. We formalise this as a controlled feature permutation operation:
\vspace{-0.2cm}
\begin{equation}
\tilde{\mathbf{X}} = \mathbf{P}_{\beta_{\text{swap}}} \cdot \mathbf{X},
\vspace{-0.2cm}
\end{equation}
where $\mathbf{P}_{\beta_{\text{swap}}}$ is a permutation matrix to swaps features between a proportion $\beta_{\text{swap}} \in [0,1]$ of nodes.

As shown in Algorithm~\ref{alg:text_swap} in Appendix, we offer three variants to \textbf{swap texts between}: \textbf{1) Intra-class:} Nodes of the same class; \textbf{2) Inter-class:} Nodes of different classes; \textbf{3) Random:} Any nodes.
This simulates real-world scenarios such as hijacked social media accounts posting out-of-character content, citation networks with papers incorrectly categorised, or product listings with mismatched descriptions. The progressive severity of misalignment allows systematic evaluation of model sensitivity to text-structure inconsistencies.

\subsection{Label Shift with Thematic Guidance} \label{sec:label_shift}
\textbf{We extend the traditional label shift approach beyond random selection} by incorporating thematic analysis through large language models (LLMs) (i.e., Claude)~\cite{Claude}. Given the complete set of classes $\mathcal{C}$, we create OOD datasets by withholding specific numbers of classes ($|\mathcal{C}_\text{OOD}| \approx 10\% \sim 40\%$ of $|\mathcal{C}|$) according to their thematic relationships:
\vspace{-0.2cm}
\begin{equation}
\mathcal{C}_{\text{ID}} = \mathcal{C} \setminus \mathcal{C}_{\text{OOD}}.
\vspace{-0.2cm}
\end{equation}
We implement 3 strategies for OOD class selection:
\paragraph{1) Random selection:} Classes are randomly designated as OOD, serving as a baseline approach.

Beyond this, we use LLMs to analyse the class names and descriptions to selects OOD classes by:
\paragraph{2) Thematic similarity:} Separating classes that are pairwise similar into ID and OOD, creating scenarios where there exists similar label concepts.
\paragraph{3) Thematic dissimilarity:} Grouping classes that most dissimilar to ID classes as OOD, creating scenarios with clearer semantic boundaries between ID and OOD.

For example, closely related topics (i.e., ML \& Theory) may share terminology and citation patterns, making them more challenging to distinguish than completely unrelated domains (i.e., ML \& Music). The LLM prompt is provided in Appendix~\ref{Appendix:ood_shifts}.

\subsection{Domain-Based Divisions}
\label{sec:domain_shifts}
Text-rich networks spans diverse domains, with unique properties that could naturally be used to devise OOD data. An example domain-based division using \textbf{temporal information} is as follows: %
\paragraph{1) Temporal Shift.}
For datasets with temporal information (e.g., arXiv citation network), we create domain shifts based on publication time:
\begin{align}
\mathcal{V}_{\text{ID}} &= \{v_i \in \mathcal{V} \mid \text{year}(v_i) \in r_{\text{ID}} \} \\
\mathcal{V}_{\text{OOD}} &= \{v_i \in \mathcal{V} \mid \text{year}(v_i) \in r_{\text{OOD}} \}, 
\end{align}
where $r_{\text{ID}}, r_{\text{OOD}}$ are disjoint time ranges. This approach creates a series of OOD instances representing increasingly distant future data, simulating the natural evolution of text-rich networks over time. 

\subsection{Comparison to Related Benchmarks}
To highlight TextTopoOOD's novelty, we contrast it with existing benchmarks. OOD-TAG~\cite{OOD_TAG} introduced shifts that are derived from common citation network properties like node degree, temporal splits, and word diversity. Recent works like GLIP~\cite{GLIP-OOD} and GSync-OOD~\cite{Gsyn_tag} focus solely on random label shifts, lacking comprehensive scenario coverage. In contrast, TextTopoOOD provides a principled evaluation framework that \textbf{explicitly targets both textual and topological dimensions} across diverse network types, including citation, e-commerce, knowledge, and social networks.

\section{TNT-OOD Methodology}
\begin{figure}[!t]
    \centering
    \vspace{-0.3cm}
    \resizebox{0.5\textwidth}{!}{%
        \includegraphics{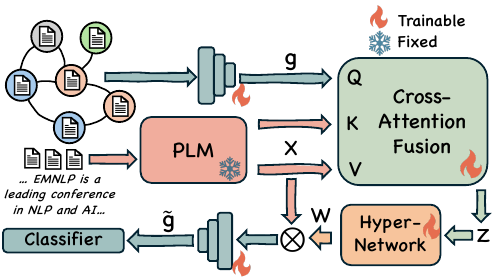}
    }
    \caption{TNT-OOD framework: 1) Structure encoders and Classifier; 2) Text-encoder PLM; 3) Cross-attention fusion module; 4) HyperNetwork projection.}
    \label{fig:tnt_framework}
    \vspace{-0.5cm}
\end{figure}

To address the complex interplay between text and topology, we propose TNT-OOD: a HyperNetwork-enhanced, cross-attentional model for OOD detection in text-rich networks. As shown in Figure~\ref{fig:tnt_framework}, the architecture comprises: \textbf{1) a GCN-based structure encoder} for structure-aware representation, \textbf{2) a Cross-Attention module} that adaptively fuses textual and graph context, and \textbf{3) a novel HyperNetwork Projection Head} that generates dedicated weights conditioned on fused representations, enabling heterogeneous alignment.

\paragraph{1) Structure and Textual Encoding.}
Let $\mathbf{x}_i \in \mathbb{R}^{d}$ denote the frozen textual embedding of node $v_i$ (e.g., via SBERT~\cite{sbert}). The structure encoder is a $(L)$-layer Graph Convolutional Network (GCN) that transforms $\mathbf{x}_i$ via neighborhood aggregation~\cite{GCN}. The layer-wise update with $n$ nodes is:
\vspace{-0.2cm}
\begin{equation}
\mathbf{g}^{(l)} = \sigma \left( \hat{\mathbf{A}} \mathbf{g}^{(l-1)} \mathbf{W}^{(l)} \right), \quad l = 1, \dots, L,
\vspace{-0.2cm}
\end{equation}
with $\mathbf{g}^{(0)} = \mathbf{X} \in \mathbb{R}^{n \times d}$, and $\hat{\mathbf{A}}$ is the symmetrically normalised adjacency matrix with self-loops, $ \mathbf{W}^{(l)}$ are trainable weights for layer $l$. This produces structural-aware embeddings $\mathbf{g} \in \mathbb{R}^{n \times d_p}$, where $d_p$ is the projection dimension.

\paragraph{2) Cross-Attentional Fusion.}
To infuse textual embeddings with localised graph-aware context, we introduce a \textbf{neighborhood-based cross-attention module}. For each node $v_i$, we aggregate textual information from its neighbors $\mathcal{N}(i)$ conditioned on the learned structural representations:
\vspace{-0.2cm}
\begin{equation}
\mathbf{q} = \mathbf{W}_q \mathbf{g}, \quad \mathbf{k} = \mathbf{W}_k \mathbf{x}, \quad \mathbf{v} = \mathbf{W}_v \mathbf{x},
\vspace{-0.2cm}
\end{equation}
\begin{equation}
\mathbf{z}_i = \mathbf{x}_i + \sum_{j \in \mathcal{N}(i)} \text{Softmax}_j\left( \frac{\langle \mathbf{q}_i, \mathbf{k}_j \rangle}{\sqrt{d_z}} \right) \cdot \mathbf{v}_j,
\label{eq:cross_attention}
\end{equation}
where $\mathbf{W}_q$, $\mathbf{W}_k$, and $\mathbf{W}_v$ are learnable projection matrices. This yields a fused representation $\mathbf{z}_i$ with structure-aware attention over neighbour content.

\paragraph{3) HyperNetwork-based Projection.}
Furthermore, to capture the difference between ID and OOD samples, we employ a HyperNetwork to generate \textit{node-specific projections} from the fused text-structure features. 
The motivation is that using projection weights learned from ID data would expose misalignment when transforming OOD nodes. The HyperNetwork is defined as:
\begin{equation}
\mathbf{W}_i = \text{MLP}_{\text{hyper}}\left( \mathbf{z}_i \right),
\end{equation}
where $\mathbf{W}_i \in \mathbb{R}^{d_p \times d}$ is a node-specific weight with projection dimension $d_p$. The projected text representation for each node $\mathbf{p}^i_t \in \mathbb{R}^{d_p}$ is given by:
\begin{equation}
\mathbf{p}_t^i = \mathbf{W}_i \mathbf{x}_i.
\end{equation}
To ensure efficiency when working with large scale networks (i.e., over 100,000 nodes), we generate a factorised low-rank representation:
\begin{equation}
\left(\mathbf{L}_i, \mathbf{R}_i\right) = \text{LowRankHyper}_{\text{MLP}}\left( \mathbf{z_i} \right),
\end{equation}
where $\mathbf{L}_i \in \mathbb{R}^{d_p \times r}$ and $\mathbf{R}_i \in \mathbb{R}^{r \times d}$ are the left and right factors respectively node $v_i$, $r$ is the low-rank parameter ($r \ll \min(d_p, d)$). The projected text representation is computed efficiently using:
\begin{equation}
\mathbf{p}^i_t = \mathbf{L}_i \cdot (\mathbf{R}_i \cdot \mathbf{x}_i).
\end{equation}
This approach reduces memory cost from $O(d_p \times d)$ to $O(d_p \times r + r \times d)$, enable efficient training. The projected representation is further passed through a GCN layer to re-enable structure-aware learning:
\begin{equation}
\tilde{\mathbf{g}} = \text{GCN}_{\text{fuse}}(\mathbf{P}_t, \mathbf{A})
\label{eq:post_gnn_encoder}
\end{equation}

\begin{table*}[t!]
\vspace{-0.2cm}
\centering
\resizebox{\textwidth}{!}{%
\begin{tabular}{c|r|ccccc||cc||cccc}
\toprule
\multirow{2}{*}{} & \multirow{2}{*}{\textbf{Metrics}} & \multicolumn{5}{c||}{\textit{\textbf{Citation Networks}}} & \multicolumn{2}{c||}{\textit{\textbf{Knowledge \& Social Networks}}} & \multicolumn{4}{c}{\textit{\textbf{E-commerce Networks}}}\\
 &  & \textbf{Cora} & \textbf{Citeseer} & \textbf{Arxiv} & \textbf{DBLP} & \textbf{PubMed} & \textbf{Reddit} & \textbf{WikiCS} & \textbf{Bookhis} & \textbf{Bookchild} & \textbf{Elephoto} & \textbf{Elecomp} \\
\midrule
\multirow{4}{*}{\rotatebox{90}{Maha}}
& AUROC $(\uparrow)$ 
    & 41.74 $\pm$ 9.24 
    & 50.63 $\pm$ 24.39 
    & 67.75 $\pm$ 35.39 
    & 58.47 $\pm$ 24.67
    & 55.32 $\pm$ 20.70 
    & 42.49 $\pm$ 19.65 
    & 52.40 $\pm$ 24.19 
    & 66.65 $\pm$ 29.71 
    & 68.96 $\pm$ 31.61 
    & 62.03 $\pm$ 26.03 
    & 43.97 $\pm$ 16.64 \\
& AUPR $(\uparrow)$ 
    & 48.75 $\pm$ 4.20 
    & 55.41 $\pm$ 13.37 
    & 78.75 $\pm$ 28.41 
    & 55.74 $\pm$ 12.53
    & 54.87 $\pm$ 14.89 
    & 51.84 $\pm$ 10.14 
    & 66.68 $\pm$ 13.61 
    & 89.08 $\pm$ 11.22 
    & 88.51 $\pm$ 12.73 
    & 86.97 $\pm$ 10.21 
    & 87.01 $\pm$ 4.18 \\
& FPR95 $(\downarrow)$ 
    & 92.15 $\pm$ 4.36 
    & 87.11 $\pm$ 15.16
    & \textbf{56.92 $\pm$ 44.11}
    & 75.73 $\pm$ 22.72
    & 88.13 $\pm$ 8.76
    & 88.77 $\pm$ 5.55 
    & 82.65 $\pm$ 15.50 
    & 70.22 $\pm$ 33.94 
    & \underline{59.06 $\pm$ 42.84 }
    & 76.87 $\pm$ 34.46 
    & 89.94 $\pm$ 13.53 \\
& ID ACC $(\uparrow)$ 
    & 86.21 $\pm$ 5.48 
    & 79.34 $\pm$ 7.47 
    & 68.18 $\pm$ 0.51 
    & 77.41 $\pm$ 0.21 
    & 78.23 $\pm$ 0.41 
    & 61.85 $\pm$ 2.32 
    & 82.22 $\pm$ 4.18 
    & 83.95 $\pm$ 0.89 
    & 58.52 $\pm$ 5.87 
    & 82.71 $\pm$ 5.81 
    & 82.58 $\pm$ 7.93\\
\midrule
\multirow{4}{*}{\rotatebox{90}{MSP}}
& AUROC $(\uparrow)$ 
    & 74.63 $\pm$ 5.05 
    & 79.33 $\pm$ 9.67 
    & 70.26 $\pm$ 14.51 
    & 84.34 $\pm$ 5.59
    & 70.49 $\pm$ 5.97 
    & 47.69 $\pm$ 12.05 
    & 71.30 $\pm$ 4.72 
    & 69.65 $\pm$ 11.95 
    & 64.49 $\pm$ 13.06 
    & 70.57 $\pm$ 12.81 
    & 69.05 $\pm$ 15.22 \\
& AUPR $(\uparrow)$ 
    & 75.58 $\pm$ 3.52 
    & 81.19 $\pm$ 8.66 
    & 81.19 $\pm$ 21.03 
    & \underline{83.68 $\pm$ 6.55}
    & 66.57 $\pm$ 5.09 
    & 55.35 $\pm$ 6.41 
    & 78.41 $\pm$ 3.47 
    & 90.44 $\pm$ 4.63 
    & 85.95 $\pm$ 4.38 
    & 89.69 $\pm$ 4.30 
    & \underline{94.20 $\pm$ 2.79} \\
& FPR95 $(\downarrow)$ 
    & 64.27 $\pm$ 9.83 
    & 61.97 $\pm$ 23.93 
    & 74.53 $\pm$ 23.04 
    & 51.13 $\pm$ 13.55 
    & \underline{70.75 $\pm$ 11.72 }
    & 90.84 $\pm$ 3.03 
    & 68.15 $\pm$ 11.83 
    & 68.58 $\pm$ 19.97 
    & 71.27 $\pm$ 25.22 
    & 72.90 $\pm$ 19.65 
    & 70.18 $\pm$ 30.06 \\
& ID ACC $(\uparrow)$ 
    & 86.21 $\pm$ 5.48 
    & 79.34 $\pm$ 7.47 
    & 68.18 $\pm$ 0.51 
    & 77.41 $\pm$ 0.21 
    & 78.23 $\pm$ 0.41 
    & 61.85 $\pm$ 2.32 
    & 82.22 $\pm$ 4.18 
    & 83.95 $\pm$ 0.89 
    & 58.52 $\pm$ 5.87 
    & 82.71 $\pm$ 5.81 
    & 82.58 $\pm$ 7.93\\
\midrule
\multirow{4}{*}{\rotatebox{90}{ODIN}}
& AUROC $(\uparrow)$ 
    & 73.71 $\pm$ 4.60 
    & 78.96 $\pm$ 9.13 
    & 69.73 $\pm$ 14.99 
    & 83.44 $\pm$ 6.94
    & 70.65 $\pm$ 4.12 
    & 46.86 $\pm$ 16.17 
    & 63.28 $\pm$ 4.05 
    & 68.27 $\pm$ 12.01 
    & 60.33 $\pm$ 13.64 
    & 70.14 $\pm$ 13.89 
    & \underline{69.09 $\pm$ 18.31} \\
& AUPR $(\uparrow)$ 
    & 74.75 $\pm$ 3.49 
    & 81.09 $\pm$ 7.94 
    & 80.96 $\pm$ 21.06 
    & 82.25 $\pm$ 8.63 
    & 66.73 $\pm$ 3.86 
    & 55.04 $\pm$ 8.17 
    & 71.81 $\pm$ 3.29 
    & 89.90 $\pm$ 4.73 
    & 84.10 $\pm$ 4.41 
    & 89.74 $\pm$ 4.72 
    & \textbf{94.41 $\pm$ 3.45} \\
& FPR95 $(\downarrow)$ 
    & 66.01 $\pm$ 8.61 
    & 63.32 $\pm$ 22.88 
    & 75.03 $\pm$ 24.10 
    & 51.82 $\pm$ 14.53 
    & 72.73 $\pm$ 8.88 
    & 89.06 $\pm$ 4.73 
    & 75.29 $\pm$ 7.97 
    & 70.21 $\pm$ 19.46 
    & 74.77 $\pm$ 23.13 
    & 74.65 $\pm$ 20.62 
    & 68.95 $\pm$ 33.84 \\
& ID ACC $(\uparrow)$ 
    & 86.21 $\pm$ 5.48 
    & 79.34 $\pm$ 7.47 
    & 68.18 $\pm$ 0.51 
    & 77.41 $\pm$ 0.21 
    & 78.23 $\pm$ 0.41 
    & 61.85 $\pm$ 2.32 
    & 82.22 $\pm$ 4.18 
    & 83.95 $\pm$ 0.89 
    & 58.52 $\pm$ 5.87 
    & 82.71 $\pm$ 5.81 
    & 82.58 $\pm$ 7.93\\
\midrule
\multirow{4}{*}{\rotatebox{90}{NECO}}
& AUROC $(\uparrow)$ 
    & 70.77 $\pm$ 6.35 
    & 68.35 $\pm$ 12.47 
    & \underline{75.19 $\pm$ 14.92}
    & 82.89 $\pm$ 6.81
    & 66.63 $\pm$ 6.48 
    & 47.67 $\pm$ 10.92 
    & 69.09 $\pm$ 4.78 
    & \underline{72.70 $\pm$ 12.18}
    & 69.15 $\pm$ 20.52 
    & \underline{73.65 $\pm$ 14.38}
    & 67.38 $\pm$ 17.91 \\
& AUPR $(\uparrow)$ 
    & 73.59 $\pm$ 6.67 
    & 70.13 $\pm$ 11.05 
    & \underline{83.40 $\pm$ 21.45}
    & 82.33 $\pm$ 8.48 
    & 64.57 $\pm$ 5.01 
    & 55.38 $\pm$ 6.25 
    & 77.81 $\pm$ 3.77 
    & \underline{91.74 $\pm$ 4.07}
    & 88.38 $\pm$ 7.98 
    & \underline{91.17 $\pm$ 4.80}
    & 93.67 $\pm$ 3.35 \\
& FPR95 $(\downarrow)$ 
    & 73.57 $\pm$ 5.56 
    & 73.31 $\pm$ 19.33 
    & 68.32 $\pm$ 28.11 
    & 52.69 $\pm$ 11.63 
    & 77.73 $\pm$ 9.29 
    & 91.08 $\pm$ 2.65 
    & 74.05 $\pm$ 9.64 
    & 68.27 $\pm$ 20.82 
    & 65.35 $\pm$ 35.14 
    & 70.66 $\pm$ 24.65 
    & 69.54 $\pm$ 32.75 \\
& ID ACC $(\uparrow)$ 
    & 86.21 $\pm$ 5.48 
    & 79.34 $\pm$ 7.47 
    & 68.18 $\pm$ 0.51 
    & 77.41 $\pm$ 0.21 
    & 78.23 $\pm$ 0.41 
    & 61.85 $\pm$ 2.32 
    & 82.22 $\pm$ 4.18 
    & 83.95 $\pm$ 0.89 
    & 58.52 $\pm$ 5.87 
    & 82.71 $\pm$ 5.81 
    & 82.58 $\pm$ 7.93\\
\midrule
\multirow{4}{*}{\rotatebox{90}{Energy}}
& AUROC $(\uparrow)$ 
    & \underline{81.60 $\pm$ 3.70}
    & \underline{80.11 $\pm$ 9.89 }
    & \textbf{76.96 $\pm$ 15.51 }
    & \underline{87.31 $\pm$ 6.04 }
    & \underline{71.18 $\pm$ 8.84 }
    & \underline{49.51 $\pm$ 6.48 }
    & \underline{73.57 $\pm$ 6.85 }
    & \textbf{74.03 $\pm$ 12.44 }
    & \underline{70.25 $\pm$ 22.36 }
    & \textbf{74.18 $\pm$ 15.15 }
    & 66.93 $\pm$ 18.00 \\
& AUPR $(\uparrow)$ 
    & \underline{83.74 $\pm$ 2.64  }
    & \underline{81.19 $\pm$ 8.64  }
    & \textbf{84.27 $\pm$ 21.89 }
    & \textbf{86.73 $\pm$ 7.28  }
    & \underline{68.03 $\pm$ 7.30  }
    & \underline{55.73 $\pm$ 3.36  }
    & \underline{80.74 $\pm$ 5.43  }
    & \textbf{92.42 $\pm$ 3.88 }
    & \underline{89.15 $\pm$ 9.21 }
    & \textbf{91.68 $\pm$ 5.17} 
    & 93.34 $\pm$ 3.28 \\
& FPR95 $(\downarrow)$ 
    & \underline{59.41 $\pm$ 12.56 }
    & \underline{61.22 $\pm$ 27.24 }
    & \underline{66.28 $\pm$ 31.54}
    & \underline{46.63 $\pm$ 17.56}
    & 71.40 $\pm$ 13.05 
    & \underline{87.85 $\pm$ 2.70 }
    & \underline{66.05 $\pm$ 13.87 }
    & \underline{64.10 $\pm$ 22.10}
    & 64.01 $\pm$ 37.25 
    & \underline{70.43 $\pm$ 25.61}
    & \underline{68.50 $\pm$ 21.34} \\
& ID ACC $(\uparrow)$ 
    & 86.21 $\pm$ 5.48 
    & 79.34 $\pm$ 7.47 
    & 68.18 $\pm$ 0.51 
    & 77.41 $\pm$ 0.21 
    & 78.23 $\pm$ 0.41 
    & 61.85 $\pm$ 2.32 
    & 82.22 $\pm$ 4.18 
    & 83.95 $\pm$ 0.89 
    & 58.52 $\pm$ 5.87 
    & 82.71 $\pm$ 5.81 
    & 82.58 $\pm$ 7.93\\
\midrule
\multicolumn{2}{c}{\textit{w/ Prop}} \\
\midrule
\multirow{4}{*}{\rotatebox{90}{GNNSafe}}
& AUROC $(\uparrow)$ & \textcolor{orange}{88.45 $\pm$ 4.02}
    & \textcolor{orange}{83.29 $\pm$ 7.54} 
    & 37.13 $\pm$ 21.81 
    & 88.70 $\pm$ 4.51 
    & 84.33 $\pm$ 3.46 
    & \textcolor{orange}{57.43 $\pm$ 26.51 }
    & \textcolor{orange}{85.51 $\pm$ 10.39 }
    & \textcolor{orange}{57.33 $\pm$ 24.26 }
    & \textcolor{orange}{67.06 $\pm$ 20.56 }
    & \textcolor{orange}{57.38 $\pm$ 14.85 }
    &59.95 $\pm$ 2.55 \\
& AUPR $(\uparrow)$ & \textcolor{orange}{87.77 $\pm$ 5.69}
    & \textcolor{orange}{79.63 $\pm$ 7.77}
    & 66.30 $\pm$ 13.65 
    & 81.71 $\pm$ 3.37
    & 80.48 $\pm$ 4.54 
    & \textcolor{orange}{65.51 $\pm$ 14.82 }
    & \textcolor{orange}{87.49 $\pm$ 6.28 }
    & \textcolor{orange}{84.72 $\pm$ 8.22 }
    & \textcolor{orange}{85.37 $\pm$ 6.40 }
    & 83.08 $\pm$ 5.25 
    & 89.56 $\pm$ 2.69 \\
& FPR95 $(\downarrow)$ & \textcolor{orange}{36.06 $\pm$ 7.21}
    & \textcolor{orange}{43.90 $\pm$ 19.75}
    & 88.51 $\pm$ 8.10 
    & 30.46 $\pm$ 16.36 
    & 39.19 $\pm$ 7.12 
    & \textcolor{orange}{68.64 $\pm$ 15.86 }
    & 38.73 $\pm$ 24.16 
    & \textcolor{orange}{65.47 $\pm$ 24.12 }
    & \textcolor{orange}{55.16 $\pm$ 34.01 }
    & \textcolor{orange}{66.26 $\pm$ 16.76 }
    & 64.41 $\pm$ 18.20 \\
& ID ACC $(\uparrow)$ & 86.21 $\pm$ 5.48 & 79.34 $\pm$ 7.47 & 68.18 $\pm$ 0.51 & 77.41 $\pm$ 0.21 & 78.23 $\pm$ 0.41 & 61.85 $\pm$ 2.32 & 82.22 $\pm$ 4.18 & 83.95 $\pm$ 0.89 & 58.52 $\pm$ 5.87 & 82.71 $\pm$ 5.81 & 82.58 $\pm$ 7.93 \\
\midrule
\multirow{4}{*}{\rotatebox{90}{NODESafe}}
& AUROC $(\uparrow)$ & 86.69 $\pm$ 6.19 
        & 82.09 $\pm$ 6.22 
        & \textcolor{orange}{38.06 $\pm$ 22.48}
        & \textcolor{RoyalBlue}{89.81 $\pm$ 3.55}
        & \textcolor{orange}{84.66 $\pm$ 3.06 }
        & 54.08 $\pm$ 32.33 
        & 80.54 $\pm$ 24.84 
        & 53.31 $\pm$ 23.37
        & 57.35 $\pm$ 12.69 
        & 57.24 $\pm$ 9.81
        & \textcolor{orange}{60.87 $\pm$ 4.05} \\
& AUPR $(\uparrow)$  & 86.11 $\pm$ 7.82
        & 77.94 $\pm$ 6.27 
        & \textcolor{orange}{67.54 $\pm$ 12.30}
        & \textcolor{RoyalBlue}{82.59 $\pm$ 2.73}
        & \textcolor{orange}{80.61 $\pm$ 4.21 }
        & 61.95 $\pm$ 20.25 
        & 84.48 $\pm$ 14.03 
        & 83.22 $\pm$ 8.18 
        & 81.55 $\pm$ 4.44 
        & \textcolor{orange}{83.24 $\pm$ 4.40}
        & \textcolor{orange}{90.04 $\pm$ 3.47} \\
& FPR95 $(\downarrow)$ & 38.54 $\pm$ 9.50 
        & 45.32 $\pm$ 17.70 
        & \textcolor{orange}{87.58 $\pm$ 8.75}
        & \textcolor{orange}{27.04 $\pm$ 12.58}
        & \textcolor{orange}{38.85 $\pm$ 6.10 }
        & 77.57 $\pm$ 28.13 
        & \textcolor{orange}{37.75 $\pm$ 34.39 }
        & 73.55 $\pm$ 18.97 
        & 70.34 $\pm$ 21.05 
        & 68.73 $\pm$ 12.63 
        & \textcolor{orange}{60.28 $\pm$ 15.51} \\
& ID ACC $(\uparrow)$ & 84.94 $\pm$ 4.43 
        & 79.30 $\pm$ 6.62 
        & 67.77 $\pm$ 1.87 
        & 76.82 $\pm$ 0.40
        & 78.90 $\pm$ 0.14 
        & 61.69 $\pm$ 0.67 
        & 80.92 $\pm$ 4.00 
        & 83.79 $\pm$ 1.26 
        & 58.21 $\pm$ 5.66 
        & 80.09 $\pm$ 7.99
        & 80.85 $\pm$ 7.96 \\
\midrule
\midrule
\multirow{4}{*}{\rotatebox{90}{\textbf{TNT-OOD}}}
& AUROC $(\uparrow)$ 
        & \textbf{\textcolor{RoyalBlue}{91.29 $\pm$ 4.21}} 
        & \textbf{\textcolor{RoyalBlue}{86.42 $\pm$ 6.53}} 
        & \textcolor{RoyalBlue}{47.65 $\pm$ 13.80}
        & \textbf{\textcolor{orange}{89.59 $\pm$ 3.50}} 
        & \textbf{\textcolor{RoyalBlue}{88.82 $\pm$ 3.22}} 
        & \textbf{\textcolor{RoyalBlue}{70.61 $\pm$ 37.62}} 
        & \textbf{\textcolor{RoyalBlue}{89.88 $\pm$ 8.19}} 
        & \textcolor{RoyalBlue}{72.63 $\pm$ 15.71}
        & \textbf{\textcolor{RoyalBlue}{79.03$\pm$ 5.09}} 
        & \textcolor{RoyalBlue}{70.09 $\pm$ 5.68} 
        & \textbf{\textcolor{RoyalBlue}{69.67 $\pm$ 6.93}} \\
& AUPR $(\uparrow)$ 
        & \textbf{\textcolor{RoyalBlue}{89.07$\pm$ 5.13}} 
        & \textbf{\textcolor{RoyalBlue}{82.48 $\pm$ 6.93}} 
        & \textcolor{RoyalBlue}{69.60 $\pm$ 13.22}
        & \textcolor{orange}{82.12 $\pm$ 3.22}
        & \textbf{\textcolor{RoyalBlue}{86.67 $\pm$ 3.98}} 
        & \textbf{\textcolor{RoyalBlue}{77.78 $\pm$ 24.26}} 
        & \textbf{\textcolor{RoyalBlue}{89.69 $\pm$ 5.10}} 
        & \textcolor{RoyalBlue}{89.47 $\pm$ 5.75}
        & \textbf{\textcolor{RoyalBlue}{89.27 $\pm$ 3.82}} 
        & \textcolor{RoyalBlue}{88.86 $\pm$ 4.96} 
        & \textcolor{RoyalBlue}{92.72 $\pm$ 4.43} \\
& FPR95 $(\downarrow)$ 
        & \textbf{\textcolor{RoyalBlue}{25.71 $\pm$ 11.89}} 
        & \textbf{\textcolor{RoyalBlue}{36.24 $\pm$ 18.76}} 
        & \textcolor{RoyalBlue}{78.61 $\pm$ 10.67} 
        & \textbf{\textcolor{RoyalBlue}{25.59 $\pm$ 9.54}} 
        & \textbf{\textcolor{RoyalBlue}{33.13 $\pm$ 7.47}} 
        & \textbf{\textcolor{RoyalBlue}{63.28 $\pm$ 36.20}} 
        & \textbf{\textcolor{RoyalBlue}{28.47 $\pm$ 23.60}} 
        & \textbf{\textcolor{RoyalBlue}{42.57 $\pm$ 24.62}} 
        & \textbf{\textcolor{RoyalBlue}{34.98$\pm$ 17.38}} 
        & \textbf{\textcolor{RoyalBlue}{45.58 $\pm$ 7.76}} 
        & \textbf{\textcolor{RoyalBlue}{47.76 $\pm$ 6.62}}
\\
& ID ACC $(\uparrow)$ & 86.32 $\pm$ 5.34 & 79.05 $\pm$ 7.73 & 68.62 $\pm$ 1.32 & 77.53 $\pm$ 0.16 & 78.85 $\pm$ 0.23 & 61.78 $\pm$ 0.16 & 82.68 $\pm$ 4.20 & 85.20 $\pm$ 1.89 & 58.93 $\pm$ 6.34 & 87.50 $\pm$ 5.69 & 88.31 $\pm$ 5.03\\
\bottomrule
\end{tabular}
\vspace{-0.3cm}
}
\caption{Overall performance of OOD detection on TextTopoOOD scenarios.
Results are reported as the \textbf{aggregated mean and standard deviation of the different OOD scenarios}, over \textbf{three runs}. The large variance in OOD detection indicates the different levels of challenging OOD scenarios from TextTopoOOD. The detection results of our TNT-OOD against \textbf{with(/without)-score propagation methods} %
are highlighted by \textcolor{RoyalBlue}{Best} and \textcolor{orange}{Runner-up} (\textbf{Best} and \underline{Runner-up}), respectively.\textbf{ Full results of each OOD scenario on each datasets is in Appendix~\ref{Appendix:extended_experiments}}.}
\label{table:overall_performance}
\vspace{-0.5cm}
\end{table*}

\subsection{Contrastive Objective and Classification }

To align text and structure-aware embeddings in a shared space, we use a symmetric contrastive loss:
\begin{equation}
\hat{\mathbf{P}}_t = \mathbf{P}_t / \|\mathbf{P}_t\|, \hat{\mathbf{g}} = \tilde{\mathbf{g}} / \|\tilde{\mathbf{g}}\|
\end{equation}
\begin{equation}
\mathcal{L}_{\text{cont}} = \frac{1}{2} \left( \text{CE}(\hat{\mathbf{P}}_t \hat{\mathbf{g}}^\top / \tau, \mathbf{I}) + \text{CE}(\hat{\mathbf{g}} \hat{\mathbf{P}}_t^\top / \tau, \mathbf{I}) \right),
\label{eq:contrastive_loss}
\end{equation}
where $\hat{\mathbf{P}}_t, \hat{\mathbf{g}} \in \mathbb{R}^{n \times d_p}$, $\tau$ is the temperature and $\mathbf{I}$ is the identity target. For ID classification, we include an additional classification layer using $\tilde{\mathbf{g}}$:
\begin{equation}
\mathcal{L}_{\text{cls}} = \text{CE}(\text{GCN}_{\text{cls}}(\tilde{\mathbf{g}}_i), y_i).
\end{equation}
The \textbf{final training objective} combines both losses, with a hyperparameter $\lambda$:
\begin{equation}
\mathcal{L}_{\text{TNT-OOD}} = \mathcal{L}_{\text{cls}} + \lambda \mathcal{L}_{\text{cont}}.
\label{eq:final_loss}
\end{equation}

\subsection{OOD Scoring Function}
At test time, we compute OOD scores as a combination of energy (Eq.~\ref{eq:energy}) and alignment score (Eq.~\ref{eq:alignment}):
\begin{equation}
\mathbf{s}_{\text{align}} = \left\langle \hat{\mathbf{P}}_t, \hat{\mathbf{g}} \right\rangle,
\label{eq:alignment}
\end{equation}
\begin{equation}
\mathbf{s}_{\text{E-lign}} = \text{$\mathbf{e}$}_{\text{energy}} - T \cdot \mathbf{s}_{\text{align}},
\end{equation}
$T$ is the temperature. Notably, energy identifies nodes with low confidence across all classes at the logits level, while alignment detects inconsistencies in text-graph relationships that energy alone might miss - a key motivation of TNT-OOD. Higher score indicates more OOD the node is (i.e., high energy and low alignment). The scores are further refined via a $K$-layer \textbf{\textit{propagation} smoothing} as in Eq.~\ref{eq:energy_prop}:
\begin{equation}
\tilde{\mathbf{s}}^{(k)} = \alpha_{\text{score}} \tilde{\mathbf{s}}^{(k-1)} + (1 - \alpha_{\text{score}}) \mathbf{D}^{-1} \mathbf{A} \tilde{\mathbf{s}}^{(k-1)},
\end{equation}
where $\tilde{\mathbf{s}}^{(0)} = \mathbf{s}_{\text{E-lign}}$, $\alpha_{\text{score}}$ controls concentration, and $\mathbf{D}^{-1} \mathbf{A}$ is the normalised graph Laplacian.

\section{Experiments} \label{sec:experiments}

\paragraph{Datasets.}
\label{sec:datasets}
TextTopoOOD and TNT-OOD evaluate OOD scenarios across 11 TrNs with varying scales, structural properties, and domains - including citation, knowledge, social, and e-commerce networks~\cite{TSGFM}. For each dataset, \textbf{OOD shifts are generated based on selected TextTopoOOD scenarios at multiple noise levels or modes}. Due to space constraint, the detailed discussion and OOD construction is in Appendix~\ref{Appendix:dataset_description}.

\paragraph{Baselines.} We compare TNT-OOD with 7 baselines, including (1) \textbf{post-hoc methods}:
Mahalanobis (Maha)~\citep{Mahalanobis}, MSP~\citep{MSP}, ODIN~\citep{ODIN}, NECO~\citep{ODIN}, and Energy~\citep{energy}; (2) \textbf{graph-specific OOD detectors} that leverage \textbf{propagation schema}: \textsc{GNNSafe}~\citep{GNNSafe}, and \textsc{NODESafe}~\citep{NODESAFE}; (3) \textbf{LLM} zero shot detection on label shift was conducted with GPT-4o mini~\cite{gpt4omini} and Gemini-2.5-flash~\cite{gemini2.5}.
\paragraph{Metrics.}
Following prior work, \textbf{AUROC} ($\uparrow$) $/$ \textbf{AUPR} ($\uparrow$) $/$ \textbf{FPR95} ($\downarrow$) was utilised to measure OOD detection, with Accuracy used for ID classification~\cite{GNNSafe, NODESAFE}. Appendix~\ref{Appendix:Metrics} provides further details on the metrics.

\begin{table*}[t!]
\centering
\resizebox{\textwidth}{!}{%
\begin{tabular}{cccc|cccc|cccc|cccc}
\toprule
\multirow{2}{*}{Cr.Attn.} & \multirow{2}{*}{HyperN.} & \multirow{2}{*}{$\mathcal{L}_{\text{cont.}}$} & \multirow{2}{*}{$\mathbf{s}_{\text{align}}$} & \multicolumn{4}{c|}{Cora} & \multicolumn{4}{c|}{Citeseer} & \multicolumn{4}{c}{Elephoto} \\
 & &  &  & AUROC$(\uparrow)$ & AUPR$(\uparrow)$ & FPR$(\downarrow)$ & ID Acc$(\uparrow)$ & AUROC$(\uparrow)$ & AUPR$(\uparrow)$ & FPR$(\downarrow)$ & ID Acc$(\uparrow)$ & AUROC$(\uparrow)$ & AUPR$(\uparrow)$ & FPR$(\downarrow)$ & ID Acc$(\uparrow)$ \\
\midrule
 \multicolumn{4}{c|}{GNNSafe}  & 89.19 & 89.59 & 34.21 & 82.64 &  85.20 & 82.12  & 40.03  & 74.06 & 56.14 & 82.62 &  65.14 & 78.60 \\
 \midrule
 \checkmark & &  & &       88.24  & 87.34 & 33.68 & 82.21 & 86.25 & 82.78 & 36.03 & 74.60 & 61.41 & 86.02 & 64.17 & 81.73  \\
\checkmark &  & \checkmark & \checkmark &         \textcolor{RoyalBlue}{\textbf{91.65}}   &  \textcolor{RoyalBlue}{\textbf{92.46}} & 27.64& 82.45&  \textcolor{RoyalBlue}{\textbf{88.56}} &  \textcolor{RoyalBlue}{\textbf{85.54}} & \textcolor{orange}{\textbf{33.63}} & 73.68 & 60.93 & 85.79 & 65.50 & 81.85 \\
\checkmark & \checkmark & & &    90.82  & 89.57 & 27.56 & 81.72 & 83.42 & 80.00 &  39.24 & 73.68 & 66.89 & 87.44 & 51.20 & 83.14 \\
\checkmark & \checkmark & \checkmark & & 91.35  & 89.93&   \textcolor{orange}{\textbf{25.49}}   &  82.54 & 84.74 & 81.14 &  36.80 &  74.11 & \textcolor{orange}{\textbf{67.18}} & \textcolor{orange}{\textbf{87.24}}  &  \textcolor{orange}{\textbf{49.79}} &  83.48 \\
\midrule
\multicolumn{4}{c|}{TNT-OOD} & \textcolor{RoyalBlue}{\textbf{91.65}} &  \textcolor{orange}{\textbf{90.55}} &  \textcolor{RoyalBlue}{\textbf{24.91}} & 82.54 &  \textcolor{orange}{\textbf{87.96}} &  \textcolor{orange}{\textbf{84.65}} &  \textcolor{RoyalBlue}{\textbf{32.28}} & 74.11 &  \textcolor{RoyalBlue}{\textbf{68.69}} &  \textcolor{RoyalBlue}{\textbf{88.05}}  &  \textcolor{RoyalBlue}{\textbf{47.40}} & 83.48  \\
\bottomrule
\end{tabular}%
}
\caption{Ablation study. \textbf{Cr.Attn.} and \textbf{HyperN.} denotes the cross attention and  HyperNetwork module, respectively.}
\label{Table:ablation_study}
\end{table*}

\paragraph{Implementation.}
All-MiniLM-L6-v2~\cite{minilm} is used to encode text embeddings. All methods' configurations are determined based on the ID classification performance via grid search. For fair comparison, post-hoc baselines used the same classifier configuration, resulting in the \textbf{same ID accuracy}.  For TNT-OOD, the HyperNetwork is implemented as a two-layer MLP and a rank $r$ of 16. A single GCN layer is used in each component. The projection sizes was set to 128. The number of score propagation $K$ is set to 3 for all relevant methods, with $\alpha_{\text{score}} = 0.5$. Full hyperparameter search and sensitivity analysis are in Appendix~\ref{Appendix:implementation_details}.

\subsection{Overall Performance}
\textbf{TNT-OOD consistently outperforms or matches baseline methods across diverse network domains and OOD scenarios.} Table~\ref{table:overall_performance} shows TNT-OOD's superior performance with significant improvements in key metrics across multiple network types. 
For citation networks (Cora, Citeseer, Pubmed), TNT-OOD surpasses all baselines, reducing average FPR95 by up to 10\%. In knowledge and social networks (Reddit, WikiCS), it delivers substantial gains in AUROC and AUPR (increasing average AUROC from 57.43\% to 70.61\% on Reddit). For e-commerce networks, TNT-OOD achieves the lowest FPR95 scores across all datasets. While maintaining competitive ID accuracy, TNT-OOD excels in OOD detection, demonstrating its effectiveness in diverse TrNs. However, TextTopoOOD also \textbf{reveals challenging scenarios} (e.g., Arxiv) where TNT-OOD and propagation-based methods underperform, validating the framework's efficacy in creating rigorous OOD cases. Full results is in Appendix~\ref{Appendix:extended_experiments}. Regarding computational resources, due to the computation of the cross attention and HyperNetwork weight generation module, TNT-OOD has inevitable come at a higher cost in memory usage and computation time as discussed in Appendix~\ref{Appendix:computational_cost}. However, as an initial work on TrN OOD detection, we believe the superior performance of TNT-OOD and the challenging nature of TextTopoOOD paves the way for more efficient and effective methods in the future.

\subsection{Extended Analysis on Score Propagation}
\textbf{Our framework reveals the nuanced impact of score propagation across different text-rich networks, demonstrating TextTopoOOD's challenging nature.} Table~\ref{Table:propagation_experiments} shows that in Cora, propagation (prop.) methods substantially enhance performance, with TNT-OOD achieving superior results (25.71\% FPR95). Conversely, for Arxiv, prop. degrades performance, with our prop-free variant TNT-wo outperforming all prop. approaches (53.13\% FPR95). Bookhis presents a hybrid case where TNT-wo excels in AUROC (88.53\%) and FPR95 (38.82\%), while TNT-OOD still outperforms prop. baselines. These findings demonstrate our method's adaptability across diverse network characteristics. As propagation is usually unavailable for test-time optimisation, we advocate for future research to enhance OOD detection in TextTopoOOD's challenging scenarios.

\begin{table}[h!]
    \centering
    \resizebox{1\linewidth}{!}{
    \begin{tabular}{c|c|cc|cc}
    \specialrule{.1em}{.05em}{.05em}
    \specialrule{.1em}{.05em}{.05em}
    & \multirow{2}{*}{\textbf{Metrics}}&\multicolumn{2}{c|}{w/o Prop.}&\multicolumn{2}{c}{w/ Prop.}\\
    &  & Energy & TNT-wo & GNNSafe & TNT-OOD \\
    \midrule
    \multirow{3}{*}{Cora}
    & AUROC $(\uparrow)$ & 81.60 & 86.24 & \textcolor{Orange}{\textbf{88.45}} & \textcolor{RoyalBlue}{\textbf{91.29}}  \\
    &AUPR $(\uparrow)$  &  83.74 & 86.57 & \textcolor{Orange}{\textbf{87.77}} & \textcolor{RoyalBlue}{\textbf{89.07}} \\
    &FPR95 $(\downarrow)$ & 59.41 & 44.67 & \textcolor{Orange}{\textbf{36.06}} & \textcolor{RoyalBlue}{\textbf{25.71}} \\
    \midrule
    \multirow{3}{*}{Arxiv}
    & AUROC $(\uparrow)$ & \textcolor{Orange}{\textbf{76.96}} & \textcolor{RoyalBlue}{\textbf{77.99}} & 37.13 & 47.65  \\
    &AUPR $(\uparrow)$  & \textcolor{RoyalBlue}{\textbf{84.27}} & \textcolor{Orange}{\textbf{84.11}} &  66.30 & 69.60 \\
    &FPR95 $(\downarrow)$  & \textcolor{Orange}{\textbf{66.28}} & \textcolor{RoyalBlue}{\textbf{53.13}} & 88.51 & 78.61 \\
    \midrule
    \multirow{3}{*}{Bookhis}
    & AUROC $(\uparrow)$ & 74.03 & \textcolor{RoyalBlue}{\textbf{88.53}} & 57.33  & \textcolor{Orange}{\textbf{75.39}} \\
    &AUPR $(\uparrow)$  &  \textcolor{Orange}{\textbf{92.42}} & \textcolor{RoyalBlue}{\textbf{96.89}} & 84.72 & 90.06 \\
    &FPR95 $(\downarrow)$  & 64.10 & \textcolor{RoyalBlue}{\textbf{38.82}} &  65.47 & \textcolor{Orange}{\textbf{40.13}}  \\
    \specialrule{.1em}{.05em}{.05em}
    \specialrule{.1em}{.05em}{.05em}
    \end{tabular}}
    \caption{Detection comparison between \textbf{with out (w/o)} and \textbf{with (w/)} score propagation. We refer the TNT-OOD method without Prop. as \textbf{TNT-wo}.}
    \label{Table:propagation_experiments}
    \vspace{-0.5cm}
\end{table}

\begin{figure*}[t!]
   \centering
   
   \begin{minipage}[c]{0.02\textwidth}
       \centering
       \rotatebox{90}{\scriptsize\textbf{TNT-OOD}}
   \end{minipage}%
   \begin{minipage}[c]{0.95\textwidth}
       \begin{subfigure}[b]{0.155\textwidth}
           \centering
           \includegraphics[width=\textwidth]{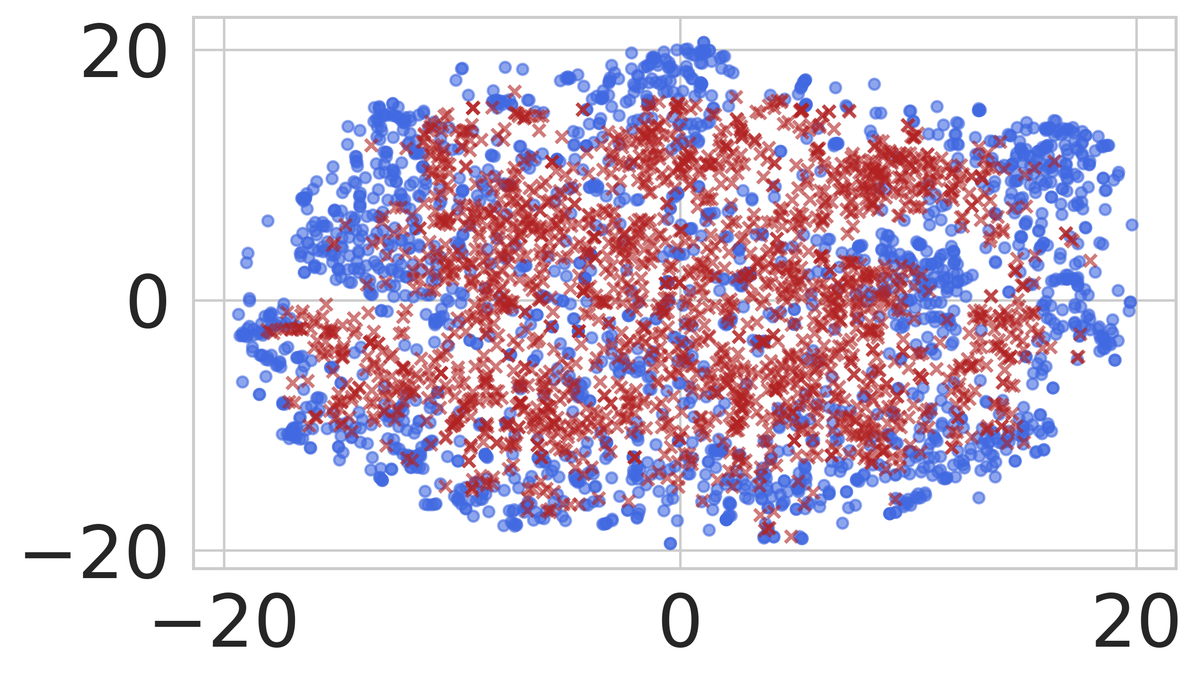}
       \end{subfigure}
       \hfill
       \begin{subfigure}[b]{0.155\textwidth}
           \centering
           \includegraphics[width=\textwidth]{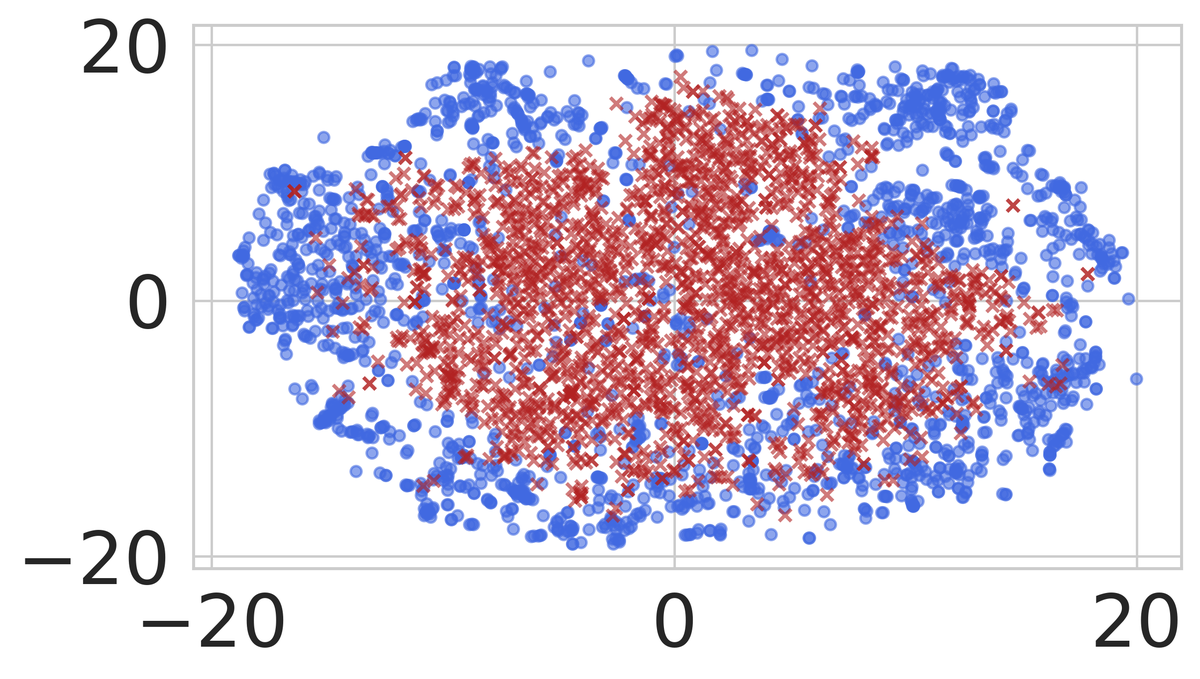}
       \end{subfigure}
       \hfill
       \begin{subfigure}[b]{0.155\textwidth}
           \centering
           \includegraphics[width=\textwidth]{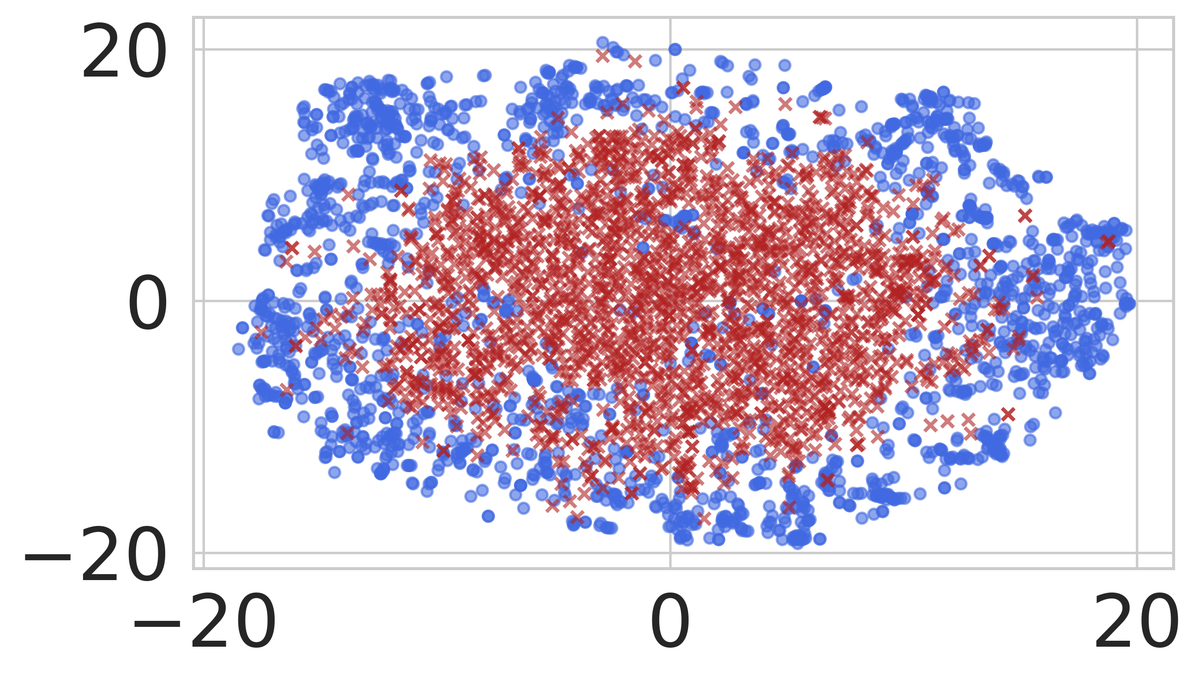}
       \end{subfigure}
       \hfill
       \begin{subfigure}[b]{0.155\textwidth}
           \centering
           \includegraphics[width=\textwidth]{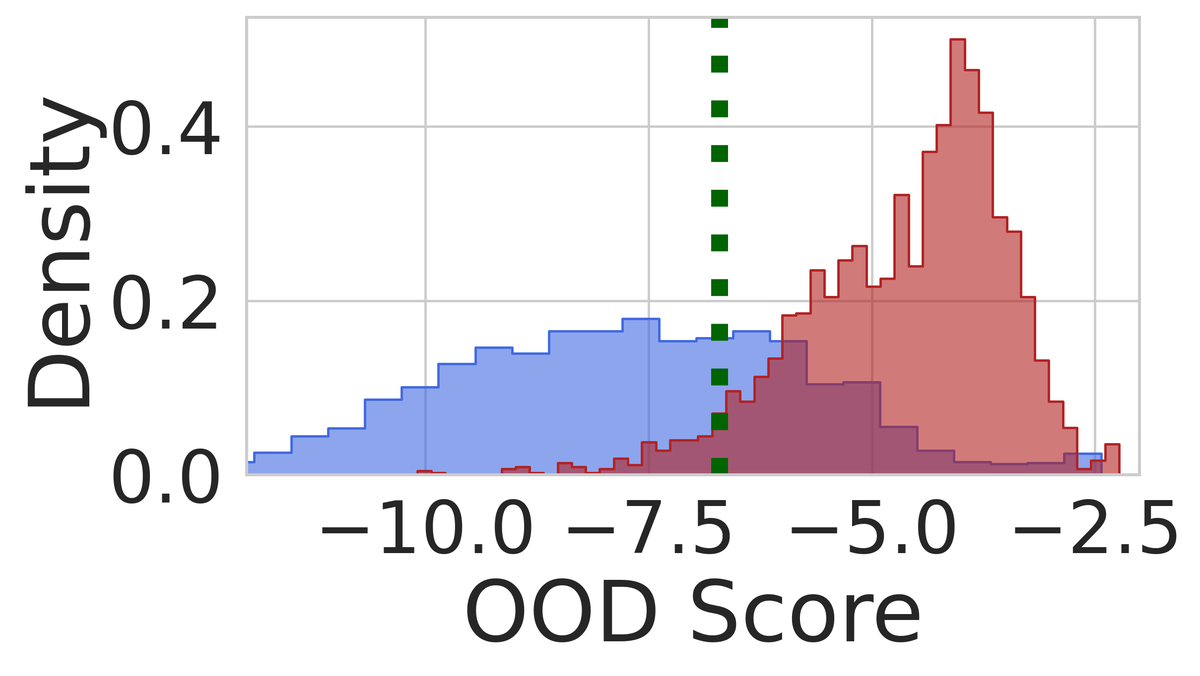}
       \end{subfigure}
       \hfill
       \begin{subfigure}[b]{0.155\textwidth}
           \centering
           \includegraphics[width=\textwidth]{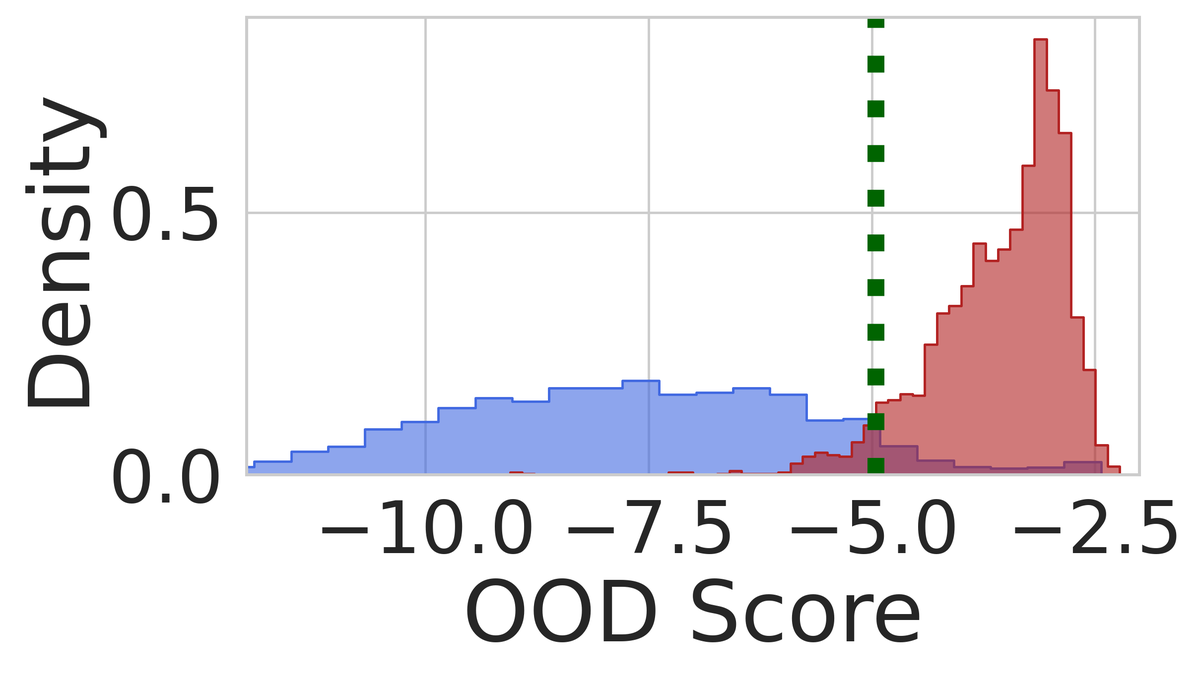}
       \end{subfigure}
       \hfill
       \begin{subfigure}[b]{0.155\textwidth}
           \centering
           \includegraphics[width=\textwidth]{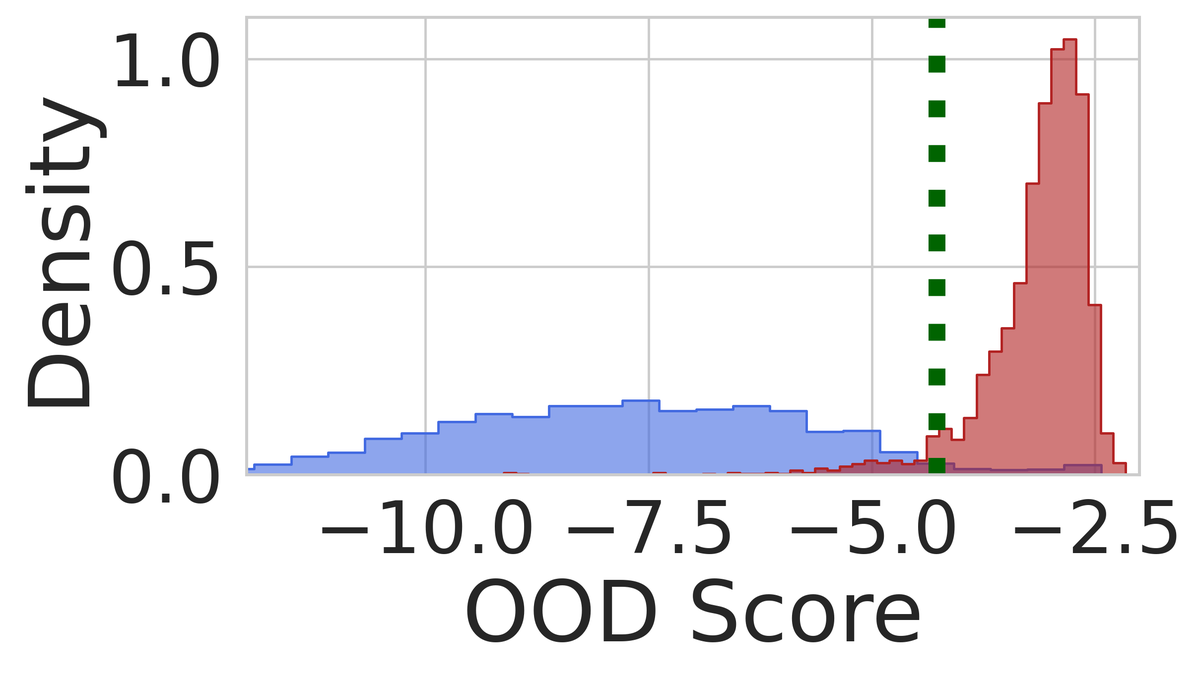}
       \end{subfigure}
   \end{minipage}
   
   \vspace{0.3cm}
   
   \begin{minipage}[c]{0.02\textwidth}
       \centering
       \rotatebox{90}{\scriptsize\textbf{GNNSafe}}
   \end{minipage}%
   \begin{minipage}[c]{0.95\textwidth}
       \begin{subfigure}[b]{0.155\textwidth}
           \centering
           \includegraphics[width=\textwidth]{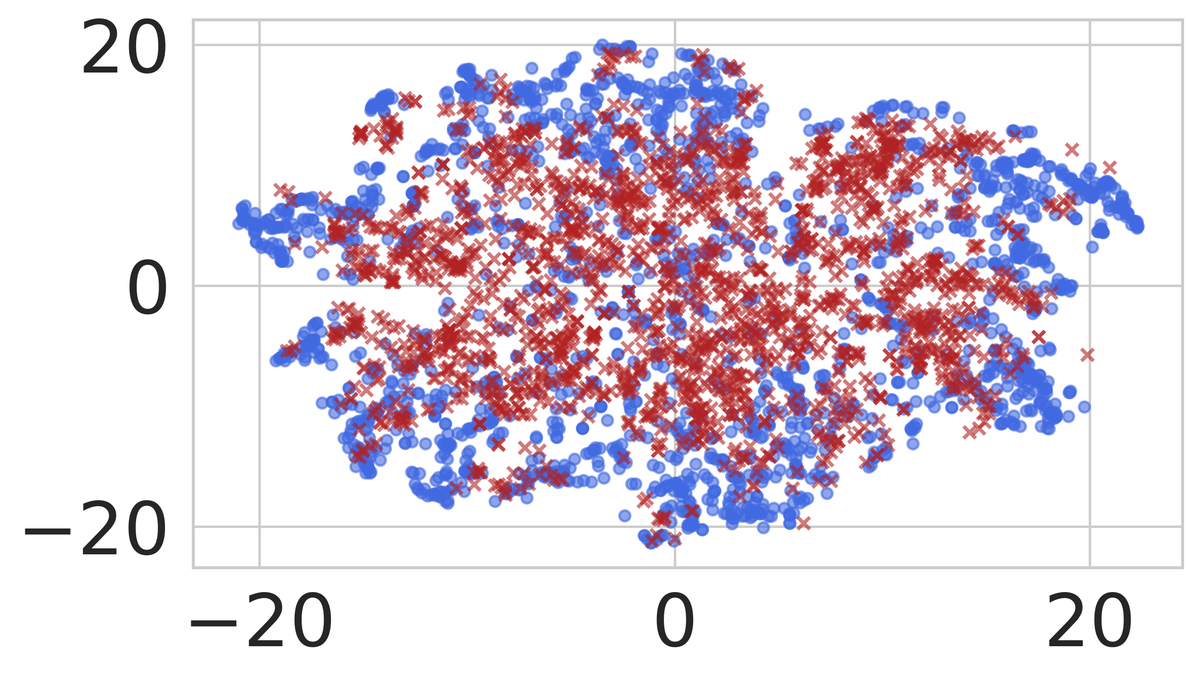}
           \caption{Emb. $\alpha=0.5$}
       \end{subfigure}
       \hfill
       \begin{subfigure}[b]{0.155\textwidth}
           \centering
           \includegraphics[width=\textwidth]{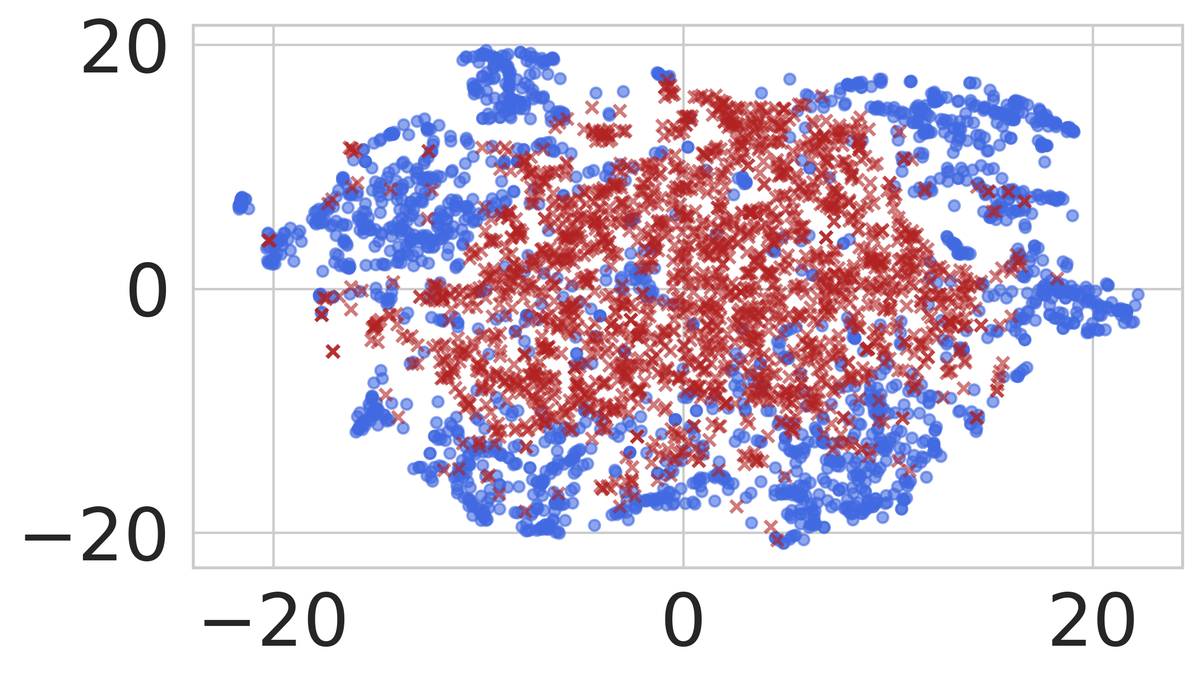}
           \caption{$\alpha=0.7$}
       \end{subfigure}
       \hfill
       \begin{subfigure}[b]{0.155\textwidth}
           \centering
           \includegraphics[width=\textwidth]{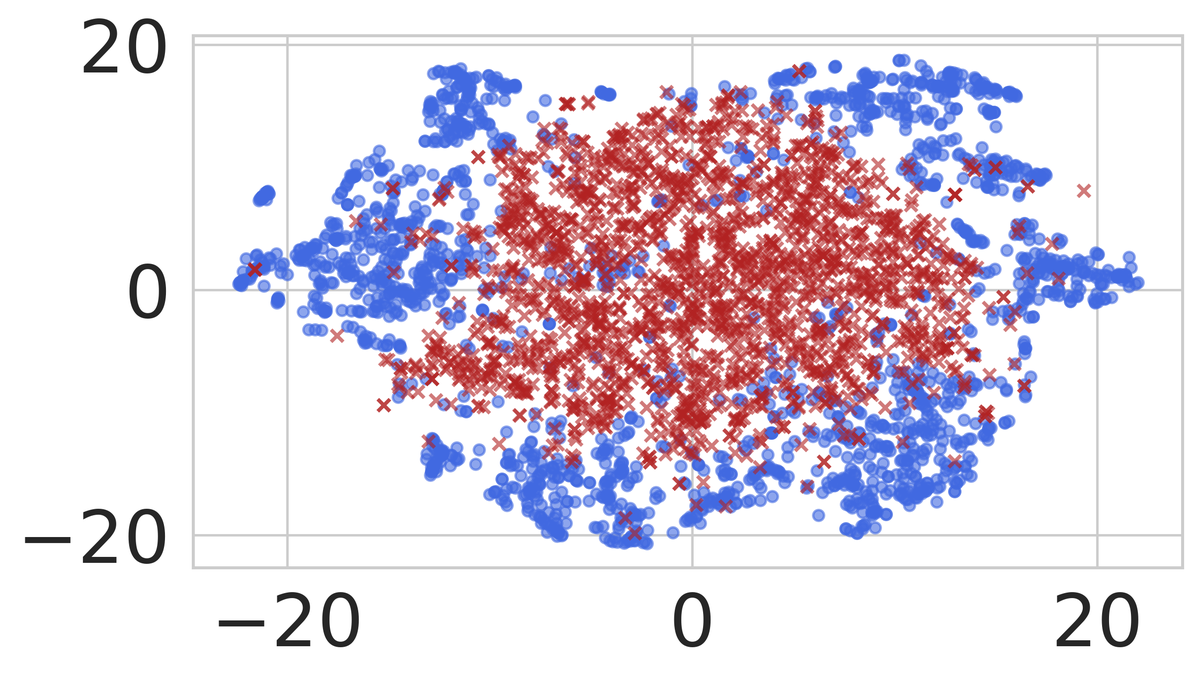}
           \caption{$\alpha=0.9$}
       \end{subfigure}
       \hfill
       \begin{subfigure}[b]{0.155\textwidth}
           \centering
           \includegraphics[width=\textwidth]{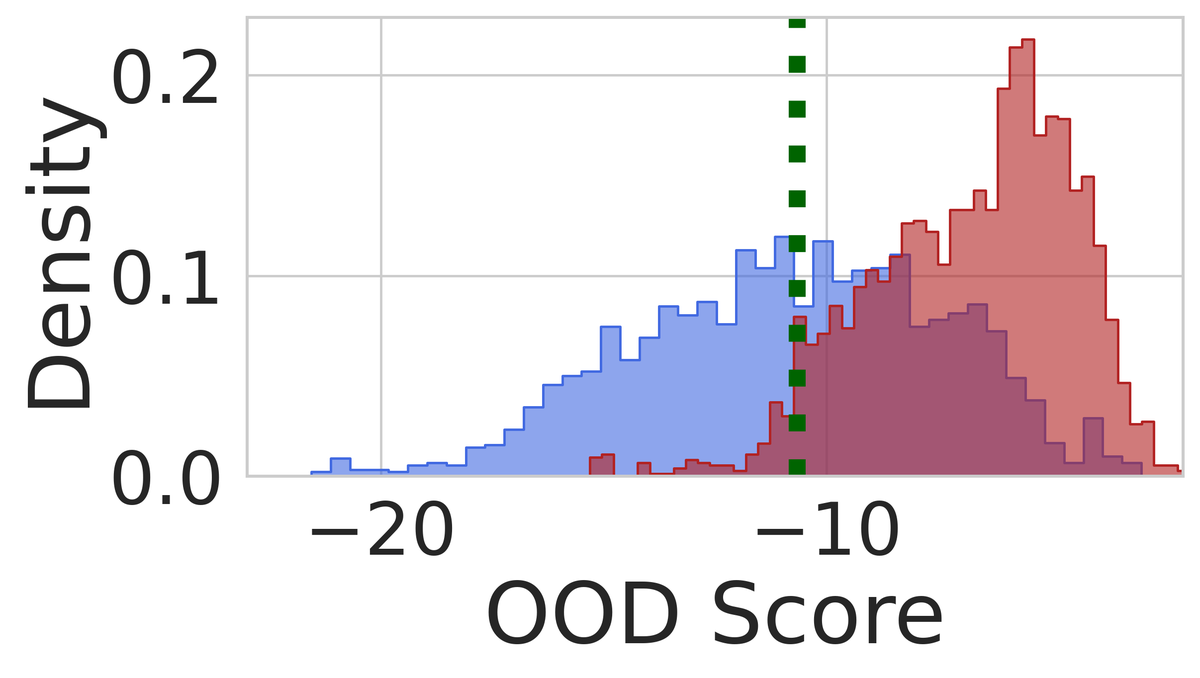}
           \caption{Score $\alpha=0.5$}
       \end{subfigure}
       \hfill
       \begin{subfigure}[b]{0.155\textwidth}
           \centering
           \includegraphics[width=\textwidth]{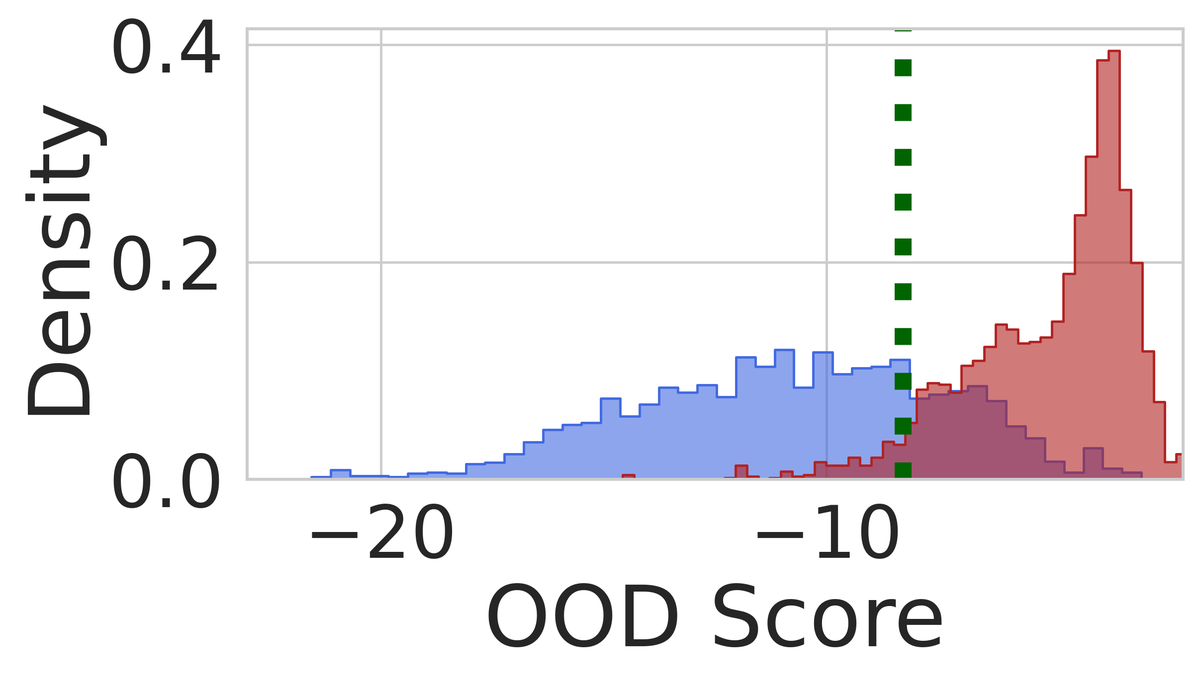}
           \caption{$\alpha=0.7$}
       \end{subfigure}
       \hfill
       \begin{subfigure}[b]{0.155\textwidth}
           \centering
           \includegraphics[width=\textwidth]{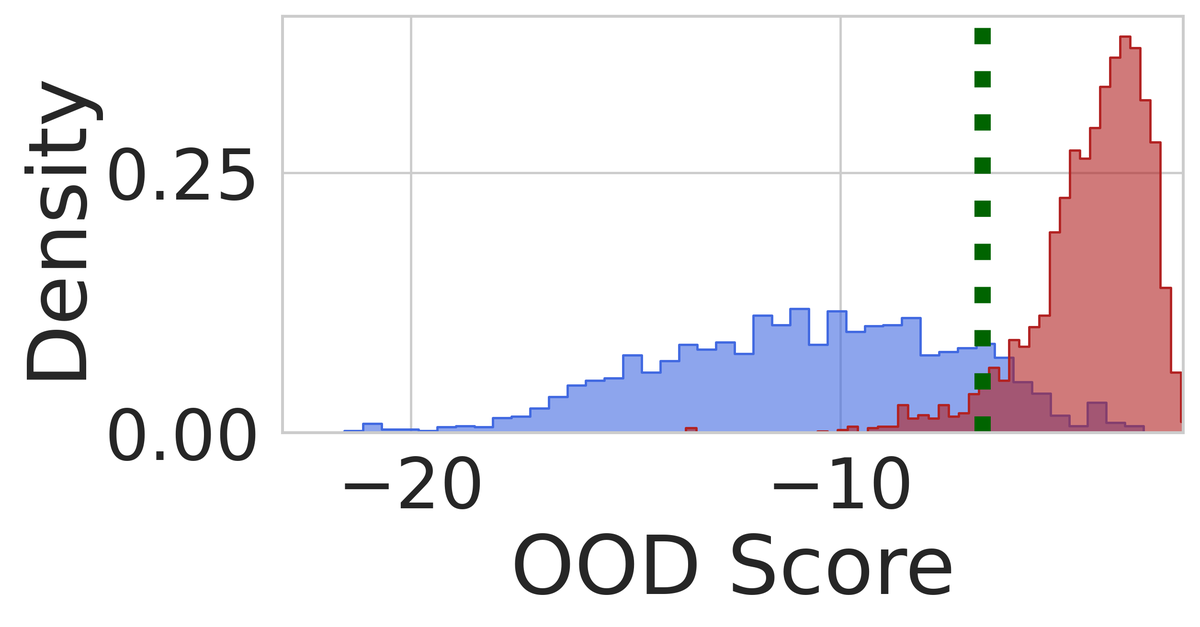}
           \caption{$\alpha=0.9$}
       \end{subfigure}
   \end{minipage}
   
   \caption{Embedding representation and OOD score visualisation of \textbf{\textcolor{RoyalBlue}{ID}} and \textbf{\textcolor{Maroon}{OOD}} data between \textbf{TNT-OOD (Top)} and \textbf{GNNSafe (Bottom)} at different noise levels on Cora Feature. \textcolor{ForestGreen}{Dash line} indicates FPR95 threshold.}
   \label{fig:score_distribution_viz}
\end{figure*}

\subsection{Ablation Study}
Table~\ref{Table:ablation_study} presents TNT-OOD's ablation study, with GNNSafe as baseline. The cross-attention mechanism alone improves OOD detection, increasing AUROC from 56.14\% to 61.41\% on Elephoto. Adding contrastive loss and alignment score delivers strong results on Cora and exceptional performance on Citeseer (88.56\% AUROC). While the HyperNetwork shows mixed results across datasets, it greatly improves Elephoto's FPR95 from 64.17\% to 51.20\%. When combined with the contrastive objective that aligns text and structure-aware embeddings, it shows consistent improvements. The full TNT-OOD achieves superior performance across metrics and datasets, confirming our components' efficacy for OOD detection in TrNs. Further experiment on the text encoder choice is in Appendix~\ref{Appendix:extended_experiments}.

\subsection{Comparison with LLM as Detector}

We further evaluate LLMs as OOD detectors using only ID categories. Since LLMs cannot produce soft OOD scores, we compare TNT-OOD with FPR calculated at equivalent TPR levels. Table~\ref{table:llm_as_detector} shows TNT-OOD outperforms GPT-4o mini on Cora (reducing FPR from 34.89\% to 25.29\%) and Gemini-2.5-flash on Wikics (reducing FPR by 10\%). %
The prompt is provided in Appendix Figure~\ref{prompt:llm_prompt_detector}.

\begin{table}[t!]
\centering
\resizebox{0.5\textwidth}{!}{%
\begin{tabular}{cc|c|c}
\toprule
 & & {Cora - Label} & Wikics - Label \\
\midrule
{\textbf{GPT-4o mini}}
 & FPR $(\downarrow)$ & 34.89 & 51.28  \\
\textbf{TNT-OOD} & FPR@ 4o mini TPR $(\downarrow)$  & \textbf{25.29} & \textbf{30.43}  \\
\midrule
{\textbf{Gemini-2.5-flash}} & FPR $(\downarrow)$ & 28.58 & 34.33  \\
\textbf{TNT-OOD} & FPR@ Gemini TPR $(\downarrow)$ & \textbf{17.30} & \textbf{24.41}  \\
\bottomrule
\end{tabular}%
}
\caption{Results of LLM as detector on Label shifts.}
\label{table:llm_as_detector}
\vspace{-0.3cm}
\end{table}

\subsection{OOD Score Visualisation}
Figure~\ref{fig:score_distribution_viz} visualises the learned embedding representations and corresponding OOD score distributions across varying noise levels on the \textbf{Cora Feature} dataset. The \textbf{top row displays results from TNT-OOD}, while the \textbf{bottom row shows the GNNSafe} baseline performance. As expected, increasing the noise level ($\alpha_{\text{feat}}$ from 0.5 to 0.9) generally reduces the difficulty in distinguishing OOD data, reflected in clearer separations between the \textcolor{RoyalBlue}{ID} and \textcolor{Maroon}{OOD} samples. The \textcolor{ForestGreen}{green dashed lines} mark the threshold corresponding to an FPR at 95\% TPR. Notably, our TNT-OOD framework consistently achieves more pronounced separation between \textcolor{RoyalBlue}{ID} and \textcolor{Maroon}{OOD} scores, evidenced by a more \textbf{distinct gap between distributions} (i.e., the \textcolor{RoyalBlue}{ID} data skewed greater to the left of the \textcolor{ForestGreen}{threshold}. In comparison, GNNSafe shows greater overlap in \textcolor{RoyalBlue}{ID} and \textcolor{Maroon}{OOD} scores across all noise levels, highlighting its relatively weaker discriminative capability.

\subsection{Extended Analysis on Coupled Shifts}

To further capture the coupled nature of real-world shifts, we conduct an additional experiment where feature mixing is superimposed on top of structure rewiring, as shown in Table~\ref{table:coupled_shifts}. A key observation is that these combined shifts amplify representation misalignment, thereby making the discrepancy between ID and OOD nodes more pronounced and ultimately improving detection performance.

\begin{table}[h]
\centering
\resizebox{0.5\textwidth}{!}{
\begin{tabular}{lccc}
\hline
\textbf{Citeseer} & \textbf{AUROC$\uparrow$} & \textbf{AUPR$\uparrow$} & \textbf{FPR95$\downarrow$} \\
\hline
\multicolumn{4}{l}{\textbf{StructureRewiring}} \\
\quad mild   & 71.91 & 69.69 & 73.45 \\
\quad strong & 94.68 & 91.74 & 15.71 \\
\hline
\multicolumn{4}{l}{\textbf{FeatureMixing}} \\
\quad $\alpha_{\text{feat}}=0.5$ & 84.23 & 78.89 & 38.11 \\
\quad $\alpha_{\text{feat}} = 0.7$ & 91.85 & 86.63 & 22.19 \\
\hline
\multicolumn{4}{l}{\textbf{Structure+Feature}} \\
\quad mild+$\alpha_{\text{feat}} = 0.5$   & 94.12 & 89.50 & 16.47 \\
\quad mild+$\alpha_{\text{feat}} = 0.7$   & 96.58 & 91.74 &  6.21 \\
\quad strong+$\alpha_{\text{feat}} = 0.5$ & 96.88 & 95.01 &  6.53 \\
\quad strong+$\alpha_{\text{feat}} = 0.7$ & 96.87 & 95.02 &  6.52 \\
\hline
\end{tabular}
}
\caption{OOD detection performance (Citeseer) under different OOD shifts.}
\label{table:coupled_shifts}    
\vspace{-0.5cm}
\end{table}

\section{Related Work}
\label{sec:related_work}
Our work explores OOD detection in TrNs across three research trajectories. \textbf{1) Post-hoc methods} construct specialised scoring functions using only ID data~\cite{MSP, KNN, Mahalanobis, ODIN, Neco, energy, Flats}, avoiding model retraining. For \textbf{2) graph-structured data}, GNNSafe introduced energy-based scoring with propagation to leverage node-topology interdependencies~\cite{GNNSafe}, while NODESafe refined extreme energy scores~\cite{NODESAFE} and GRASP enhanced edge augmentation~\cite{grasp} - though we exclude comparison with the latter as it uses transductive settings while we focus on inductive tasks. More complex algorithm designs have also been explored, including GOLD and DeGEM~\cite{GOLD, DEGEM}, which presents a data synthesis strategy to simulate OOD scenarios without requiring actual auxiliary OOD samples. \textbf{3) OOD detection in NLP} has made great progress over the year~\cite{NLPOODD}, employing both established post-hoc methods and newer approaches that leverage LLMs' semantic understanding capabilities~\cite{LLMOOD1,llmood_survey, LLMOOD2}. \textbf{4) Text-rich Network OOD detection} remains underexplored, with recent LLM-based approaches focusing mainly on label shifts~\cite{GLIP-OOD, OOD_TAG}, neglecting text-structure interactions. We address this gap with TextTopoOOD, which systematically investigates diverse textual and structural shifts, and TNT-OOD, which models text-structure interplay for improved detection. A detailed related work is in Appendix~\ref{Appendix:extended_related_work}.

\section{Conclusion}
We propose \textbf{TextTopoOOD}, the first comprehensive framework tailored for evaluating OOD detection in text-rich networks across diverse OOD scenarios; this encompasses feature, structural, label, and domain-based shifts. Moreover, to address the unique challenges posed by the interplay of text and topology, we proposed \textbf{TNT-OOD}; a novel detection method that fuses structure-aware context into textual embeddings via a cross-attention mechanism and dynamically adapts to node-level heterogeneity using a HyperNetwork-driven projection module. Extensive experiments on 11 datasets under four distinct OOD scenarios, demonstrates the nuance and challenging nature of TextTopoOOD and the efficacy of TNT-OOD. This work paves the way for more robust and semantically aware models in real-world text rich network applications.

\section*{Acknowledgements}
This work is supported by Australian Research Council DP230101196, DE250100919 and CE200100025.

\newpage
\section*{Limitations}

This study focuses on node-level OOD detection, we hope to expand the research into graph-level OOD detection. Additionally, due to hardware constraints (single NVIDIA RTX A6000 GPU (48GB)), batch-wise and approximation strategies were utilised for similarity calculations and to construct the OOD shifts on larger scale datasets, which may potentially impact the performance. Nevertheless, we have provided extensive experiments across diverse scale and domain datasets, validating the nuance of our TextTopoOOD framework and efficacy of TNT-OOD. Additionally, we have only evaluated a selection of configurations for the proposed TextTopoOOD splits and evaluated on one primary text encoder method, we hope to extend with more experiments as future work. Additionally, in TNT-OOD, the text encoder remain fixed, as text is a key element of TrNs, we leave explorations on fine-tuning the PLMs as future work. Furthermore, with consideration of the textual and structural features, the proposed cross-attention and HyperNetwork architecture inevitably resulted in higher computational cost, both in memory usage and training time as discussed in Section~\ref{sec:experiments}. However, we have made an attempted to reduce cost by implementing an efficient low-rank HyperNetwork and using batch calculations. As an initial work that demonstrates great performance and presenting with challenging and nuanced TrN OOD detection scenarios, we hope this can pave the way for future explorations into effective and efficient OOD detection methods on text-rich networks.

\bibliography{custom}

\begin{thebibliography}{61}
\providecommand{\natexlab}[1]{#1}

\bibitem[{Abbas et~al.(2025)Abbas, Azmat, Horesh, and Yurochkin}]{LLM_OOD_Syn}
Momin Abbas, Muneeza Azmat, Raya Horesh, and Mikhail Yurochkin. 2025.
\newblock \href {https://doi.org/10.48550/ARXIV.2502.03323} {Out-of-distribution detection using synthetic data generation}.
\newblock \emph{CoRR}, abs/2502.03323.

\bibitem[{Abbe(2017)}]{SBM}
Emmanuel Abbe. 2017.
\newblock Community detection and stochastic block models: Recent developments.
\newblock \emph{JMLR}.

\bibitem[{Ammar et~al.(2024)Ammar, Belkhir, Popescu, Manzanera, and Franchi}]{Neco}
Mou{\"{\i}}n~Ben Ammar, Nacim Belkhir, Sebastian Popescu, Antoine Manzanera, and Gianni Franchi. 2024.
\newblock {NECO:} neural collapse based out-of-distribution detection.
\newblock In \emph{ICLR}.

\bibitem[{Anthropic(2025)}]{Claude}
Anthropic. 2025.
\newblock Claude 3.7 sonnet.
\newblock Large language model, available at \url{https://www.anthropic.com/claude}.

\bibitem[{Bao et~al.(2024)Bao, Wu, Jiang, Chen, Sun, and Yan}]{TopoOOD}
Tianyi Bao, Qitian Wu, Zetian Jiang, Yiting Chen, Jiawei Sun, and Junchi Yan. 2024.
\newblock Graph out-of-distribution detection goes neighborhood shaping.
\newblock In \emph{ICML}.

\bibitem[{Cai et~al.(2025)Cai, Jiang, Liu, Li, Huang, and Pan}]{GOODD-survey}
Tingyi Cai, Yunliang Jiang, Yixin Liu, Ming Li, Changqin Huang, and Shirui Pan. 2025.
\newblock Out-of-distribution detection on graphs: {A} survey.
\newblock \emph{CoRR}.

\bibitem[{Chauhan et~al.(2024)Chauhan, Zhou, Lu, Molaei, and Clifton}]{hypernetwork_survey}
Vinod Kumar Chauhan, Jiandong Zhou, Ping Lu, Soheila Molaei, and David A. Clifton. 2024.
\newblock A brief review of hypernetworks in deep learning.
\newblock \emph{Artificial Intelligence Review}.

\bibitem[{Chen et~al.(2022)Chen, Bi, Gao, and Sun}]{BERT_OOD}
Sishuo Chen, Xiaohan Bi, Rundong Gao, and Xu~Sun. 2022.
\newblock Holistic sentence embeddings for better out-of-distribution detection.
\newblock In \emph{EMNLP Findings}.

\bibitem[{Chen et~al.(2025)Chen, Luo, Song, Dai, Tang, and Cao}]{DEGEM}
Yuhan Chen, Yihong Luo, Yifan Song, Pengwen Dai, Jing Tang, and Xiaochun Cao. 2025.
\newblock Decoupled graph energy-based model for node out-of-distribution detection on heterophilic graphs.
\newblock \emph{ICLR}.

\bibitem[{Chen et~al.(2024{\natexlab{a}})Chen, Mao, Liu, Song, Li, Jin, Fatemi, Tsitsulin, Perozzi, Liu, and Tang}]{TSGFM}
Zhikai Chen, Haitao Mao, Jingzhe Liu, Yu~Song, Bingheng Li, Wei Jin, Bahare Fatemi, Anton Tsitsulin, Bryan Perozzi, Hui Liu, and Jiliang Tang. 2024{\natexlab{a}}.
\newblock Text-space graph foundation models: Comprehensive benchmarks and new insights.
\newblock In \emph{NeurIPS}.

\bibitem[{Chen et~al.(2024{\natexlab{b}})Chen, Wang, Liang, Risius, Demartini, and Yin}]{getfair}
Tong Chen, Danny Wang, Xurong Liang, Marten Risius, Gianluca Demartini, and Hongzhi Yin. 2024{\natexlab{b}}.
\newblock Hate Speech Detection with Generalizable Target-aware Fairness.
\newblock In \emph{KDD}.

\bibitem[{Dai et~al.(2023)Dai, Lang, Zeng, Huang, and Li}]{LLMOOD1}
Yi~Dai, Hao Lang, Kaisheng Zeng, Fei Huang, and Yongbin Li. 2023.
\newblock Exploring large language models for multi-modal out-of-distribution detection.
\newblock In \emph{EMNLP Findings}.

\bibitem[{Du et~al.(2023)Du, Sun, Zhu, and Li}]{Dream-OOD}
Xuefeng Du, Yiyou Sun, Jerry Zhu, and Yixuan Li. 2023.
\newblock Dream the impossible: Outlier imagination with diffusion models.
\newblock In \emph{NeurIPS}.

\bibitem[{Giles et~al.(1998)Giles, Bollacker, and Lawrence}]{Citeseer}
C.~Lee Giles, Kurt~D. Bollacker, and Steve Lawrence. 1998.
\newblock Citeseer: An automatic citation indexing system.
\newblock In \emph{ACM Digital Libraries}.

\bibitem[{{Google}(2024)}]{gemini2.5}
{Google}. 2024.
\newblock Gemini 2.5 flash.
\newblock Large language model, available at \url{https://gemini.google.com/}.

\bibitem[{Gui et~al.(2022)Gui, Li, Wang, and Ji}]{GOOD}
Shurui Gui, Xiner Li, Limei Wang, and Shuiwang Ji. 2022.
\newblock {GOOD:} {A} graph out-of-distribution benchmark.
\newblock In \emph{NeurIPS}.

\bibitem[{Guo et~al.(2023)Guo, Yang, Chen, Liu, Shi, and Du}]{AAGOD}
Yuxin Guo, Cheng Yang, Yuluo Chen, Jixi Liu, Chuan Shi, and Junping Du. 2023.
\newblock A data-centric framework to endow graph neural networks with out-of-distribution detection ability.
\newblock In \emph{KDD}.

\bibitem[{Ha et~al.(2017)Ha, Dai, and Le}]{hypernetwork}
David Ha, Andrew M. Dai, and Quoc V. Le. 2017.
\newblock HyperNetworks.
\newblock In \emph{ICLR}.

\bibitem[{Hendrycks and Gimpel(2017)}]{MSP}
Dan Hendrycks and Kevin Gimpel. 2017.
\newblock A baseline for detecting misclassified and out-of-distribution examples in neural networks.
\newblock In \emph{ICLR}.

\bibitem[{Hu et~al.(2020)Hu, Fey, Zitnik, Dong, Ren, Liu, Catasta, and Leskovec}]{Arxiv}
Weihua Hu, Matthias Fey, Marinka Zitnik, Yuxiao Dong, Hongyu Ren, Bowen Liu, Michele Catasta, and Jure Leskovec. 2020.
\newblock Open graph benchmark: Datasets for machine learning on graphs.
\newblock In \emph{NeurIPS}.

\bibitem[{Huang et~al.(2024)Huang, Han, Yang, Bao, Tao, Chai, and Zhu}]{reddit_text}
Xuanwen Huang, Kaiqiao Han, Yang Yang, Dezheng Bao, Quanjin Tao, Ziwei Chai, and Qi~Zhu. 2024.
\newblock Can gnn be good adapter for llms?
\newblock In \emph{WWW}.

\bibitem[{Ji et~al.(2010)Ji, Sun, Danilevsky, Han, and Gao}]{dblp}
Ming Ji, Yizhou Sun, Marina Danilevsky, Jiawei Han, and Jing Gao. 2010.
\newblock Graph regularized transductive classification on heterogeneous information networks.
\newblock In \emph{KDD}.

\bibitem[{Jin et~al.(2023)Jin, Zhang, Zhang, Meng, Zhang, Zhu, and Han}]{patton}
Bowen Jin, Wentao Zhang, Yu~Zhang, Yu~Meng, Xinyang Zhang, Qi~Zhu, and Jiawei Han. 2023.
\newblock Patton: Language model pretraining on text-rich networks.
\newblock In \emph{ACL}.

\bibitem[{Kingma and Ba(2015)}]{Adam}
Diederik~P. Kingma and Jimmy Ba. 2015.
\newblock Adam: {A} method for stochastic optimization.
\newblock In \emph{ICLR}.

\bibitem[{Kipf and Welling(2017)}]{GCN}
Thomas~N. Kipf and Max Welling. 2017.
\newblock Semi-supervised classification with graph convolutional networks.
\newblock In \emph{ICLR}.

\bibitem[{Lang et~al.(2023)Lang, Zheng, Li, Sun, Huang, and Li}]{NLPOODD}
Hao Lang, Yinhe Zheng, Yixuan Li, Jian Sun, Fei Huang, and Yongbin Li. 2023.
\newblock A survey on out-of-distribution detection in {NLP}.
\newblock \emph{CoRR}.

\bibitem[{Lee et~al.(2018)Lee, Lee, Lee, and Shin}]{Mahalanobis}
Kimin Lee, Kibok Lee, Honglak Lee, and Jinwoo Shin. 2018.
\newblock A simple unified framework for detecting out-of-distribution samples and adversarial attacks.
\newblock In \emph{NeurIPS}.

\bibitem[{Li et~al.(2022{\natexlab{a}})Li, Wang, Zhang, and Zhu}]{GOODSurvey}
Haoyang Li, Xin Wang, Ziwei Zhang, and Wenwu Zhu. 2022{\natexlab{a}}.
\newblock Out-of-distribution generalization on graphs: {A} survey.
\newblock \emph{CoRR}.

\bibitem[{Li et~al.(2024)Li, Wang, Zhu, Chen, Jiang, Cai, Chan, and Li}]{GLBENCH}
Yuhan Li, Peisong Wang, Xiao Zhu, Aochuan Chen, Haiyun Jiang, Deng Cai, Victor Wai~Kin Chan, and Jia Li. 2024.
\newblock Glbench: A comprehensive benchmark for graph with large language models.
\newblock In \emph{NeurIPS}.

\bibitem[{Li et~al.(2022{\natexlab{b}})Li, Wu, Nie, and Yan}]{GraphDE}
Zenan Li, Qitian Wu, Fan Nie, and Junchi Yan. 2022{\natexlab{b}}.
\newblock Graphde: {A} generative framework for debiased learning and out-of-distribution detection on graphs.
\newblock In \emph{NeurIPS}.

\bibitem[{Liang et~al.(2018)Liang, Li, and Srikant}]{ODIN}
Shiyu Liang, Yixuan Li, and R.~Srikant. 2018.
\newblock Enhancing the reliability of out-of-distribution image detection in neural networks.
\newblock In \emph{ICLR}.

\bibitem[{Lin and Gu(2023)}]{Flats}
Haowei Lin and Yuntian Gu. 2023.
\newblock {FL}at{S}: Principled out-of-distribution detection with feature-based likelihood ratio score.
\newblock In \emph{EMNLP}.

\bibitem[{Liu et~al.(2024{\natexlab{a}})Liu, Zhan, Lu, Feng, Xue, and Wu}]{LLMOOD2}
Bo~Liu, Liming Zhan, Zexin Lu, Yujie Feng, Lei Xue, and Xiao-Ming Wu. 2024{\natexlab{a}}.
\newblock How good are llms at out-of-distribution detection?
\newblock In \emph{COLING}.

\bibitem[{Liu et~al.(2024{\natexlab{b}})Liu, Feng, Kong, Liang, Tao, Chen, and Zhang}]{OFA}
Hao Liu, Jiarui Feng, Lecheng Kong, Ningyue Liang, Dacheng Tao, Yixin Chen, and Muhan Zhang. 2024{\natexlab{b}}.
\newblock One for all: Towards training one graph model for all classification tasks.
\newblock In \emph{ICLR}.

\bibitem[{Liu et~al.(2020)Liu, Wang, Owens, and Li}]{energy}
Weitang Liu, Xiaoyun Wang, John~D. Owens, and Yixuan Li. 2020.
\newblock Energy-based out-of-distribution detection.
\newblock In \emph{NeurIPS}.

\bibitem[{Liu et~al.(2023{\natexlab{a}})Liu, Qiu, and Huang}]{CaT}
Yilun Liu, Ruihong Qiu, and Zi~Huang. 2023{\natexlab{a}}.
\newblock Cat: Balanced continual graph learning with graph condensation.
\newblock In \emph{ICDM}.

\bibitem[{Liu et~al.(2025)Liu, Qiu, Tang, Yin, and Huang}]{Puma}
Yilun Liu, Ruihong Qiu, Yanran Tang, Hongzhi Yin, and Zi~Huang. 2025.
\newblock {PUMA:} efficient continual graph learning for node classification with graph condensation.
\newblock \emph{TKDE}.

\bibitem[{Liu et~al.(2023{\natexlab{b}})Liu, Ding, Liu, and Pan}]{GOOD-D}
Yixin Liu, Kaize Ding, Huan Liu, and Shirui Pan. 2023{\natexlab{b}}.
\newblock {GOOD-D:} on unsupervised graph out-of-distribution detection.
\newblock In \emph{WSDM}.

\bibitem[{Ma et~al.(2024)Ma, Sun, Ding, Liu, and Wu}]{grasp}
Longfei Ma, Yiyou Sun, Kaize Ding, Zemin Liu, and Fei Wu. 2024.
\newblock Revisiting score propagation in graph out-of-distribution detection.
\newblock In \emph{NeurIPS}.

\bibitem[{Ni et~al.(2019)Ni, Li, and McAuley}]{amazon_data_text}
Jianmo Ni, Jiacheng Li, and Julian~J. McAuley. 2019.
\newblock Justifying recommendations using distantly-labeled reviews and fine-grained aspects.
\newblock In \emph{EMNLP-IJCNLP}.

\bibitem[{{OpenAI}(2024)}]{gpt4omini}
{OpenAI}. 2024.
\newblock Gpt-4o mini.
\newblock Large language model, available at \url{https://openai.com/}.

\bibitem[{Reimers and Gurevych(2019)}]{sbert}
Nils Reimers and Iryna Gurevych. 2019.
\newblock Sentence-bert: Sentence embeddings using siamese bert-networks.
\newblock In \emph{EMNLP}.

\bibitem[{Sen et~al.(2008)Sen, Namata, Bilgic, Getoor, Gallagher, and Eliassi{-}Rad}]{Cora_Citeseer_pubmed}
Prithviraj Sen, Galileo Namata, Mustafa Bilgic, Lise Getoor, Brian Gallagher, and Tina Eliassi{-}Rad. 2008.
\newblock Collective classification in network data.
\newblock \emph{{AI} Mag.}, 29(3):93--106.

\bibitem[{Shen et~al.(2024)Shen, Wang, Zhou, Pan, and Wang}]{Graph_molecule_ood}
Xu~Shen, Yili Wang, Kaixiong Zhou, Shirui Pan, and Xin Wang. 2024.
\newblock Optimizing {OOD} detection in molecular graphs: {A} novel approach with diffusion models.
\newblock In \emph{KDD}.

\bibitem[{Song and Wang(2022)}]{OODGAT}
Yu~Song and Donglin Wang. 2022.
\newblock Learning on graphs with out-of-distribution nodes.
\newblock In \emph{KDD}.

\bibitem[{Stadler et~al.(2021)Stadler, Charpentier, Geisler, Z{\"{u}}gner, and G{\"{u}}nnemann}]{GPN}
Maximilian Stadler, Bertrand Charpentier, Simon Geisler, Daniel Z{\"{u}}gner, and Stephan G{\"{u}}nnemann. 2021.
\newblock Graph posterior network: Bayesian predictive uncertainty for node classification.
\newblock In \emph{NeurIPS}.

\bibitem[{Sun et~al.(2022)Sun, Ming, Zhu, and Li}]{KNN}
Yiyou Sun, Yifei Ming, Xiaojin Zhu, and Yixuan Li. 2022.
\newblock Out-of-distribution detection with deep nearest neighbors.
\newblock In \emph{ICML}.

\bibitem[{Tang et~al.(2024{\natexlab{a}})Tang, Qiu, Liu, Li, and Huang}]{CaseGNN}
Yanran Tang, Ruihong Qiu, Yilun Liu, Xue Li, and Zi~Huang. 2024{\natexlab{a}}.
\newblock Casegnn: Graph neural networks for legal case retrieval with text-attributed graphs.
\newblock In \emph{ECIR}.

\bibitem[{Tang et~al.(2024{\natexlab{b}})Tang, Qiu, Yin, Li, and Huang}]{CaseLink}
Yanran Tang, Ruihong Qiu, Hongzhi Yin, Xue Li, and Zi~Huang. 2024{\natexlab{b}}.
\newblock Caselink: Inductive graph learning for legal case retrieval.
\newblock In \emph{SIGIR}.

\bibitem[{Um et~al.(2025)Um, Lim, Kim, Yeo, and Jung}]{spread_ood}
Daeho Um, Jongin Lim, Sunoh Kim, Yuneil Yeo, and Yoonho Jung. 2025.
\newblock Spreading out-of-distribution detection on graphs.
\newblock In \emph{ICLR}.

\bibitem[{Wang et~al.(2025)Wang, Qiu, Bai, and Huang}]{GOLD}
Danny Wang, Ruihong Qiu, Guangdong Bai, and Zi~Huang. 2025.
\newblock {GOLD:} graph out-of-distribution detection via implicit adversarial latent generation.
\newblock \emph{ICLR}.

\bibitem[{Wang et~al.(2024{\natexlab{a}})Wang, Yang, Huang, Yang, Majumder, and Wei}]{e5-small}
Liang Wang, Nan Yang, Xiaolong Huang, Linjun Yang, Rangan Majumder, and Furu Wei. 2024{\natexlab{a}}.
\newblock Multilingual {E5} text embeddings: {A} technical report.
\newblock \emph{CoRR}.

\bibitem[{Wang et~al.(2024{\natexlab{b}})Wang, He, Zhang, Liu, Wang, Pan, Jin, and Chua}]{GOODAT}
Luzhi Wang, Dongxiao He, He~Zhang, Yixin Liu, Wenjie Wang, Shirui Pan, Di~Jin, and Tat{-}Seng Chua. 2024{\natexlab{b}}.
\newblock {GOODAT:} towards test-time graph out-of-distribution detection.
\newblock In \emph{AAAI}.

\bibitem[{Wang et~al.(2020)Wang, Wei, Dong, Bao, Yang, and Zhou}]{minilm}
Wenhui Wang, Furu Wei, Li~Dong, Hangbo Bao, Nan Yang, and Ming Zhou. 2020.
\newblock Minilm: Deep self-attention distillation for task-agnostic compression of pre-trained transformers.
\newblock In \emph{NeurIPS}.

\bibitem[{Wang et~al.(2024{\natexlab{c}})Wang, Zhang, Xie, Shi, Wang, Liu, Zhu, and Tan}]{OOD_TAG}
Yiqi Wang, Jiaxin Zhang, Nianhao Xie, Yu~Shi, Siwei Wang, Xinwang Liu, En~Zhu, and Yusong Tan. 2024{\natexlab{c}}.
\newblock \href {https://openreview.net/forum?id=FGIBKpOj8m} {Towards the effect of large language models on out-of-distribution challenge in text-attributed graphs}.

\bibitem[{Wu et~al.(2023)Wu, Chen, Yang, and Yan}]{GNNSafe}
Qitian Wu, Yiting Chen, Chenxiao Yang, and Junchi Yan. 2023.
\newblock Energy-based out-of-distribution detection for graph neural networks.
\newblock In \emph{ICLR}.

\bibitem[{Xu et~al.(2025{\natexlab{a}})Xu, Yao, Wang, Cheng, Hu, Li, and Zhao}]{Gsyn_tag}
Haoyan Xu, Zhengtao Yao, Ziyi Wang, Zhan Cheng, Xiyang Hu, Mengyuan Li, and Yue Zhao. 2025{\natexlab{a}}.
\newblock Graph synthetic out-of-distribution exposure with large language models.
\newblock \emph{CoRR}.

\bibitem[{Xu et~al.(2025{\natexlab{b}})Xu, Yao, Zhang, Wang, He, Dong, Yu, Li, and Zhao}]{GLIP-OOD}
Haoyan Xu, Zhengtao Yao, Xuzhi Zhang, Ziyi Wang, Langzhou He, Yushun Dong, Philip~S. Yu, Mengyuan Li, and Yue Zhao. 2025{\natexlab{b}}.
\newblock Glip-ood: Zero-shot graph ood detection with foundation model.
\newblock \emph{CoRR}.

\bibitem[{Xu and Ding(2025)}]{llmood_survey}
Ruiyao Xu and Kaize Ding. 2025.
\newblock Large language models for anomaly and out-of-distribution detection: A survey.
\newblock In \emph{NAACL Findings}.

\bibitem[{Yan et~al.(2023)Yan, Li, Long, Yan, Zhao, Zhuang, Yin, Zhang, Han, Sun, Deng, Zhang, Sun, Xie, and Wang}]{cstag}
Hao Yan, Chaozhuo Li, Ruosong Long, Chao Yan, Jianan Zhao, Wenwen Zhuang, Jun Yin, Peiyan Zhang, Weihao Han, Hao Sun, Weiwei Deng, Qi~Zhang, Lichao Sun, Xing Xie, and Senzhang Wang. 2023.
\newblock A comprehensive study on text-attributed graphs: Benchmarking and rethinking.
\newblock In \emph{NeurIPS}.

\bibitem[{Yang et~al.(2021)Yang, Zhou, Li, and Liu}]{OODD_cv_survey}
Jingkang Yang, Kaiyang Zhou, Yixuan Li, and Ziwei Liu. 2021.
\newblock Generalized out-of-distribution detection: {A} survey.
\newblock \emph{CoRR}.

\bibitem[{Yang et~al.(2024)Yang, Liang, Liu, Gui, Yao, and Zhang}]{NODESAFE}
Shenzhi Yang, Bin Liang, An~Liu, Lin Gui, Xingkai Yao, and Xiaofang Zhang. 2024.
\newblock Bounded and uniform energy-based out-of-distribution detection for graphs.
\newblock In \emph{ICML}.

\bibitem[{Zhao et~al.(2020)Zhao, Chen, Hu, and Cho}]{GKDE}
Xujiang Zhao, Feng Chen, Shu Hu, and Jin{-}Hee Cho. 2020.
\newblock Uncertainty aware semi-supervised learning on graph data.
\newblock In \emph{NeurIPS}.

\bibitem[{Zou et~al.(2023)Zou, Yu, Huang, Sun, and Du}]{TRN_Paper}
Tao Zou, Le~Yu, Yifei Huang, Leilei Sun, and Bowen Du. 2023.
\newblock Pretraining language models with text-attributed heterogeneous graphs.
\newblock In \emph{EMNLP Findings}.

\end{thebibliography}

\appendix

\section{Impact Statement}
Our work aims to inspire and pave the way for future works on text-rich network OOD detection for real-world applications. This foundational research uses only publicly available datasets and models, with all sources properly acknowledged. We do not identify any direct negative societal impacts requiring specific safeguards or emphasis in this study, though we encourage continuing ethical assessment as the field develops.

\begin{table*}[h!]
\centering
\setlength{\fboxsep}{8pt}
\setlength{\tabcolsep}{6pt}
\begin{tabular}{p{0.31\textwidth}|p{0.31\textwidth}|p{0.31\textwidth}}
\multicolumn{1}{c|}{\textbf{Original Text}} & 
\multicolumn{1}{c|}{\textbf{Synonym ($\alpha=0.5, p_{text}=0.3$)}} & 
\multicolumn{1}{c}{\textbf{Antonym ($\alpha=0.3, p_{text}=0.3$)}} \\
\midrule
Graph neural networks have shown promising results on semi-supervised node classification tasks. &
Graph \textcolor{blue}{nerve-based} netorks have shown promi\textcolor{blue}{ss}ing results on \textcolor{blue}{semi-labeled} node \textcolor{blue}{categorisation} tasks. &
Graph neural networks have shown \textcolor{red}{failesd} rsults on \textcolor{red}{unsupervised} node classifica\textcolor{red}{tt}ion \textcolor{red}{tz}sks. \\
\midrule
\rowcolor{gray!8}
Contrastive learning has emerged as a powerful paradigm for self-supervised representation learning. &
\textcolor{blue}{Contrative} learning has e\textcolor{blue}{n}merged as a \textcolor{blue}{strong} pardigm for self-\textcolor{blue}{labeled} \textcolor{blue}{feature} learning. &
\textcolor{red}{Unifying} learning has \textcolor{red}{collapsed} as a \textcolor{red}{weak} \textcolor{red}{heuristic} for  self-\textcolor{red}{unsupervised} representation  \textcolor{red}{lee}arning. \\
\midrule
Pre-trained language models are widely used for various natural language understanding tasks. &
Pre-\textcolor{blue}{traind} \textcolor{blue}{speech} mo\textcolor{blue}{a}ls are \textcolor{blue}{commonly} used for various natural \textcolor{blue}{textual} understanding \textcolor{blue}{assignments}. &
\textcolor{red}{Untraned} \textcolor{red}{imzge} \textcolor{red}{generaotrs} are widely used for \textcolor{red}{unnatural} language \textcolor{red}{forgetting} tasks. \\
\end{tabular}
\caption{Additional examples of text-level feature shift created by \textsc{TextAugmenT}.}
\label{Table:extended_text_augmentation_examples}
\end{table*}
\section{Extended Related Work} \label{Appendix:extended_related_work}
Our work explores the challenging problem of OOD detection in text-rich networks, intersecting with several important research trajectories that have evolved independently but are now converging in this complex domain.
\paragraph{Post-hoc OOD Detection Methods.}
Post-hoc methods represent a fundamental approach where detection relies solely on in-distribution data to construct specialised scoring functions~\cite{MSP, KNN, Mahalanobis, ODIN, Neco, energy, Flats}. These methods avoid re-training models, providing adaptive detection capabilities across diverse scenarios. Maximum Softmax Probability (MSP)\cite{MSP} utilises confidence scores from softmax outputs, while ODIN\cite{ODIN} enhances this with temperature scaling and input perturbation. Mahalanobis distance-based approaches~\cite{Mahalanobis} measure deviation from class-conditional Gaussian distributions. These approaches form the methodological foundation upon which specialised network-based detection methods build.

\paragraph{NLP OOD Detection.}
Recent OOD detection techniques in NLP increasingly leverage PLM/LLMs to improve robustness on novel inputs. Distance-based detectors applied to a model’s embedding space have shown strong performance with pre-trained encoders like BERT~\cite{BERT_OOD}. Moreover, research has been conducted on the effectiveness and identifying scenarios where LLM exhibit OOD detection capabilities~\cite{LLMOOD2}. Another emerging trend is harnessing LLMs to generate synthetic outlier examples for outlier exposure training, which markedly reduces false positives and improves detection without requiring real OOD data~\cite{LLM_OOD_Syn}. Notably, LLM-enhanced strategies has greatly improved the diversity in textual OOD detection, enabling more reliable identification of OOD instances in open-world settings.

\paragraph{Graph-Structured Data OOD Detection.}
In the domain of graph-structured data, diverse approaches have emerged to leverage network topology alongside node features~\cite{AAGOD, GraphDE,GPN,GKDE,Graph_molecule_ood, CaT, Puma, TopoOOD, spread_ood}. GNNSafe~\cite{GNNSafe} introduced an energy-score based method with a propagation schema to effectively harness the intrinsic interdependencies between node features and network topology. NODESafe~\cite{NODESAFE} extended this approach by refining extreme energy scores. GRASP~\cite{grasp} proposed an enhanced edge augmentation method to improve propagation effectiveness at test time. More complex algorithm designs have also been explored, including GOLD~\cite{GOLD}, which presents a data synthesis strategy to simulate OOD scenarios without requiring actual auxiliary OOD samples. DeGEM~\cite{DEGEM} provides a novel energy-based modelling architecture featuring multi-hop graph encoders coupled with dedicated energy heads, designed to improve detection over heterophilic network structures. We exclude direct comparison with transductive methods like GRASP as our work focuses on inductive tasks where OOD node features or edge connections are unavailable during training, representing a more challenging but realistic scenario.
\paragraph{Text-Rich Network OOD Detection.}
Text-rich network OOD detection methods remain relatively underexplored, despite their critical importance in real-world applications. Recent attempts have leveraged Large Language Models (LLMs) for generating OOD data or performing zero-shot OOD detection~\cite{GLIP-OOD, Gsyn_tag}. However, current research has primarily focused on label shift scenarios, ignoring the intricate interplay between textual semantics and structural properties that characterises real-world distribution shifts~\cite{OOD_TAG}. In real-world TrN environments, textual content and network structure evolve in tandem, creating complex patterns of distribution shift that cannot be captured by approaches designed for either modality in isolation.
To bridge this gap, we propose the TextTopoOOD framework to systematically and thoroughly investigate OOD detection under diverse textual, feature, and structural shifts. Complementing this framework, we propose TNT-OOD, a novel OOD detection method that explicitly models the interplay between textual and structural information for improved detection performance on TrNs.
Our work thus uniquely contributes to advancing OOD detection capabilities in text-rich network environments, addressing a  blind spot in current approaches and establishing both evaluation standards and methodological foundations for this important research direction.

\paragraph{HyperNetwork.} 
Hypernetworks~\cite{hypernetwork} are neural networks that generate the weights of another target model, enabling dynamic parameterisation and improved adaptability. They have been widely applied in few-shot learning and domain adaptation, where generalisation to unseen tasks or domains is essential~\cite{hypernetwork_survey}. For instance, it is used in tasks such as hate speech detection to enable detecting unseen domain samples~\cite{getfair}.

\section{Extended Discussion on OOD Shifts in TextTopoOOD.}
\label{Appendix:ood_shifts}
\paragraph{Label Shift.} The LLM prompt for selecting the thematically-guided label-leave-out classes is provided in Prompt~\ref{prompt:llm_prompt}. We provide three strategies to select OOD classes: 1) Random Selection, 2) Thematically similar labels w.r.t. ID labels, and 3) Thematically dissimilar labels to the ID labels. We used GPT-4o as the LLM model, giving the category names and descriptions (if available) as input. 

\paragraph{Text Shift.} We provide further examples of text-level shifts involving synonym, antonym replacements as well as character level edits in Table~\ref{Table:extended_text_augmentation_examples}.

\paragraph{Domain-Based Sentiment Shift.}
For datasets with underlying sentiment properties (e.g., review ratings), we can create domain shifts based on the sentiment expressed in the text (i.e., Positive vs. Negative reviews). Node sets can be defined based on their sentiment $s(v_i)$:
\begin{align}
\mathcal{V}_{\text{ID}} &= \{v_i \in \mathcal{V} \mid s(v_i) \in S_{\text{ID}}\} \\
\mathcal{V}_{\text{OOD}} &= \{v_i \in \mathcal{V} \mid s(v_i) \in S_{\text{OOD}}\},
\end{align}
where $S_{\text{ID}}$ is the set of ID sentiments (e.g., \{`neutral', `positive'\}) and $S_{\text{OOD}}$ is a disjoint set of OOD sentiments (e.g., \{`negative'\}). The sentiments can be obtained via a sentiment analysis model (i.e., sentiment pre-trained Bert). The textual content associated with these different sentiment groups often exhibits distinct vocabulary, stylistic features, and potentially different graph connectivity patterns if sentiment influences interactions.

\begin{table*}[t!]
\centering
\resizebox{0.9\textwidth}{!}{%
\begin{tabular}{lccccl}
\toprule
\textbf{Name} & \textbf{\#Nodes} & \textbf{\#Edges} & \textbf{\#Classes} & \textbf{Domain} & \textbf{Text Characteristics} \\
\midrule
\multicolumn{6}{c}{\textit{Citation Networks}} \\
\midrule
Cora & 2,708 & 10,556 & 7 & CS Citation & Paper title \& abstracts  \\
CiteSeer & 3,186 & 8,450 & 6 & CS Citation  & Paper title \& abstracts  \\
DBLP & 14,376 & 431,326 & 4 & CS Citation & Paper title \& abstracts \\
Arxiv & 169,343 & 1,166,243 & 40 & CS Citation & Paper title \& abstracts  \\
PubMed & 19,717 & 88,648 & 3 & Bio Citation & Medical title \& abstracts \\
\midrule
\multicolumn{6}{c}{\textit{E-commerce Networks}} \\
\midrule
History (BookHis) & 41,551 & 358,574 & 12 & E-commerce & Book title \& descriptions \\
Child (BookChild) & 76,875 & 1,554,578 & 24 & E-commerce & Book title \& descriptions  \\
Computers (EleComp) & 87,229 & 721,081 & 10 & E-commerce & Product reviews \\
Photo (ElePhoto) & 48,362 & 500,939 & 12 & E-commerce & Product reviews \\
\midrule
\multicolumn{6}{c}{\textit{Other Networks}} \\
\midrule
WikiCS & 11,701 & 431,726 & 10 & Knowledge & Encyclopedia articles \\
Reddit & 33,434 & 302,876 & 2 & Social Forum & Social media posts \\
\bottomrule
\end{tabular}
}
\caption{Statistics of datasets used in TextTopoOOD. The datasets span citation networks, e-commerce networks, and knowledge graphs.}
\label{tab:dataset_stats}
\end{table*}

\section{Extended Dataset Description} \label{Appendix:dataset_description}
We have utilised the publicly available text-rich networks provided by TSGFM and GLBench~\cite{TSGFM, GLBENCH}, acknowledge and follow the MIT license. We adhere to the provided splits for ID classification, except for label shift and time-shift where we would filter out the associated OOD labels from the ID data. The dataset statistics are presented in Table~\ref{tab:dataset_stats}.

\paragraph{Scale:} The datasets range from small-scale networks (Cora: $\sim$2.7K nodes) to large-scale networks (Products: $\sim$316K nodes), with edge counts ranging from thousands to millions.
\paragraph{Domains:} The datasets span four primary domains, dataset details are provided below:
\begin{itemize}[leftmargin=*, itemsep=0pt, topsep=2pt, parsep=0pt, partopsep=0pt]
    \item \textbf{Citation networks} (Cora, CiteSeer, DBLP, Arxiv, PubMed)
    \item \textbf{E-commerce networks} (Bookhis, Bookchild, Elecomp, Elephoto)
    \item \textbf{Knowledge and social networks} (WikiCS, Reddit)
\end{itemize}

\subsection{OOD Test Data Construction.} In this paper, we consider the following configurations of the OOD scenarios of TextTopoOOD as presented in Sec~\ref{sec:TextTopoOOD}. For the OOD shifts, we consider an inductive setting, where OOD node feature/structure (depending on the shift type) and labels are not available during training.
\paragraph{Text Augmentation.} For text augmentation (sec~\ref{sec:text_aug}), we utilise NLTK v.3.9.1's WordNet\footnote{\url{https://www.nltk.org/}} to build a cache of possible \textbf{synonyms} and \textbf{antonyms} for all unique words in the datasets. Since academic content often lacks suitable replacement terms, we set $$\alpha_{\text{text}} = 1 \quad \text{and} \quad p_{\text{char}} = 1,$$ 
in Eq.~\ref{eq:textaugment} to introduce sufficient diversity for simulating suitable OOD scenarios.

\paragraph{Feature Mixing.} For feature mixing (sec~\ref{sec:feat_mix}), we implement three different noise levels for each data: 
$$\alpha_{\text{feat}} \in \{0.5, 0.7, 0.9\}.$$
This would gradually increase the diversity between the original ID embedding and the transformed OOD embeddings. Example embeddings from a trained classifier are shown in Figure~\ref{fig:score_distribution_viz}.

\paragraph{Structure Rewiring.} For structure rewiring shift (sec~\ref{sec:struct_rewire}), we calculate the density as:
$$\rho = \frac{\text{num ID edges}}{\text{num possible edges}}.$$
This ensures our generated structure does not deviate too far in terms of graph density, mitigating undesired OOD test cases.
Following this, we define \textbf{three levels of structure shifts} with $\beta, f_{ii}, f_{ij}$ as follows:
\begin{itemize}
    \item \textbf{Mild}: $\beta = 0.2 , f_{ii} = 0.7 , f_{ij} = 0.5$
    \item \textbf{Medium}: $\beta = 0.5, f_{ii} = 0.6, f_{ij} = 0.3$
    \item \textbf{Strong}: $\beta = 1.0, f_{ii}= 0.4, f_{ij} = 0.7$
\end{itemize}

\paragraph{Semantic Connection.} For semantic connection shift (sec~\ref{sec:semantic_connect}), to ensure a challenging and valid OOD shifts, $k$ was selected to be the same number of edges as the original ID graph. Moreover, we define three thresholds to test:
$$ \text{threshold} \in \{0.75, 0.85, 0.95\},$$
where we would select edges based on the similarity values at these given thresholds (i.e., $k/2$ edges directly greater \& lower than the similarity value at the 0.75\% threshold.) This introduces diversity based on a gradual increase in similarity between the connected nodes in the OOD test set.

\paragraph{Text Swap.} For text swap shift (sec~\ref{sec:text_swap}), we experiment with all three introduced variants: \textbf{1)Intra-class, 2)Inter-class, 3)Random Swap}, all with a swap ratio of 
$$\beta_{\text{swap}} = 1.$$
This ensures a maximum distinction against the original ID network.

\paragraph{Label Shift.} For label shifts (sec~\ref{sec:label_shift}), we again test the three strategies proposed: \textbf{1) Random selection, 2) Thematic similarity, 3) Thematic dissimilarity}. We perform label shift on datasets above four available classes (i.e., excluding datasets like Pubmed, DBLP, and Reddit), this ensures that there is enough ID data for training and our OOD contains diverse labels. The number of OOD classes selected are dependent on the dataset, in general, we select 10\% to $\approx$ 40\% of the available classes as OOD (i.e., for Cora with 7 classes, we select 3 OOD classes and 4 ID classes). Using the prompt in Figure~\ref{prompt:llm_prompt}, we can select the thematically similar and dissimilar ID and OOD classes. The detailed OOD classes are shown in Table~\ref{tab:ood_classes}:
\begin{table}[h]
    \centering
    \resizebox{1\columnwidth}{!}{%
    \begin{tabular}{lccc}
        \toprule
        \textbf{Dataset} & \textbf{Random} & \textbf{Similar} & \textbf{Dissimilar} \\
        \midrule
        Cora & $\{0$-$2\}$ & $\{1,3,4\}$ & $\{2,5,6\}$ \\
        Citeseer & $\{1,3,5\}$ & $\{2$-$4\}$ & $\{0,1,5\}$ \\
        WikiCS & $\{1,3$-$5,8\}$ & $\{0,1,8,9\}$ & $\{2$-$4,6\}$ \\
        Bookhis & $\{0,1,3,4,7\}$ & $\{1,2,6,10,11\}$ & $\{0,3,5,8,9\}$ \\
        Bookchild & $\{0,2,5,9$-$11,15,18\}$ & $\{1,3,4,6,12,19,21,23\}$ & $\{0,2,5,7$-$10,22\}$ \\
        Elephoto & $\{1,3,5,7,9\}$ & $\{0,4,6,10,11\}$ & $\{1$-$3,5,9\}$ \\
        Elecomp & $\{1$-$5\}$ & $\{2$-$4,8,9\}$ & $\{0,1,5$-$7\}$ \\
        \bottomrule
    \end{tabular}%
    }
    \caption{OOD Class Configurations by Dataset}
    \label{tab:ood_classes}
\end{table}

\paragraph{Domain Shift.} For domain shifts (sec~\ref{sec:domain_shifts}), we split the Arxiv-dataset based on temporal information. The time ranges are as follow:
\begin{itemize}
    \item ID Time Range:  1960-2015
    \item OOD Time Range 1: 2017-2018
    \item OOD Time Range 2: 2018-2019
    \item OOD Time Range 3: 2019-2020
\end{itemize}
Notably, we evaluate three distinct time periods as OOD scenarios. For each temporal split, we include only nodes published within that specific period, with accessible edges limited to citations that existed up to the end date of that period - accurately representing the available information at that point in time.

We provide dataset statistics for each OOD shift type in Table~\ref{Table:ood_stat}, using Cora as an example. Text and feature-level shifts preserve graph structure, maintaining identical node and edge counts as the ID data. Structure-level shifts modify edge counts based on connectivity intensity, though semantic connection shifts maintain the original edge count while reorganising connections. For label shifts, ID statistics vary depending on which classes are designated as OOD, while OOD data maintains the complete edge set to accurately represent the inductive setting where the full graph structure remains accessible during inference with its associated OOD nodes.
\begin{table}[h!]
    \centering
    \begin{tabular}{lcc}
        \toprule
        \textbf{Data Split} & \textbf{Nodes} & \textbf{Edges} \\
        \midrule
        ID (Main) & 2,708 & 10,556 \\
        \midrule
        Feature \& Text shifts & 2,708 & 10,556 \\
        \midrule
        \multicolumn{3}{l}{\textit{Structure-level Shifts}} \\
        Structural (Medium) & 2,708 & 11,690 \\
        Semantic Connect & 2,708 & 10,556 \\
        \midrule
        \multicolumn{3}{l}{\textit{Label Shifts}} \\
        Label Shift 1 (ID) & 1,412 & 5,314 \\
        Label Shift 1 (OOD) & 1,296 & 10,556 \\
        Label Shift 2 (ID) & 1,767 & 6,248 \\
        Label Shift 2 (OOD) & 941 & 10,556 \\
        Label Shift 3 (ID) & 1,121 & 3,720 \\
        Label Shift 3 (OOD) & 1,587 & 10,556 \\
        \bottomrule
    \end{tabular}
    \caption{ID/OOD Dataset Statistics for Cora}
    \label{Table:ood_stat}
\end{table}

\subsection{Dataset description}
The detailed descriptions of the dataset 
are as follows: 
    \paragraph{\textsc{Cora}} This dataset represents a citation network where nodes correspond to academic papers with raw text being the title and abstract, and edges denote citation relationships~\cite{Cora_Citeseer_pubmed}. Each paper is classified into one of seven categories.

\paragraph{\textsc{Citeseer}} This is another citation network~\cite{Citeseer, Cora_Citeseer_pubmed}, where nodes represent scientific papers classified into one of six classes, and edges denote citation relationships, text data are also title and abstract. This dataset contains more nodes but fewer edges than Cora.

\paragraph{\textsc{DBLP}} This is citation network~\cite{dblp}, where nodes represent scientific paper classified into one of four categories (area of work), and edges denote co-authorship relationship, text data are paper titles.

\paragraph{\textsc{Arxiv}} This large-scale citation network spans research papers from 1960 to 2020~\cite{Arxiv}, with nodes representing papers categorised by subject area, edges indicating citations, and the raw text are titles and abstracts.

\paragraph{\textsc{Pubmed}} This is a biomedical paper citation network~\cite{Cora_Citeseer_pubmed}, each node is classified into one of three categories. The nodes represent academic papers with title and abstract as text data, while edges are citation relationships.

\paragraph{\textsc{Bookhis} \& \textsc{BookChild}} These are e-commerce network originated from the Amazon-books dataset: History and Children subsets~\cite{cstag}. This dataset models an item co-purchasing network, where nodes represent products (books), and edges indicate frequently co-purchased/co-viewed items. Node text capture product descriptions, and labels correspond to product categories~\cite{amazon_data_text}.

\paragraph{\textsc{Elephoto} \& \textsc{Elecomp}} These are e-commerce network from the Amazon-electronics dataset. The former is sourced from the Photo subset, while the latter comes from the Computers subset~\cite{cstag}. This dataset models an item co-purchasing network, where nodes represent products, and edges indicate frequently co-purchased/co-viewed items. Node text capture product reviews (i.e., highest upvoted review of the item), and labels correspond to product categories~\cite{amazon_data_text}.

\paragraph{\textsc{WikiCS}} This a knowledge network representing Wikipedia linkage, where nodes represent pages and edges denote reference links between them. Node text include page titles and content, while labels correspond to Wikipedia entry categories~\cite{GLBENCH, OFA}.

\paragraph{\textsc{Reddit}} This is a social network where nodes represent users and edges indicate reply interactions. Text features include content from users' last three subreddit posts, with labels categorising users as either popular or normal~\cite{GLBENCH, reddit_text}.

\section{Metrics} \label{Appendix:Metrics}
For evaluation of OOD detection methods, we adopt three widely used \textbf{manual threshold-independent metrics}~\cite{energy, GNNSafe, NODESAFE, grasp, Neco}: \textbf{AUROC, AUPR, and FPR95}.
The Area Under the Receiver Operating Characteristic curve (AUROC) provides a global view of the trade-off between true positive rate (TPR) and false positive rate (FPR) across all thresholds, indicating the model's overall discrimination capability. However, AUROC may be less reliable in settings with imbalance scenarios. To address this, we also consider the Area Under the Precision-Recall curve (AUPR), which better reflects model performance in imbalanced scenarios by focusing on both precision and recall. Lastly, the False Positive Rate at 95\% True Positive Rate (FPR95) quantifies the proportion of OOD samples falsely detected as ID when the model achieves high TPR, highlighting the robustness of the detection system under stringent sensitivity demands. Collectively, these metrics offer a balanced evaluation of both general and high-sensitivity OOD detection performance.

\section{Implementation Details} \label{Appendix:implementation_details}
Experiments were conducted using Python 3.9.2 and PyTorch 2.5.1 with Cuda 12.2 on a single NVIDIA RTX A6000 GPU with 48GB of memory. All baseline and TNT-OOD's hyperparameters are searched over the following parameters: num layers $\in \{2, 3, 4\}$, hidden dimension $\in \{64, 128, 256\}$, learning rate $\in \{0.1,\dots, 0.0001\}$, dropout $\in \{0,\dots,0.7 \}$. We ran the baseline results to the best of our ability. For text encoding, we use the All-MiniLM-L6-v2; a 22.7M parameter model provided on huggingface under the Apache license 2.0 license\footnote{\url{https://huggingface.co/sentence-transformers/all-MiniLM-L6-v2}}, this generates a 384 dimensional dense vector for each text sample. All experiments were conducted over three seeds, we report the mean and standard deviation in our paper. The Adam optimiser is used for training~\citep{Adam}. As described in Section~\ref{sec:experiments}, the number of propagation iterations $K$ is set to 3, with $\alpha_{\text{score}}$ fixed to 0.5. We further provide the hyperparameter sensitivity analysis for $\lambda$ in Eq.~\ref{eq:final_loss}, and $\tau$ in Eq.~\ref{eq:contrastive_loss} in Tables~\ref{table:hyper_lambda} and~\ref{table:hyper_tau}. The experiment were conduct on all shifts except for label shift.

\begin{table}[h!]
\centering
\resizebox{1\linewidth}{!}{
\begin{tabular}{c|c|ccc}
\midrule
$\lambda$ & Metric & Cora & Pubmed & Wikics \\
\midrule
\multirow{4}{*}{0} & AUROC & 90.93 $\pm$ 4.29 & 86.05 $\pm$ 3.92 & 89.80 $\pm$ 11.10 \\
& AUPR & 89.20 $\pm$ 3.85 & 82.59 $\pm$ 5.14 & 89.94 $\pm$ 6.67 \\
& FPR & 26.72 $\pm$ 11.29 & 36.38 $\pm$ 7.88 & 26.18 $\pm$ 27.85 \\
& ID Acc & 82.32 $\pm$ 0.20 & 78.43 $\pm$ 0.14 & 80.15 $\pm$ 0.23 \\
\midrule
\multirow{4}{*}{0.001} & AUROC & 91.01 $\pm$ 4.30 & 86.25 $\pm$ 3.79 & 89.92 $\pm$ 10.89 \\
& AUPR & 89.40 $\pm$ 3.82 & 82.93 $\pm$ 4.85 & 90.01 $\pm$ 6.52 \\
& FPR & 26.47 $\pm$ 11.37 & 36.12 $\pm$ 7.89 & 26.03 $\pm$ 27.65 \\
& ID Acc & 82.50 $\pm$ 0.18 & 78.41 $\pm$ 0.12 & 80.19 $\pm$ 0.22 \\
\midrule
\multirow{4}{*}{0.01} & AUROC & 90.62 $\pm$ 4.08 & 86.90 $\pm$ 3.79 & \textbf{90.77 $\pm$ 8.82} \\
& AUPR & 89.06 $\pm$ 3.76 & 83.99 $\pm$ 4.76 & \textbf{90.59 $\pm$ 5.15} \\
& FPR & 27.50 $\pm$ 10.75 & 35.23 $\pm$ 7.98 & \textbf{25.03 $\pm$ 24.65} \\
& ID Acc & 82.11 $\pm$ 0.97 & 78.44 $\pm$ 0.08 & \textbf{80.21 $\pm$ 0.28} \\
\midrule
\multirow{4}{*}{0.1} & AUROC & 91.35 $\pm$ 4.37 & 88.24 $\pm$ 3.56 & 92.01 $\pm$ 6.91 \\
& AUPR & 90.01 $\pm$ 3.92 & 86.18 $\pm$ 4.50 & 91.42 $\pm$ 3.94 \\
& FPR & 25.40 $\pm$ 11.95 & 34.70 $\pm$ 7.73 & 23.31 $\pm$ 23.55 \\
& ID Acc & 82.51 $\pm$ 0.91 & 78.49 $\pm$ 0.18 & 80.12 $\pm$ 0.21 \\
\midrule
\multirow{4}{*}{0.5} & AUROC & 91.87 $\pm$ 4.77 & \textbf{88.82 $\pm$ 3.22} & 91.64 $\pm$ 9.00 \\
& AUPR & 90.81 $\pm$ 4.19 & \textbf{86.67 $\pm$ 3.98} & 91.30 $\pm$ 5.06 \\
& FPR & 23.81 $\pm$ 13.58 & \textbf{33.13 $\pm$ 7.47} & 23.52 $\pm$ 29.60 \\
& ID Acc & 82.50 $\pm$ 0.78 & \textbf{78.85 $\pm$ 0.23} & 80.04 $\pm$ 0.21 \\
\midrule
\multirow{4}{*}{1} & AUROC & \textbf{91.65 $\pm$ 4.61} & 88.17 $\pm$ 3.17 & 85.19 $\pm$ 20.53 \\
& AUPR & \textbf{90.55 $\pm$ 4.06} & 86.04 $\pm$ 3.77 & 87.52 $\pm$ 12.01 \\
& FPR & \textbf{24.91 $\pm$ 13.11} & 35.20 $\pm$ 7.40 & 33.35 $\pm$ 34.78 \\
& ID Acc & \textbf{82.54 $\pm$ 0.25} & 78.59 $\pm$ 0.28 & 72.95 $\pm$ 0.61 \\
\midrule
\end{tabular}
}
\caption{Hyperparameter analysis for $\lambda$. Bold highlights the parameters selected.}
\label{table:hyper_lambda}
\end{table}

\begin{table}[h!]
\centering
\resizebox{1\linewidth}{!}{
\begin{tabular}{c|c|cc}
\midrule
$\tau$ & Metric & Bookhis & Elecomp \\
\midrule
\multirow{4}{*}{0.0001} & AUROC & 34.20 $\pm$ 19.87 & 47.13 $\pm$ 19.74 \\
& AUPR & 76.50 $\pm$ 7.89 & 87.63 $\pm$ 7.68 \\
& FPR & 91.91 $\pm$ 6.77 & 77.85 $\pm$ 2.47 \\
& ID Acc & 72.51 $\pm$ 1.42 & 58.84 $\pm$ 0.70 \\
\midrule
\multirow{4}{*}{0.001} & AUROC & 61.55 $\pm$ 9.23 & 64.06 $\pm$ 7.06 \\
& AUPR & 85.14 $\pm$ 4.44 & 91.28 $\pm$ 4.74 \\
& FPR & 60.87 $\pm$ 10.08 & 57.93 $\pm$ 10.39 \\
& ID Acc & 78.86 $\pm$ 4.70 & 78.63 $\pm$ 0.20 \\
\midrule
\multirow{4}{*}{0.01} & AUROC & 71.24 $\pm$ 11.23 & 66.37 $\pm$ 6.63 \\
& AUPR & 88.14 $\pm$ 4.49 & 91.72 $\pm$ 4.71 \\
& FPR & 49.01 $\pm$ 18.80 & 53.43 $\pm$ 10.97 \\
& ID Acc & 83.18 $\pm$ 0.63 & 83.14 $\pm$ 0.50 \\
\midrule
\multirow{4}{*}{0.1} & AUROC & \textbf{71.75 $\pm$ 17.40} & \textbf{68.27 $\pm$ 6.73} \\
& AUPR & \textbf{88.51 $\pm$ 5.87} & \textbf{92.11 $\pm$ 4.66} \\
& FPR & \textbf{45.57 $\pm$ 26.28} & \textbf{47.63 $\pm$ 7.39} \\
& ID Acc & \textbf{83.86 $\pm$ 0.12} & \textbf{84.75 $\pm$ 0.09} \\
\midrule
\multirow{4}{*}{1} & AUROC & 70.89 $\pm$ 11.44 & 67.20 $\pm$ 6.06 \\
& AUPR & 88.19 $\pm$ 4.22 & 91.92 $\pm$ 4.43 \\
& FPR & 49.71 $\pm$ 18.09 & 51.58 $\pm$ 12.03 \\
& ID Acc & 83.60 $\pm$ 0.30 & 84.70 $\pm$ 0.18 \\
\midrule
\end{tabular}
}
\caption{Hyperparameter analysis for $\tau$. Bold highlights the optimal parameter.}
\label{table:hyper_tau}
\end{table}

\begin{algorithm}[!t]
\caption{Character-level Edit}
\label{alg:char_edit}
\begin{algorithmic}[1]
\Function{CharEdit}{$w$}
    \State \textbf{Input:} Word $w$
    \State \textbf{Output:} Edited word $w'$
    
    \State $\text{ops} \gets \{\text{"ins"}, \text{"del"}, \text{"replace"}, \text{"swap"}\}$
    \State $\text{op} \gets \text{Sample}(\text{ops})$
    \State $\text{pos} \gets \text{RandInt}(0, |w|-1)$ 
    
    \If{$\text{op} = \text{"ins"}$}
        \State $w' \gets w[0:\text{pos}] + \text{RandChar}() + w[\text{pos}:]$
    \ElsIf{$\text{op} = \text{"del"} \text{ and } |w| > 1$}
        \State $w' \gets w[0:\text{pos}] + w[\text{pos}+1:]$
    \ElsIf{$\text{op} = \text{"replace"}$}
        \State $c \gets \text{RandChar}() \neq w[\text{pos}]$
        \State $w' \gets w[0:\text{pos}] + c + w[\text{pos}+1:]$
    \ElsIf{$\text{op} = \text{"swap"} \text{ and } |w| > 1$}
        \State $\text{pos} \gets \min(\text{pos}, |w|-2)$
        \State $w' \gets w[0:\text{pos}] + w[\text{pos}+1] + w[\text{pos}]$
        \State $w' \gets w' + w[\text{pos}+2:]$
    \Else
        \State $w' \gets w$
    \EndIf
    
    \State \Return $w'$
\EndFunction
\end{algorithmic}
\end{algorithm}

\begin{algorithm}[!t]
\caption{Text Augmentation}
\label{alg:text_augment}
\begin{algorithmic}[1]
\Function{TextAugment}{$t$, type, $\alpha$, $p_{\text{char}}$}
    \State \textbf{Input:} Text $t$, type, noise level $\alpha$, edit prob $p_{\text{char}}$
    \State \textbf{Output:} Augmented text $\tilde{t}$
    
    \State $\text{words}, \text{tags} \gets \text{PosTag}(\text{Tokenize}(t))$ 
    
    \State $\text{candidates} \gets \{i : \text{IsEligible}(\text{words}[i])\}$
    \State $n \gets \lfloor \alpha \cdot |\text{candidates}| \rfloor$
    \State $\text{idxs} \gets \text{Sample}(\text{candidates}, n)$
    
    \For{$i \in \text{idxs}$}
        \State $\text{alts} \gets \text{Wordnet}(\text{words}[i], \text{type})$
        \If{$\text{alts} \neq \emptyset$}
            \State $\text{words}[i] \gets \text{Sample}(\text{alts})$
            
            \If{$\text{Random}() < p_{\text{char}}$}
                \State $\text{words}[i] \gets \text{CharEdit}(\text{words}[i])$
            \EndIf
        \EndIf
    \EndFor
    
    \State \Return $\text{Detokenize}(\text{words})$
\EndFunction
\end{algorithmic}
\end{algorithm}

\begin{algorithm}[!t]
\caption{Text Swap for OOD Generation}
\label{alg:text_swap}
\begin{algorithmic}[1]
\Function{TextSwap}{$\mathcal{G}(\mathcal{V}, \mathcal{T}, \mathbf{A})$, $\mathbf{Y}$, scope, $\beta$}
    \State \textbf{Input:} Network $\mathcal{G}$, labels $\mathbf{Y}$, swap scope; ratio $\beta$
    \State \textbf{Output:} Network $\mathcal{G'}$ with swapped texts $\tilde{\mathcal{T}}$
    
    \State $\text{pairs} \gets \{\}$ \Comment{Initialize eligible node pairs}
    
    \If{scope = "intra-class"}
        \State $\text{pairs} \gets \{(i,j) \mid i,j \in \mathcal{V}, \mathbf{Y}_i = \mathbf{Y}_j, i \neq j\}$ \Comment{Same class}
    \ElsIf{scope = "inter-class"}
        \State $\text{pairs} \gets \{(i,j) \mid i,j \in \mathcal{V}, \mathbf{Y}_i \neq \mathbf{Y}_j\}$ \Comment{Different classes}
    \ElsIf{scope = "random"}
        \State $\text{pairs} \gets \{(i,j) \mid i,j \in \mathcal{V}, i \neq j\}$ \Comment{Any nodes}
    \EndIf
    
    \State $n_{\text{swap}} \gets \lfloor \beta \cdot |\mathcal{V}| \rfloor$ \Comment{Number of nodes to swap}
    \State $n_{\text{pairs}} \gets \lceil n_{\text{swap}}/2 \rceil$ \Comment{Number of pairs needed}
    
    \State $\text{selected\_pairs} \gets \text{SampleWithoutReplacement}(\text{pairs}, n_{\text{pairs}})$
    \State $\tilde{\mathcal{T}} \gets \mathcal{T}$ \Comment{Initialize with original texts}
    
    \For{each $(i,j)$ in $\text{selected\_pairs}$}
        \State $\tilde{\mathcal{T}}_i, \tilde{\mathcal{T}}_j \gets \mathcal{T}_j, \mathcal{T}_i$ \Comment{Swap texts}
    \EndFor
    
    \State $\mathcal{G'} \gets (\mathcal{V}, \tilde{\mathcal{T}}, \mathbf{A})$ \Comment{Update with swapped features}
    \State \Return $\mathcal{G'}$
\EndFunction
\end{algorithmic}
\end{algorithm}

\section{Algorithm}
\label{Appendix:text_augment}
Algorithms for the \textsc{TextAugment} function in Eq.~\ref{eq:textaugment} is provided in Algorithm~\ref{alg:text_augment}. The associated character-level edit function is presented in Algorithm~\ref{alg:char_edit}. The \textsc{TextSwap} function given in Algorithm~\ref{alg:text_swap}.

\section{Computational Cost} \label{Appendix:computational_cost}
Table~\ref{Table:computational_cost} compares the computational costs of TNT-OOD against GNNSafe and NODESafe. As expected, TNT-OOD's cross-attention mechanism and HyperNetwork module increase memory usage and training time compared to standard GCN models with post-hoc detection. However, we significantly reduced these costs by implementing batch-based contrastive objective calculations and a LowRank approach for the HyperNetwork, decreasing memory requirements by up to 80\%. Importantly, inference speeds remain comparable across methods, demonstrating TNT-OOD's practical viability for real-world applications -- specially with the superior detection performance demonstrated by TNT-OOD. As an early contribution to this underexplored field, we believe our work establishes a foundation for developing both effective and efficient OOD detection methods for text-rich networks.
\begin{table}[!t]
    \centering
    \resizebox{\columnwidth}{!}{%
    \begin{tabular}{l|c|cccc}
        \toprule
        & & \textbf{GNNSafe} & \textbf{NODESafe} & \textbf{TNT-OOD} & \textbf{TNT w/ batch} \\
        \midrule
        \multirow{3}{*}{\textbf{Cora}} 
        & Train (s) & 2.95 & 3.50 & 11.85 & 8.95 \\
        & Infer (s) & 0.01 & 0.02 & 0.05 & 0.06 \\
        & Mem (MiB) & 590 & 590 & 1022 & 852 \\
        \midrule
        \multirow{3}{*}{\textbf{Citeseer}} 
        & Train (s) & 2.60 & 2.79 & 11.32 & 12.90 \\
        & Infer (s) & 0.03 & 0.04 & 0.07 & 0.07 \\
        & Mem (MiB) & 654 & 654 & 2029 & 893 \\
        \midrule
        \multirow{3}{*}{\textbf{Pubmed}} 
        & Train (s) & 4.50 & 4.60 & 36.20 & 17.58 \\
        & Infer (s) & 0.15 & 0.16 & 0.20 & 0.22 \\
        & Mem (MiB) & 852 & 852 & 9045 & 1956 \\
        \midrule
        \multirow{3}{*}{\textbf{Arxiv}} 
        & Train (s) & 15.00 & 16.00 & OOM & 276.48 \\
        & Infer (s) & 0.76 & 0.75 & OOM & 0.82 \\
        & Mem (MiB) & 3326 & 3326 & OOM & 16352 \\
        \midrule
        \multirow{3}{*}{\textbf{Bookhis}} 
        & Train (s) & 4.50 & 4.62 & 522.34 & 49.82 \\
        & Infer (s) & 0.18 & 0.18 & 0.27 & 0.25 \\
        & Mem (MiB) & 1742 & 1742 & 12936 & 5222 \\
        \bottomrule
    \end{tabular}%
    }
    \caption{Computational Performance on a single NVIDIA RTX A6000 GPU (48GB). Train: Convergence time; Infer: Inference time; Mem: Peak memory usage; OOM: Out of memory}
    \label{Table:computational_cost}
\end{table}

\section{Extended Experiment Results} \label{Appendix:extended_experiments}
In this section, we provide the extended results complementing the overall performance table provided in Table~\ref{table:overall_performance} in the main paper. Specifically, the OOD detection performance for each OOD scenario at different configurations on each dataset are provided in Tables \ref{table:cora_ood_full} to \ref{table:elecomp_ood_full}. The scores reported in the subset tables are averaged across three runs, with variance reflecting performance deviation across the three seeds. 

\begin{table}[!t]
   \centering
   \resizebox{\columnwidth}{!}{%
   \begin{tabular}{cl|cccc|cccc}
       \toprule
       & & \multicolumn{4}{c|}{\textbf{multilingual-E5-Small}} & \multicolumn{4}{c}{\textbf{All-MiniLM-L6-v2}} \\
       \cmidrule{3-10}
       \textbf{Dataset} & \textbf{Method} & \textbf{AUROC} & \textbf{AUPR} & \textbf{FPR95} & \textbf{ID Acc} & \textbf{AUROC} & \textbf{AUPR} & \textbf{FPR95} & \textbf{ID Acc} \\
       \midrule
       \multirow{3}{*}{\textbf{Cora}} 
       & Energy & 82.60 & 85.89 & 60.36 & \multirow{2}{*}{82.50} & 81.17 & 83.95 & 60.42 & \multirow{2}{*}{82.64} \\
       & GNNSafe & 90.01 & 91.73 & 36.66 & & 89.19 & 89.58 & 34.21 & \\
       & TNT-OOD & \textbf{89.52} & \textbf{90.31} & \textbf{33.61} & \textbf{83.27} & \textbf{91.65} & \textbf{90.55} & \textbf{24.91} & 82.54 \\
       \midrule
       \multirow{3}{*}{\textbf{PubMed}} 
       & Energy & 71.88 & 68.98 & 74.13 & \multirow{2}{*}{79.07} & 71.18 & 68.03 & 71.40 & \multirow{2}{*}{78.23} \\
       & GNNSafe & 84.73 & 81.44 & 42.74 & & 84.33 & 80.48 & 39.19 & \\
       & TNT-OOD & \textbf{87.29} & \textbf{83.64} & \textbf{35.76} & \textbf{74.93} & \textbf{88.82} & \textbf{86.67} & \textbf{33.13} & 78.85 \\
       \midrule
       \multirow{3}{*}{\textbf{DBLP}} 
       & Energy & 72.82 & 68.43 & 63.66 & \multirow{2}{*}{75.91} & 87.31 & 86.73 & 46.63 & \multirow{2}{*}{77.41} \\
       & GNNSafe & 80.90 & 74.67 & 42.66 & & 88.70 & 81.71 & 30.46 & \\
       & TNT-OOD & \textbf{82.94} & \textbf{76.85} & \textbf{38.29} & \textbf{73.07} & \textbf{89.40} & \textbf{81.91} & \textbf{25.78} & 77.42 \\
       \bottomrule
   \end{tabular}%
   }
  \caption{Performance Comparison of Text Encoders.}
   \label{Table:text_encoder_analysis}
\end{table}

Additionally, to investigate the effectiveness and adptatibility of TNT-OOD and OOD scenarios, we substitute language models from  All-MiniLM-L6-v2 to Multilingual-E5-Small~\cite{e5-small} with the same embedding size. We tuned the learning rate to ensure comparable ID Acc. As in Table~\ref{Table:text_encoder_analysis}, TNT-OOD shows great performance on both PLM text encoders against the best non-propagation and propagation-based baselines.

\begin{figure*}[h]
\begin{tcolorbox}[colback=gray!10!white, colframe=gray!50!black, title=\textbf{LLM Prompt for Thematically-Guided OOD Label Selection}]
\textbf{You are tasked with selecting [NUM\_OOD\_CLASSES] classes from the following list to be designated as Out-of-Distribution (OOD), while the remaining classes will be In-Distribution (ID). Your selection must be based on [SELECTION\_CRITERIA].}

\textbf{Here are the categories and their descriptions:}

\noindent [CATEGORY\_NAMES\_AND\_DESCRIPTIONS]

\textbf{Selection Criteria:}
\begin{itemize}
    \item If [SELECTION\_CRITERIA] is ``thematic\_similarity'': Select OOD classes that are pairwise similar to some of the remaining ID classes. This creates a challenging scenario where semantically related classes are separated into ID and OOD.
    \item If [SELECTION\_CRITERIA] is ``thematic\_dissimilarity'': Select OOD classes that are most thematically dissimilar from the remaining ID classes. This creates clearer semantic boundaries between ID and OOD.
\end{itemize}

\textbf{Important instructions:}
\begin{enumerate}
    \item You must select EXACTLY [NUM\_OOD\_CLASSES] classes as OOD.
    \item Provide your response as a JSON object with two keys: ``id\_classes'' and ``ood\_classes''.
    \item Each key should contain an array of integers representing class indices (0-indexed).
    \item Do not include any explanations, reasoning, or additional text.
\end{enumerate}

\textbf{Example output format:}

\texttt{\{``id\_classes'': [0, 1, 3, 5], ``ood\_classes'': [2, 4, 6]\}}
\end{tcolorbox}
\caption{LLM prompt for OOD class selection.}
\label{prompt:llm_prompt}
\end{figure*}

\begin{figure*}[h]
\begin{tcolorbox}[colback=gray!10!white, colframe=gray!50!black, title=\textbf{LLM Prompt for LLM as OOD Detector on Label Shifts}]
\textbf{You are given a set of known categories indexed from 0 to [NUM\_ID\_CLASSES]-1, where they are in-distribution (ID) classes. For each test case, you will be given a 'Text'. The given set of known categories are: [CATEGORIES]. Your task is: If the text clearly belongs to one of the ID classes, output the numeric label index (from 0 to [NUM\_ID\_CLASSES]-1). If the text does not belong to any ID class (i.e., is out-of-distribution), output [NUM\_ID\_CLASSES].}\\

\textbf{Important instructions:}
Output only the predicted label index (an integer from 0 to [NUM\_ID\_CLASSES]), with no explanation or extra text.\\

\textbf{Example output format:}

\texttt{3}

\end{tcolorbox}
\caption{LLM prompt for LLM as Detector methods on label shifts.}
\label{prompt:llm_prompt_detector}
\end{figure*}

\begin{table*}[h!]
\centering
\resizebox{1\linewidth}{!}{
\begin{tabular}{c|c|ccccc|ccc}
\toprule
\toprule
\multirow{2}{*}{OOD Type} & \multirow{2}{*}{Metric} & \multicolumn{5}{c|}{w/o Prop.} & \multicolumn{3}{c}{w/ Prop.} \\
& & MAHA & MSP & ODIN & NECO & Energy & GNNSafe & NodeSafe & TNT-OOD \\
\midrule
\multirow{4}{*}{Feature (0.50)} & AUROC & 52.80 $\pm$ 1.32 & 68.77 $\pm$ 0.60 & 66.80 $\pm$ 0.51 & 69.37 $\pm$ 3.48 & 78.16 $\pm$ 0.42 & 82.26 $\pm$ 0.77 & 84.36 $\pm$ 0.83 & 91.22 $\pm$ 0.23 \\
& AUPR & 54.18 $\pm$ 0.92 & 70.82 $\pm$ 0.35 & 69.12 $\pm$ 0.08 & 73.27 $\pm$ 2.89 & 81.18 $\pm$ 0.40 & 84.37 $\pm$ 0.03 & 86.03 $\pm$ 0.70 & 89.56 $\pm$ 0.16 \\
& FPR95 & 86.22 $\pm$ 1.18 & 73.55 $\pm$ 1.71 & 75.71 $\pm$ 1.48 & 80.13 $\pm$ 7.35 & 65.99 $\pm$ 1.41 & 61.54 $\pm$ 8.34 & 58.16 $\pm$ 1.96 & 32.08 $\pm$ 2.55 \\

\midrule
\multirow{4}{*}{Feature (0.70)} & AUROC & 49.58 $\pm$ 2.40 & 79.24 $\pm$ 1.02 & 76.69 $\pm$ 1.12 & 75.22 $\pm$ 7.73 & 89.09 $\pm$ 0.21 & 93.37 $\pm$ 0.22 & 93.92 $\pm$ 0.53 & 97.22 $\pm$ 0.19 \\
& AUPR & 51.81 $\pm$ 1.38 & 80.61 $\pm$ 0.95 & 78.06 $\pm$ 0.94 & 79.53 $\pm$ 6.34 & 91.11 $\pm$ 0.21 & 94.33 $\pm$ 0.24 & 94.57 $\pm$ 0.30 & 95.89 $\pm$ 0.22 \\
& FPR95 & 85.82 $\pm$ 1.40 & 55.29 $\pm$ 2.93 & 60.01 $\pm$ 2.53 & 69.91 $\pm$ 13.27 & 42.26 $\pm$ 3.45 & 28.08 $\pm$ 2.09 & 27.61 $\pm$ 4.07 & 8.48 $\pm$ 1.35 \\
\bottomrule
\bottomrule
\end{tabular}
}
\caption{Performance comparison of OOD detection methods on \textbf{Cora dataset} across different OOD shifts and configurations. MAHA is short for Mahalanobis.}
\label{table:cora_ood_full}
\end{table*}

\begin{table*}[h!]
\centering
\resizebox{1\linewidth}{!}{
\begin{tabular}{c|c|ccccc|ccc}
\toprule
\toprule
\multirow{2}{*}{OOD Type} & \multirow{2}{*}{Metric} & \multicolumn{5}{c|}{w/o Prop.} & \multicolumn{3}{c}{w/ Prop.} \\
& & MAHA & MSP & ODIN & NECO & Energy & GNNSafe & NodeSafe & TNT-OOD \\
\midrule
\multirow{3}{*}{Feature (0.50)} & AUROC & 74.14 $\pm$ 0.09 & 78.06 $\pm$ 0.27 & 76.59 $\pm$ 0.25 & 69.42 $\pm$ 2.61 & 81.87 $\pm$ 0.27 & 82.43 $\pm$ 0.83 & 79.95 $\pm$ 3.56 & 84.23 $\pm$ 6.18 \\
& AUPR & 70.51 $\pm$ 0.94 & 78.86 $\pm$ 0.27 & 77.5 $\pm$ 0.31 & 71.34 $\pm$ 2.32 & 82.16 $\pm$ 0.09 & 79.05 $\pm$ 0.53 & 76.21 $\pm$ 3.78 & 78.89 $\pm$ 6.01 \\
& FPR95 & 61.04 $\pm$ 0.29 & 62.95 $\pm$ 0.99 & 66.03 $\pm$ 0.77 & 77.73 $\pm$ 4.8 & 60.56 $\pm$ 1.82 & 52.47 $\pm$ 4.11 & 55.85 $\pm$ 5.62 & 38.11 $\pm$ 11.54 \\
\midrule
\multirow{3}{*}{Feature (0.70)} & AUROC & 77.41 $\pm$ 0.26 & 88.38 $\pm$ 0.57 & 87.06 $\pm$ 0.51 & 78.71 $\pm$ 5.74 & 90.31 $\pm$ 0.18 & 91.2 $\pm$ 0.74 & 88.26 $\pm$ 3.11 & 91.85 $\pm$ 3.62 \\
& AUPR & 72.91 $\pm$ 0.85 & 89.14 $\pm$ 0.44 & 87.9 $\pm$ 0.41 & 80.05 $\pm$ 5.96 & 90.6 $\pm$ 0.19 & 86.61 $\pm$ 0.62 & 82.93 $\pm$ 3.69 & 86.63 $\pm$ 3.63 \\
& FPR95 & 55.2 $\pm$ 1.03 & 42.97 $\pm$ 1.58 & 46.47 $\pm$ 1.27 & 65.55 $\pm$ 10.61 & 40.59 $\pm$ 2.34 & 27.06 $\pm$ 4.12 & 30.24 $\pm$ 5.71 & 22.19 $\pm$ 11.22 \\
\bottomrule
\bottomrule
\end{tabular}
}
\caption{Performance comparison of OOD detection methods on \textbf{Citeseer dataset} across different OOD shifts and configurations.}
\label{table:citeseer_ood_full}
\end{table*}

\begin{table*}[h!]
\centering
\resizebox{1\linewidth}{!}{
\begin{tabular}{c|c|ccccc|ccc}
\toprule
\toprule
\multirow{2}{*}{OOD Type} & \multirow{2}{*}{Metric} & \multicolumn{5}{c|}{w/o Prop.} & \multicolumn{3}{c}{w/ Prop.} \\
& & MAHA & MSP & ODIN & NECO & Energy & GNNSafe & NodeSafe & TNT-OOD \\
\midrule
\multirow{3}{*}{Feature (0.50)} & AUROC & 89.92 $\pm$ 0.23 & 65.82 $\pm$ 0.02 & 64.10 $\pm$ 0.05 & 69.06 $\pm$ 0.16 & 71.08 $\pm$ 0.23 & 44.97 $\pm$ 0.07 & 44.89 $\pm$ 0.09 & 54.98 $\pm$ 1.42 \\
& AUPR & 96.73 $\pm$ 0.07 & 87.31 $\pm$ 0.03 & 86.49 $\pm$ 0.01 & 89.16 $\pm$ 0.04 & 90.29 $\pm$ 0.07 & 80.16 $\pm$ 0.04 & 80.09 $\pm$ 0.03 & 84.03 $\pm$ 0.50 \\
& FPR95 & 59.73 $\pm$ 1.35 & 92.93 $\pm$ 0.16 & 94.48 $\pm$ 0.14 & 94.87 $\pm$ 0.17 & 94.94 $\pm$ 0.15 & 95.88 $\pm$ 0.13 & 96.08 $\pm$ 0.09 & 85.40 $\pm$ 3.03 \\
\midrule
\multirow{3}{*}{Feature (0.70)} & AUROC & 91.90 $\pm$ 0.23 & 72.14 $\pm$ 0.13 & 70.76 $\pm$ 0.08 & 77.19 $\pm$ 0.16 & 79.59 $\pm$ 0.24 & 46.08 $\pm$ 0.08 & 45.98 $\pm$ 0.09 & 58.29 $\pm$ 1.47 \\
& AUPR & 97.39 $\pm$ 0.07 & 89.88 $\pm$ 0.07 & 89.29 $\pm$ 0.05 & 92.19 $\pm$ 0.04 & 93.37 $\pm$ 0.07 & 80.14 $\pm$ 0.04 & 80.15 $\pm$ 0.02 & 85.32 $\pm$ 0.49 \\
& FPR95 & 42.58 $\pm$ 1.75 & 83.09 $\pm$ 0.34 & 86.07 $\pm$ 0.24 & 84.60 $\pm$ 0.55 & 84.96 $\pm$ 0.51 & 90.92 $\pm$ 0.20 & 90.55 $\pm$ 0.09 & 73.91 $\pm$ 3.65 \\
\bottomrule
\bottomrule
\end{tabular}
}
\caption{Performance comparison of OOD detection methods on \textbf{Arxiv dataset} across different OOD shifts and configurations.}
\label{table:arxiv_ood_full}
\end{table*}

\begin{table*}[h!]
\centering
\resizebox{1\linewidth}{!}{
\begin{tabular}{c|c|ccccc|ccc}
\toprule
\toprule
\multirow{2}{*}{OOD Type} & \multirow{2}{*}{Metric} & \multicolumn{5}{c|}{w/o Prop.} & \multicolumn{3}{c}{w/ Prop.} \\
& & MAHA & MSP & ODIN & NECO & Energy & GNNSafe & NodeSafe & TNT-OOD \\
\midrule
\multirow{3}{*}{Feature (0.50)} & AUROC & 72.25 $\pm$ 0.54 & 80.08 $\pm$ 2.10 & 77.91 $\pm$ 2.00 & 79.85 $\pm$ 5.36 & 85.43 $\pm$ 8.88 & 88.54 $\pm$ 9.34 & 90.29 $\pm$ 9.64 & 89.06 $\pm$ 1.16 \\
& AUPR & 66.23 $\pm$ 0.53 & 78.59 $\pm$ 1.73 & 75.60 $\pm$ 1.62 & 78.64 $\pm$ 4.96 & 85.09 $\pm$ 11.85 & 82.74 $\pm$ 9.95 & 83.66 $\pm$ 9.86 & 82.66 $\pm$ 1.11 \\
& FPR95 & 82.39 $\pm$ 1.13 & 62.10 $\pm$ 5.84 & 65.42 $\pm$ 5.71 & 62.64 $\pm$ 8.88 & 55.12 $\pm$ 7.63 & 42.40 $\pm$ 13.10 & 32.93 $\pm$ 16.75 & 35.09 $\pm$ 2.64 \\
\midrule
\multirow{3}{*}{Feature (0.70)} & AUROC & 75.05 $\pm$ 0.66 & 90.10 $\pm$ 1.52 & 87.84 $\pm$ 1.57 & 88.99 $\pm$ 0.44 & 93.31 $\pm$ 10.47 & 93.89 $\pm$ 9.47 & 94.32 $\pm$ 9.53 & 93.66 $\pm$ 0.41 \\
& AUPR & 68.40 $\pm$ 0.70 & 89.99 $\pm$ 1.06 & 86.83 $\pm$ 1.10 & 89.19 $\pm$ 0.67 & 93.74 $\pm$ 15.03 & 87.15 $\pm$ 10.85 & 87.16 $\pm$ 10.63 & 86.85 $\pm$ 0.37 \\
& FPR95 & 80.37 $\pm$ 1.06 & 40.68 $\pm$ 5.57 & 44.71 $\pm$ 5.45 & 44.65 $\pm$ 5.15 & 35.74 $\pm$ 12.70 & 14.05 $\pm$ 20.05 & 8.64 $\pm$ 21.20 & 13.88 $\pm$ 3.24 \\
\bottomrule
\bottomrule
\end{tabular}
}
\caption{Performance comparison of OOD detection methods on \textbf{DBLP dataset} across different OOD shifts and configurations.}
\label{table:dblp_ood_full}
\end{table*}

\begin{table*}[h!]
\centering
\resizebox{1\linewidth}{!}{
\begin{tabular}{c|c|ccccc|ccc}
\toprule
\toprule
\multirow{2}{*}{OOD Type} & \multirow{2}{*}{Metric} & \multicolumn{5}{c|}{w/o Prop.} & \multicolumn{3}{c}{w/ Prop.} \\
& & MAHA & MSP & ODIN & NECO & Energy & GNNSafe & NodeSafe & TNT-OOD \\
\midrule
\multirow{3}{*}{Feature (0.50)} & AUROC & 79.57 $\pm$ 0.54 & 71.24 $\pm$ 0.51 & 68.81 $\pm$ 0.65 & 62.69 $\pm$ 5.54 & 72.43 $\pm$ 1.48 & 77.47 $\pm$ 2.08 & 77.96 $\pm$ 2.58 & 84.04 $\pm$ 1.80 \\
& AUPR & 77.08 $\pm$ 0.82 & 67.25 $\pm$ 0.46 & 65.34 $\pm$ 0.51 & 61.67 $\pm$ 4.63 & 70.16 $\pm$ 1.15 & 76.68 $\pm$ 1.77 & 76.52 $\pm$ 1.88 & 82.75 $\pm$ 0.53 \\
& FPR95 & 81.08 $\pm$ 2.29 & 71.15 $\pm$ 1.39 & 77.18 $\pm$ 1.17 & 83.88 $\pm$ 3.24 & 77.82 $\pm$ 2.31 & 59.22 $\pm$ 2.47 & 58.39 $\pm$ 2.30 & 49.64 $\pm$ 6.66 \\
\midrule
\multirow{3}{*}{Feature (0.70)} & AUROC & 81.92 $\pm$ 0.59 & 78.50 $\pm$ 0.64 & 76.31 $\pm$ 0.80 & 71.16 $\pm$ 1.22 & 78.63 $\pm$ 1.93 & 85.90 $\pm$ 2.32 & 86.38 $\pm$ 2.98 & 93.26 $\pm$ 1.17 \\
& AUPR & 78.81 $\pm$ 1.03 & 73.33 $\pm$ 0.69 & 71.91 $\pm$ 0.71 & 68.33 $\pm$ 0.21 & 74.27 $\pm$ 2.09 & 83.04 $\pm$ 2.66 & 82.97 $\pm$ 3.17 & 92.31 $\pm$ 0.98 \\
& FPR95 & 72.57 $\pm$ 3.70 & 54.01 $\pm$ 1.20 & 61.24 $\pm$ 1.24 & 69.90 $\pm$ 3.53 & 61.41 $\pm$ 3.71 & 36.53 $\pm$ 3.54 & 35.28 $\pm$ 4.04 & 22.64 $\pm$ 4.33 \\
\bottomrule
\bottomrule
\end{tabular}
}
\caption{Performance comparison of OOD detection methods on \textbf{PubMed dataset} across different OOD shifts and configurations.}
\label{table:pubmed_ood_full}
\end{table*}

\begin{table*}[h!]
\centering
\resizebox{1\linewidth}{!}{
\begin{tabular}{c|c|ccccc|ccc}
\toprule
\toprule
\multirow{2}{*}{OOD Type} & \multirow{2}{*}{Metric} & \multicolumn{5}{c|}{w/o Prop.} & \multicolumn{3}{c}{w/ Prop.} \\
& & MAHA & MSP & ODIN & NECO & Energy & GNNSafe & NodeSafe & TNT-OOD \\
\midrule
\multirow{3}{*}{Feature (0.50)} & AUROC & 52.44 $\pm$ 4.42 & 48.52 $\pm$ 10.26 & 53.00 $\pm$ 10.29 & 47.70 $\pm$ 10.73 & 51.00 $\pm$ 11.35 & 65.45 $\pm$ 25.84 & 49.36 $\pm$ 1.63 & 76.44 $\pm$ 2.30 \\
& AUPR & 54.83 $\pm$ 3.38 & 55.88 $\pm$ 6.55 & 58.93 $\pm$ 7.57 & 55.46 $\pm$ 6.60 & 55.46 $\pm$ 7.33 & 69.94 $\pm$ 20.29 & 53.84 $\pm$ 1.42 & 78.38 $\pm$ 2.56 \\
& FPR95 & 86.72 $\pm$ 1.48 & 90.05 $\pm$ 5.98 & 88.67 $\pm$ 4.67 & 90.20 $\pm$ 6.17 & 85.25 $\pm$ 8.34 & 63.98 $\pm$ 42.59 & 95.10 $\pm$ 0.12 & 84.41 $\pm$ 0.87 \\
\midrule
\multirow{3}{*}{Feature (0.70)} & AUROC & 51.60 $\pm$ 6.24 & 48.96 $\pm$ 11.44 & 55.36 $\pm$ 13.72 & 48.02 $\pm$ 11.93 & 51.24 $\pm$ 12.31 & 66.01 $\pm$ 26.33 & 50.23 $\pm$ 1.99 & 84.86 $\pm$ 2.42 \\
& AUPR & 53.99 $\pm$ 4.66 & 56.31 $\pm$ 7.29 & 61.00 $\pm$ 10.14 & 55.74 $\pm$ 7.30 & 55.49 $\pm$ 7.81 & 70.32 $\pm$ 20.39 & 54.26 $\pm$ 1.71 & 88.05 $\pm$ 3.18 \\
& FPR95 & 85.71 $\pm$ 2.12 & 88.84 $\pm$ 7.74 & 85.36 $\pm$ 8.90 & 88.96 $\pm$ 7.94 & 84.03 $\pm$ 9.05 & 63.79 $\pm$ 42.50 & 94.89 $\pm$ 0.13 & 77.23 $\pm$ 1.52 \\
\bottomrule
\bottomrule
\end{tabular}
}
\caption{Performance comparison of OOD detection methods on \textbf{Reddit dataset} across different OOD shifts and configurations.}
\label{table:reddit_ood_full}
\end{table*}

\begin{table*}[h!]
\centering
\resizebox{1\linewidth}{!}{
\begin{tabular}{c|c|ccccc|ccc}
\toprule
\toprule
\multirow{2}{*}{OOD Type} & \multirow{2}{*}{Metric} & \multicolumn{5}{c|}{w/o Prop.} & \multicolumn{3}{c}{w/ Prop.} \\
& & MAHA & MSP & ODIN & NECO & Energy & GNNSafe & NodeSafe & TNT-OOD \\
\midrule
\multirow{3}{*}{Feature (0.50)} & AUROC & 64.60 $\pm$ 0.74 & 65.46 $\pm$ 0.62 & 62.03 $\pm$ 0.71 & 62.19 $\pm$ 2.48 & 67.71 $\pm$ 0.60 & 82.58 $\pm$ 0.98 & 84.92 $\pm$ 1.33 & 91.69 $\pm$ 0.15 \\
& AUPR & 76.28 $\pm$ 0.86 & 76.84 $\pm$ 0.27 & 74.01 $\pm$ 0.39 & 73.96 $\pm$ 2.45 & 79.24 $\pm$ 0.49 & 87.07 $\pm$ 0.57 & 89.24 $\pm$ 0.65 & 92.05 $\pm$ 0.05 \\
& FPR95 & 89.68 $\pm$ 0.36 & 80.34 $\pm$ 0.90 & 83.11 $\pm$ 0.99 & 84.82 $\pm$ 0.95 & 80.65 $\pm$ 0.61 & 62.59 $\pm$ 3.49 & 71.80 $\pm$ 1.02 & 29.10 $\pm$ 2.99 \\
\midrule
\multirow{3}{*}{Feature (0.70)} & AUROC & 68.79 $\pm$ 1.73 & 76.27 $\pm$ 0.43 & 69.96 $\pm$ 0.70 & 72.49 $\pm$ 3.01 & 77.99 $\pm$ 0.77 & 92.55 $\pm$ 0.46 & 91.78 $\pm$ 1.46 & 96.05 $\pm$ 0.15 \\
& AUPR & 78.97 $\pm$ 1.90 & 84.00 $\pm$ 0.17 & 79.00 $\pm$ 0.15 & 81.36 $\pm$ 3.11 & 86.40 $\pm$ 0.66 & 92.65 $\pm$ 0.18 & 92.25 $\pm$ 0.76 & 94.13 $\pm$ 0.05 \\
& FPR95 & 86.63 $\pm$ 0.47 & 62.89 $\pm$ 2.56 & 73.23 $\pm$ 1.96 & 79.68 $\pm$ 3.64 & 67.42 $\pm$ 1.59 & 22.58 $\pm$ 1.83 & 34.49 $\pm$ 11.90 & 6.01 $\pm$ 0.11 \\
\bottomrule
\bottomrule
\end{tabular}
}
\caption{Performance comparison of OOD detection methods on \textbf{WikiCS dataset} across different OOD shifts and configurations.}
\label{table:wikics_ood_full}
\end{table*}

\begin{table*}[h!]
\centering
\resizebox{1\linewidth}{!}{
\begin{tabular}{c|c|ccccc|ccc}
\toprule
\toprule
\multirow{2}{*}{OOD Type} & \multirow{2}{*}{Metric} & \multicolumn{5}{c|}{w/o Prop.} & \multicolumn{3}{c}{w/ Prop.} \\
& & MAHA & MSP & ODIN & NECO & Energy & GNNSafe & NodeSafe & TNT-OOD \\
\midrule
\multirow{3}{*}{Feature (0.50)} & AUROC & 84.88 $\pm$ 0.68 & 60.38 $\pm$ 0.71 & 58.73 $\pm$ 0.68 & 62.06 $\pm$ 0.92 & 62.66 $\pm$ 0.96 & 61.82 $\pm$ 0.76 & 60.37 $\pm$ 0.70 & 76.98 $\pm$ 1.44 \\
& AUPR & 95.89 $\pm$ 0.19 & 87.46 $\pm$ 0.26 & 86.58 $\pm$ 0.27 & 88.24 $\pm$ 0.33 & 88.61 $\pm$ 0.35 & 87.12 $\pm$ 0.23 & 86.45 $\pm$ 0.25 & 91.71 $\pm$ 0.39 \\
& FPR95 & 48.00 $\pm$ 1.66 & 88.73 $\pm$ 1.06 & 89.88 $\pm$ 1.02 & 89.75 $\pm$ 1.05 & 89.79 $\pm$ 1.14 & 70.92 $\pm$ 1.21 & 82.61 $\pm$ 0.97 & 46.20 $\pm$ 3.31 \\
\midrule
\multirow{3}{*}{Feature (0.70)} & AUROC & 87.53 $\pm$ 0.69 & 60.04 $\pm$ 1.03 & 57.95 $\pm$ 0.98 & 63.08 $\pm$ 1.07 & 64.44 $\pm$ 1.08 & 64.08 $\pm$ 0.84 & 61.57 $\pm$ 0.92 & 80.46 $\pm$ 1.41 \\
& AUPR & 96.81 $\pm$ 0.20 & 86.35 $\pm$ 0.50 & 85.27 $\pm$ 0.50 & 88.08 $\pm$ 0.46 & 89.11 $\pm$ 0.44 & 87.38 $\pm$ 0.26 & 86.47 $\pm$ 0.31 & 92.60 $\pm$ 0.41 \\
& FPR95 & 46.57 $\pm$ 1.65 & 86.51 $\pm$ 0.75 & 88.49 $\pm$ 1.06 & 88.33 $\pm$ 0.98 & 88.31 $\pm$ 1.02 & 65.84 $\pm$ 1.40 & 77.09 $\pm$ 1.48 & 36.06 $\pm$ 3.93 \\
\bottomrule
\bottomrule
\end{tabular}
}
\caption{Performance comparison of OOD detection methods on \textbf{BookHis dataset} across different OOD shifts and configurations.}
\label{table:bookhis_ood_full}
\end{table*}

\begin{table*}[h!]
\centering
\resizebox{1\linewidth}{!}{
\begin{tabular}{c|c|ccccc|ccc}
\toprule
\toprule
\multirow{2}{*}{OOD Type} & \multirow{2}{*}{Metric} & \multicolumn{5}{c|}{w/o Prop.} & \multicolumn{3}{c}{w/ Prop.} \\
& & MAHA & MSP & ODIN & NECO & Energy & GNNSafe & NodeSafe & TNT-OOD \\
\midrule
\multirow{3}{*}{Feature (0.50)} & AUROC & 78.01 $\pm$ 0.44 & 64.64 $\pm$ 0.89 & 61.26 $\pm$ 1.08 & 70.09 $\pm$ 0.97 & 70.76 $\pm$ 0.96 & 74.84 $\pm$ 0.89 & 61.87 $\pm$ 3.68 & 81.16 $\pm$ 2.66 \\
& AUPR & 93.24 $\pm$ 0.14 & 88.58 $\pm$ 0.43 & 86.91 $\pm$ 0.46 & 90.64 $\pm$ 0.40 & 91.12 $\pm$ 0.37 & 91.08 $\pm$ 0.28 & 86.73 $\pm$ 1.16 & 92.77 $\pm$ 0.83 \\
& FPR95 & 56.93 $\pm$ 0.95 & 81.84 $\pm$ 0.87 & 84.26 $\pm$ 1.35 & 78.09 $\pm$ 1.43 & 78.07 $\pm$ 1.54 & 61.11 $\pm$ 1.46 & 78.84 $\pm$ 2.97 & 35.42 $\pm$ 5.42 \\
\midrule
\multirow{3}{*}{Feature (0.70)} & AUROC & 82.07 $\pm$ 0.38 & 69.01 $\pm$ 1.33 & 64.08 $\pm$ 1.52 & 74.85 $\pm$ 1.13 & 75.58 $\pm$ 1.07 & 80.22 $\pm$ 0.79 & 65.52 $\pm$ 5.14 & 83.88 $\pm$ 3.14 \\
& AUPR & 94.76 $\pm$ 0.13 & 89.78 $\pm$ 0.69 & 87.37 $\pm$ 0.67 & 92.29 $\pm$ 0.50 & 92.92 $\pm$ 0.40 & 92.27 $\pm$ 0.24 & 87.53 $\pm$ 1.54 & 93.52 $\pm$ 0.93 \\
& FPR95 & 52.06 $\pm$ 0.38 & 74.97 $\pm$ 0.80 & 79.11 $\pm$ 2.07 & 70.11 $\pm$ 2.04 & 69.93 $\pm$ 2.26 & 39.11 $\pm$ 2.21 & 70.38 $\pm$ 4.87 & 23.91 $\pm$ 8.06 \\
\bottomrule
\bottomrule
\end{tabular}
}
\caption{Performance comparison of OOD detection methods on \textbf{BookChild dataset} across different OOD shifts and configurations.}
\label{table:bookchild_ood_full}
\end{table*}

\begin{table*}[h!]
\centering
\resizebox{1\linewidth}{!}{
\begin{tabular}{c|c|ccccc|ccc}
\toprule
\toprule
\multirow{2}{*}{OOD Type} & \multirow{2}{*}{Metric} & \multicolumn{5}{c|}{w/o Prop.} & \multicolumn{3}{c}{w/ Prop.} \\
& & MAHA & MSP & ODIN & NECO & Energy & GNNSafe & NodeSafe & TNT-OOD \\
\midrule
\multirow{3}{*}{Feature (0.50)} & AUROC & 69.18 $\pm$ 0.51 & 65.44 $\pm$ 0.18 & 64.30 $\pm$ 0.32 & 68.04 $\pm$ 0.54 & 68.34 $\pm$ 0.55 & 66.29 $\pm$ 0.34 & 61.19 $\pm$ 0.21 & 73.13 $\pm$ 0.62 \\
& AUPR & 89.82 $\pm$ 0.10 & 88.81 $\pm$ 0.07 & 88.58 $\pm$ 0.15 & 90.05 $\pm$ 0.24 & 90.42 $\pm$ 0.26 & 88.33 $\pm$ 0.20 & 87.32 $\pm$ 0.20 & 91.97 $\pm$ 0.25 \\
& FPR95 & 72.72 $\pm$ 2.09 & 84.24 $\pm$ 0.25 & 86.55 $\pm$ 0.08 & 86.13 $\pm$ 0.59 & 86.28 $\pm$ 0.54 & 63.75 $\pm$ 0.82 & 80.31 $\pm$ 0.09 & 47.91 $\pm$ 1.97 \\
\midrule
\multirow{3}{*}{Feature (0.70)} & AUROC & 70.67 $\pm$ 0.52 & 65.73 $\pm$ 0.33 & 64.44 $\pm$ 0.28 & 71.28 $\pm$ 0.41 & 72.00 $\pm$ 0.44 & 68.51 $\pm$ 0.23 & 65.23 $\pm$ 0.28 & 75.18 $\pm$ 0.53 \\
& AUPR & 90.59 $\pm$ 0.13 & 88.87 $\pm$ 0.11 & 88.50 $\pm$ 0.10 & 91.16 $\pm$ 0.16 & 91.90 $\pm$ 0.18 & 88.73 $\pm$ 0.16 & 88.04 $\pm$ 0.17 & 92.72 $\pm$ 0.20 \\
& FPR95 & 70.07 $\pm$ 1.83 & 83.79 $\pm$ 0.66 & 87.02 $\pm$ 0.57 & 80.90 $\pm$ 0.84 & 81.13 $\pm$ 0.70 & 54.22 $\pm$ 0.44 & 63.58 $\pm$ 0.63 & 41.30 $\pm$ 1.53 \\
\bottomrule
\bottomrule
\end{tabular}
}
\caption{Performance comparison of OOD detection methods on \textbf{ElePhoto dataset} across different OOD shifts and configurations.}
\label{table:elephoto_ood_full}
\end{table*}

\begin{table*}[h!]
\centering
\resizebox{1\linewidth}{!}{
\begin{tabular}{c|c|ccccc|ccc}
\toprule
\toprule
\multirow{2}{*}{OOD Type} & \multirow{2}{*}{Metric} & \multicolumn{5}{c|}{w/o Prop.} & \multicolumn{3}{c}{w/ Prop.} \\
& & MAHA & MSP & ODIN & NECO & Energy & GNNSafe & NodeSafe & TNT-OOD \\
\midrule
\multirow{3}{*}{Feature (0.50)} & AUROC & 43.68 $\pm$ 1.27 & 60.82 $\pm$ 0.52 & 58.66 $\pm$ 0.49 & 58.57 $\pm$ 0.39 & 57.93 $\pm$ 0.34 & 61.49 $\pm$ 0.25 & 61.83 $\pm$ 0.30 & 73.64 $\pm$ 0.13 \\
& AUPR & 86.70 $\pm$ 0.40 & 92.38 $\pm$ 0.23 & 92.04 $\pm$ 0.23 & 91.33 $\pm$ 0.16 & 90.80 $\pm$ 0.09 & 90.69 $\pm$ 0.01 & 92.10 $\pm$ 0.03 & 95.78 $\pm$ 0.04 \\
& FPR95 & 97.54 $\pm$ 0.16 & 89.51 $\pm$ 0.32 & 90.87 $\pm$ 0.29 & 91.80 $\pm$ 0.37 & 91.85 $\pm$ 0.36 & 75.82 $\pm$ 0.40 & 76.73 $\pm$ 0.93 & 53.69 $\pm$ 0.57 \\
\midrule
\multirow{3}{*}{Feature (0.70)} & AUROC & 42.86 $\pm$ 1.82 & 68.76 $\pm$ 0.73 & 67.76 $\pm$ 0.62 & 64.96 $\pm$ 0.79 & 63.93 $\pm$ 0.76 & 66.17 $\pm$ 0.44 & 66.78 $\pm$ 0.25 & 76.45 $\pm$ 0.08 \\
& AUPR & 86.17 $\pm$ 0.59 & 94.21 $\pm$ 0.28 & 94.31 $\pm$ 0.21 & 93.04 $\pm$ 0.29 & 92.36 $\pm$ 0.19 & 91.27 $\pm$ 0.05 & 92.72 $\pm$ 0.03 & 96.27 $\pm$ 0.04 \\
& FPR95 & 97.45 $\pm$ 0.30 & 76.36 $\pm$ 0.84 & 77.50 $\pm$ 0.82 & 80.87 $\pm$ 1.35 & 81.15 $\pm$ 1.33 & 56.57 $\pm$ 1.37 & 51.13 $\pm$ 0.67 & 43.30 $\pm$ 0.15 \\
\bottomrule
\bottomrule
\end{tabular}
}
\caption{Performance comparison of OOD detection methods on \textbf{EleComp dataset} across different OOD shifts and configurations.}
\label{table:elecomp_ood_full}
\end{table*}

\end{document}